\documentclass[10pt,twocolumn,letterpaper]{article}

\usepackage{cvpr}

\makeatletter
\@namedef{ver@everyshi.sty}{}
\makeatother

\usepackage{graphicx}
\usepackage{amsmath}
\usepackage{amssymb}
\usepackage{booktabs}
\usepackage{comment}
\usepackage{color}
\usepackage{tabularx}
\usepackage{colortbl}
\usepackage{tabu}
\usepackage{multirow}
\usepackage{setspace}
\usepackage[font=small,labelsep=period]{caption}
\usepackage[dvipsnames]{xcolor}
\usepackage{tikz}
\usepackage{pgfplots}
\usepackage{pgfkeys}
\usepackage{xcolor}
\usepackage{xspace}
\usepackage{cuted}
\usepackage{arydshln}
\usepackage{makecell}
\usepackage{pifont}%

\usepackage[accsupp]{axessibility}  %

\usepackage[moderate]{savetrees}

\definecolor{bblue}{rgb}{0.0,0.2,0.8}
\definecolor{myred}{rgb}{0.9,0.1,0.1}
\definecolor{ccol}{rgb}{0.91,0.91,0.91}
\usepackage[pagebackref=true,breaklinks=true,colorlinks=true,urlcolor=magenta,bookmarks=false,citecolor=bblue]{hyperref}

\usepackage[accsupp]{axessibility} %

\newcolumntype{Y}{>{\centering\arraybackslash}X}
\newcolumntype{R}{>{\raggedleft\arraybackslash}X}
\newcolumntype{L}{>{\raggedright\arraybackslash}X}

\definecolor{lightgray}{rgb}{0.835, 0.835, 0.835}
\definecolor{lightergray}{rgb}{0.935, 0.935, 0.935}
\definecolor{lightgreen}{rgb}{0.85, 1.0, 0.85}

\makeatletter
\def\adl@drawiv#1#2#3{%
        \hskip.5\tabcolsep
        \xleaders#3{#2.5\@tempdimb #1{1}#2.5\@tempdimb}%
                #2\z@ plus1fil minus1fil\relax
        \hskip.5\tabcolsep}
\newcommand{\cdashlinelr}[1]{%
  \noalign{\vskip\aboverulesep
           \global\let\@dashdrawstore\adl@draw
           \global\let\adl@draw\adl@drawiv}
  \cdashline{#1}
  \noalign{\global\let\adl@draw\@dashdrawstore
           \vskip\belowrulesep}}
\makeatother

\makeatletter
\newlength{\qrr@dimen@}
\expandafter\pretocmd\csname tabular*\endcsname{\setlength{\qrr@dimen@}{#1}}{}{}
\newcommand*{\Rowcolor}[2][\tabcolsep]{%
    \ifx\relax#1\relax\else
        \kern-\the\dimexpr#1\relax
    \fi
    \makebox[0pt][l]{%
        \fboxsep=0pt
        \colorbox{#2}{%
            \strut\kern\qrr@dimen@
        }%
    }%
    \ifx\relax#1\relax\else
        \kern\the\dimexpr#1\relax
    \fi
    \ignorespaces
}
\makeatother

\newcommand\customparagraph[1]{\vspace{0.7em}\noindent\textbf{#1}}
\newcommand{\mytilde}{\raise.17ex\hbox{$\scriptstyle\sim$}}

\newcommand{\cmark}{\ding{51}}%
\newcommand{\xmark}{\ding{55}}%

\def\addlegendimage{\csname pgfplots@addlegendimage\endcsname}

\usepackage[capitalize]{cleveref}
\crefname{section}{Sec.}{Secs.}
\Crefname{section}{Section}{Sections}
\Crefname{table}{Table}{Tables}
\crefname{table}{Tab.}{Tabs.}

\makeatletter
\renewcommand{\fnum@figure}{Figure \thefigure}
\makeatother

\def\cvprPaperID{7209}
\def\confName{CVPR}
\def\confYear{2025}

\usepackage{cuted}
\usepackage{capt-of}

\begin{document}

\title{HOT3D: Hand and Object Tracking in 3D from Egocentric Multi-View Videos}

\newcommand{\namesep}{, }

\author{
Prithviraj Banerjee\namesep
Sindi Shkodrani\namesep
Pierre Moulon\namesep
Shreyas Hampali\namesep
Shangchen Han,\vspace{0.2em}\\
Fan Zhang\namesep
Linguang Zhang\namesep
Jade Fountain\namesep
Edward Miller\namesep
Selen Basol,\vspace{0.2em}\\
Richard Newcombe\namesep
Robert Wang\namesep
Jakob Julian Engel\namesep
Tomas Hodan\vspace{9.3pt} \\
{\normalsize Meta Reality Labs\hspace{1.6em}\href{https://facebookresearch.github.io/hot3d/}{facebookresearch.github.io/hot3d}}
}

\maketitle

\begin{strip}
\begin{minipage}{\textwidth}\centering
\vspace{-30.5pt}
\centering
\begin{minipage}{0.195\linewidth}%
\includegraphics[width=\linewidth]{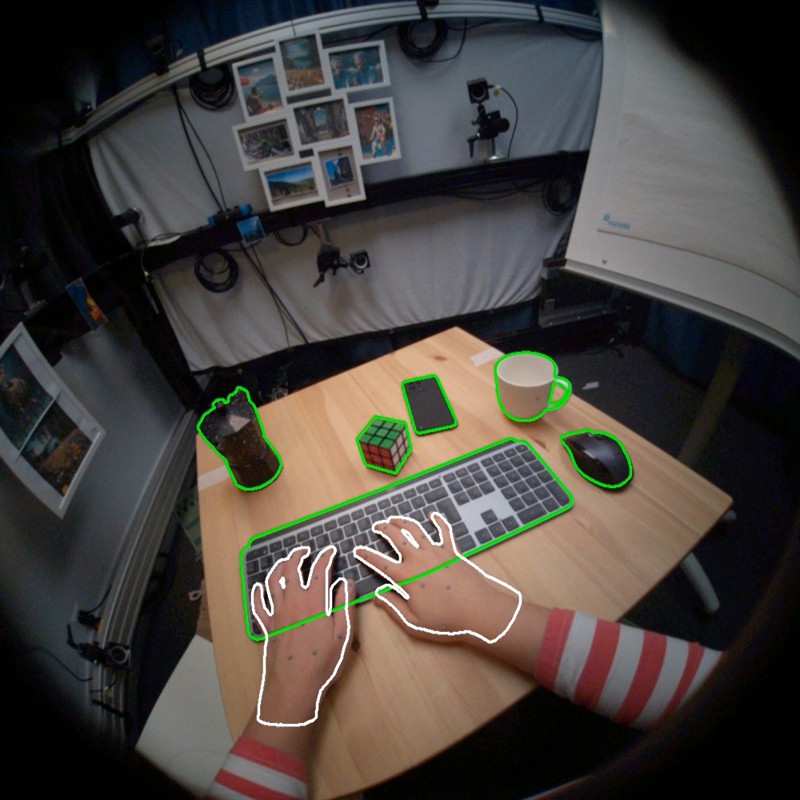}\\[0.2mm]
\includegraphics[width=0.49\linewidth]{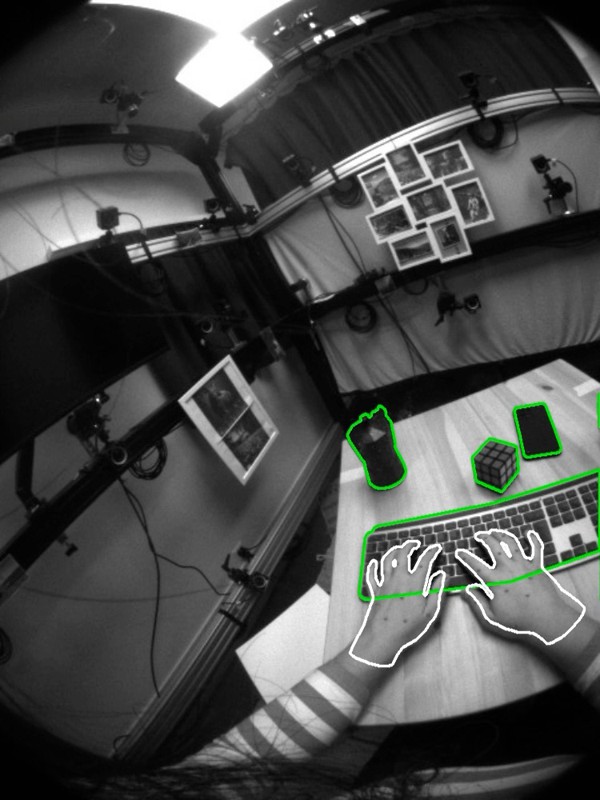}
\includegraphics[width=0.49\linewidth]{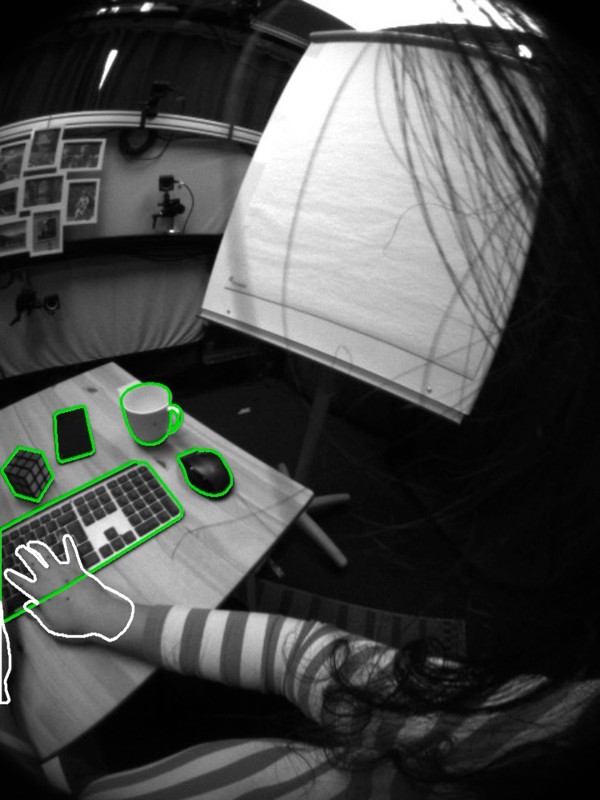}%
\end{minipage}\hfill%
\begin{minipage}{0.195\linewidth}%
\includegraphics[width=\linewidth]{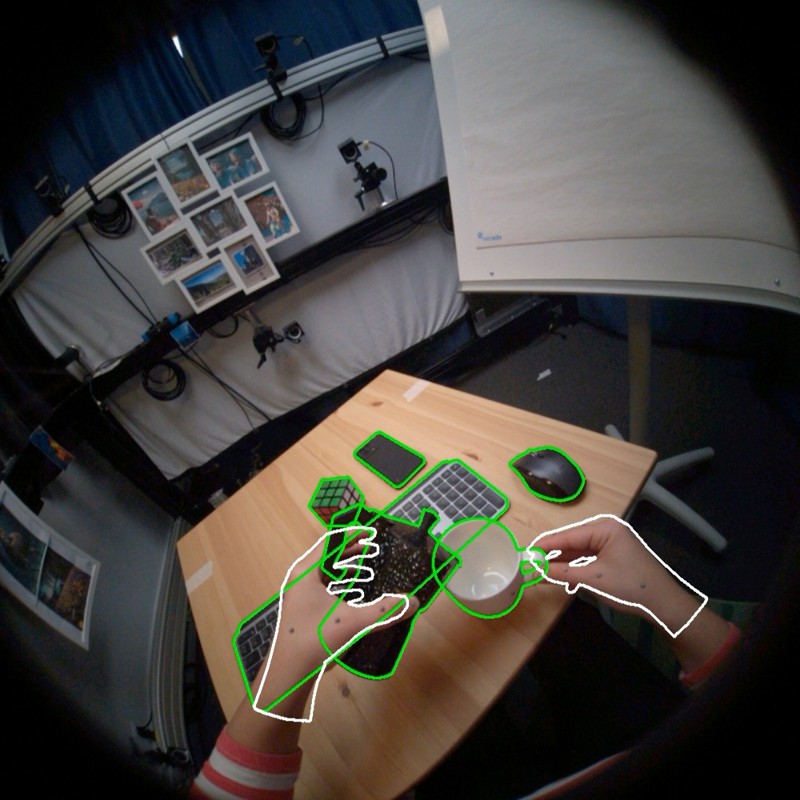}\\[0.2mm]
\includegraphics[width=0.49\linewidth]{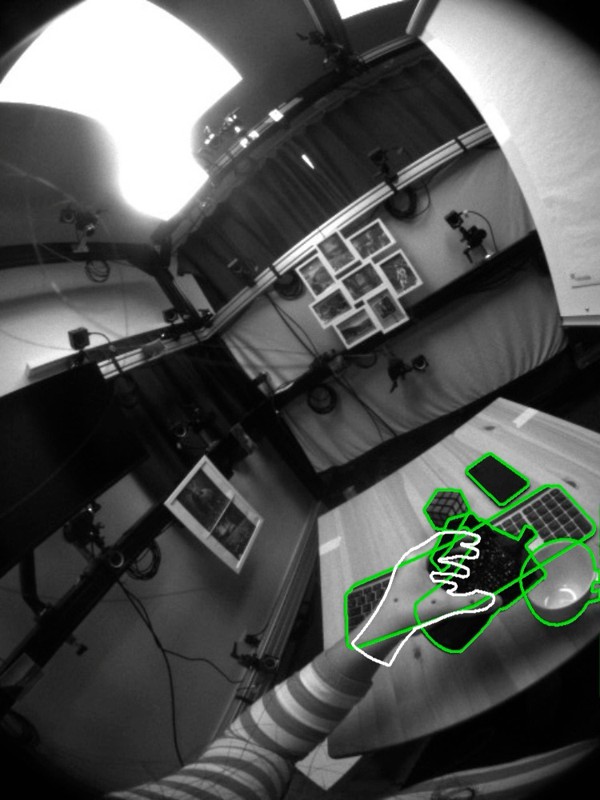}
\includegraphics[width=0.49\linewidth]{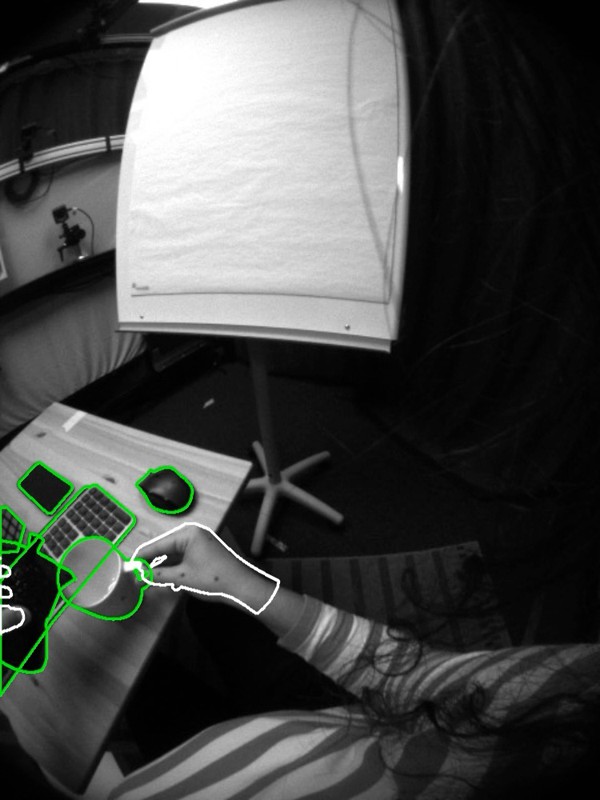}%
\end{minipage}\hfill%
\begin{minipage}{0.195\linewidth}%
\includegraphics[width=\linewidth]{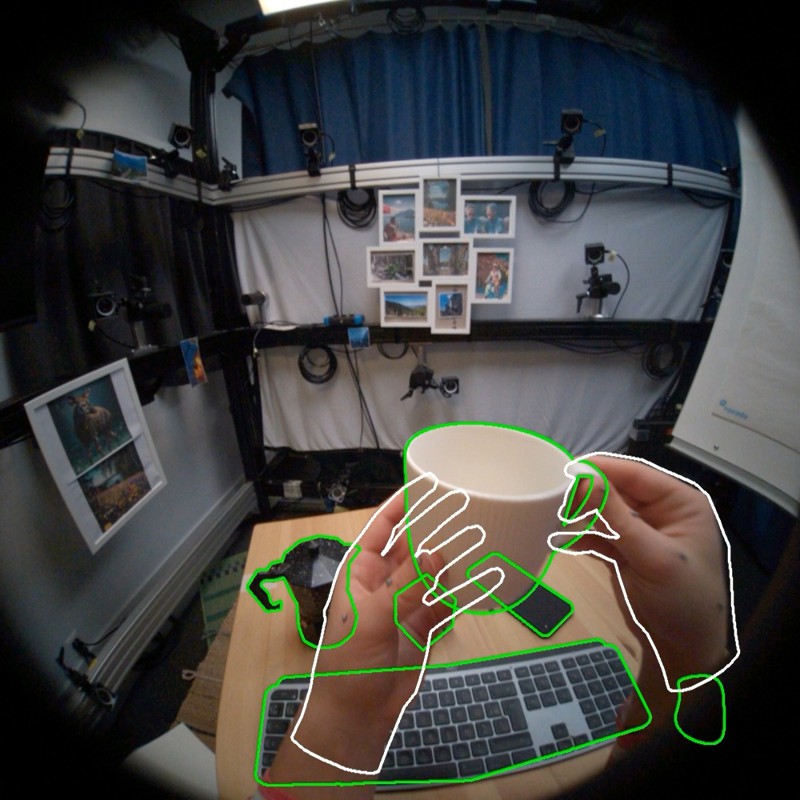}\\[0.2mm]
\includegraphics[width=0.49\linewidth]{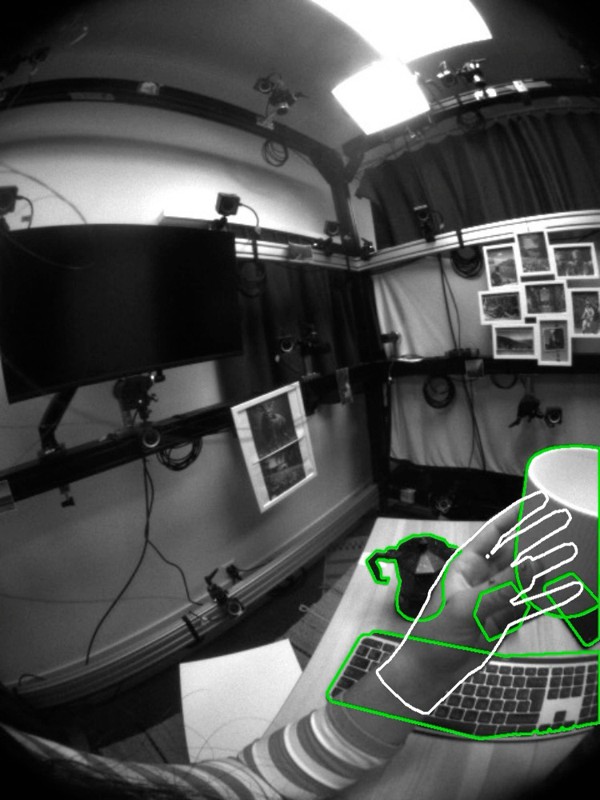}
\includegraphics[width=0.49\linewidth]{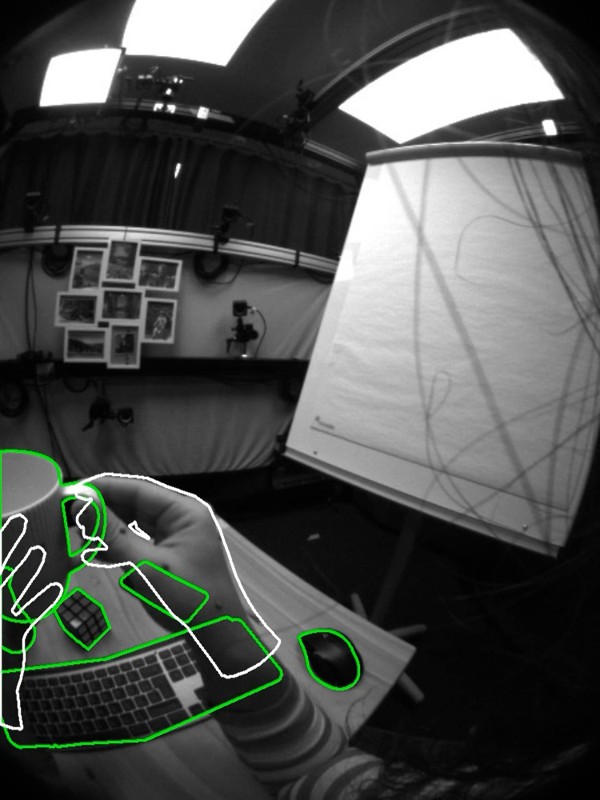}%
\end{minipage}\hfill%
\begin{minipage}{0.393\linewidth}%
{\setlength{\fboxsep}{0pt}\setlength{\fboxrule}{0.5pt}
\fbox{\includegraphics[width=\linewidth]{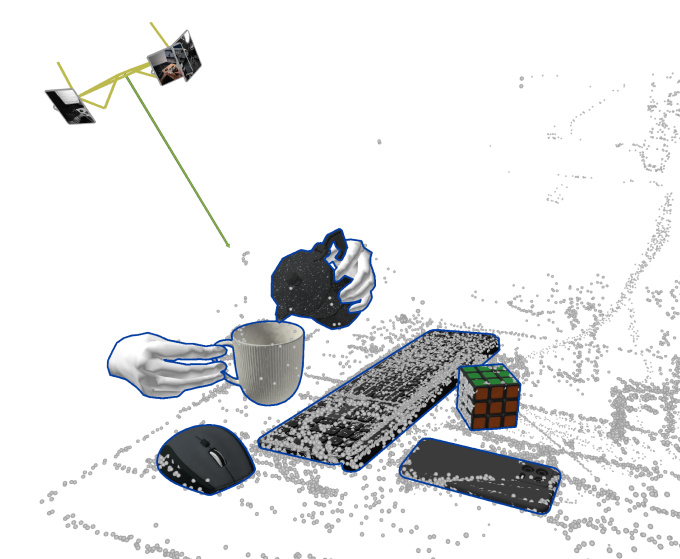}}}%
\end{minipage}
\captionof{figure}{
    \textbf{HOT3D overview.} The dataset includes multi-view egocentric 
    image streams from Aria~\cite{engel2023project} and Quest 3~\cite{Quest3}
    annotated with high-quality ground-truth 3D poses and models of hands and objects.
    Three multi-view frames from Aria are shown on the left, with contours of 3D models of hands and objects in the ground-truth poses in white and green, respectively.
    Aria also provides 3D point clouds from SLAM and eye gaze information (right).
}
\label{fig:overview}
\end{minipage}
\vspace{-2pt}
\end{strip}

\makebox[0pt][c]{%
\hspace{0.925\columnwidth}
\begin{minipage}[b]{\columnwidth}
\vspace{2pt}
\begin{abstract}
\vspace{-5pt}
We introduce HOT3D, a publicly available dataset for egocentric hand and object tracking in 3D. The dataset offers over 833 minutes (3.7M+ images) of recordings that feature 19 subjects interacting with 33 diverse rigid objects. In addition to simple pick-up, observe, and put-down actions, the subjects perform actions typical for a kitchen, office, and living room environment. The recordings include multiple synchronized data streams containing egocentric multi-view RGB/monochrome images, eye gaze signal, scene point clouds, and 3D poses of cameras, hands, and objects. The dataset is recorded with two headsets from Meta: Project Aria, which is a research prototype of AI glasses, and Quest 3, a~virtual-reality headset that has shipped millions of units. Ground-truth poses were obtained by a motion-capture system using small optical markers attached to hands and objects. Hand annotations are provided in the UmeTrack and MANO formats, and objects are represented by 3D meshes with PBR materials obtained by an in-house scanner. In our experiments, we demonstrate the effectiveness of multi-view egocentric data for three popular tasks: 3D hand tracking, model-based 6DoF object pose estimation, and 3D lifting of unknown in-hand objects. The evaluated multi-view methods, whose benchmarking is uniquely enabled by HOT3D, significantly outperform their single-view counterparts.
\end{abstract}%
\end{minipage}
}%

\section{Introduction}

We use our hands to communicate, interact with objects, or utilize objects as tools to act upon
our environment.
The dexterity with which we can manipulate objects is unmatched by other species and has been a key factor in our evolution~\cite{bardo2022precision}.
Hand-object interaction has therefore naturally received considerable attention from various research fields, including computer vision~\cite{oikonomidis2018hands18}.

\begin{figure*}[t]
    \centering
    \setlength{\tabcolsep}{1pt} %
    \renewcommand{\arraystretch}{0.6} %

    \begin{tabular}{cccccc}
        \includegraphics[width=0.163\textwidth]{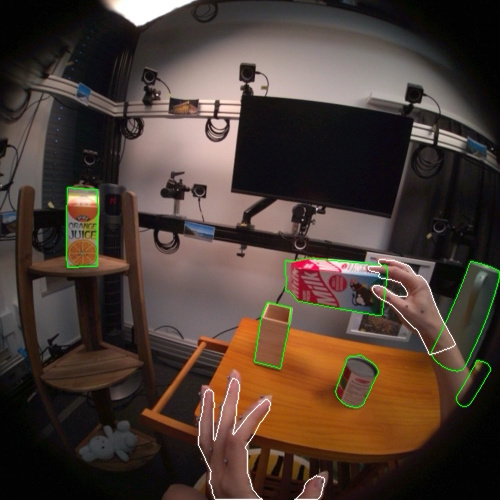} &
        \includegraphics[width=0.163\textwidth]{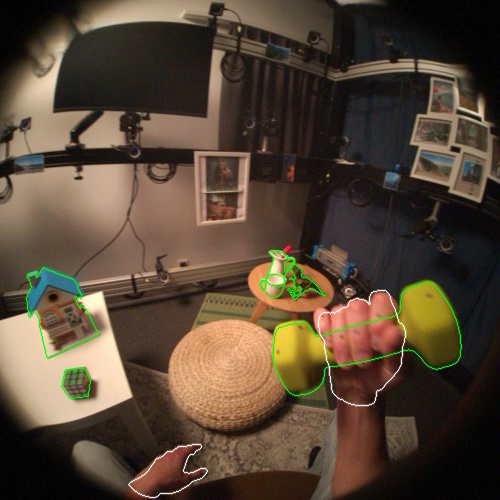} &
        \includegraphics[width=0.163\textwidth]{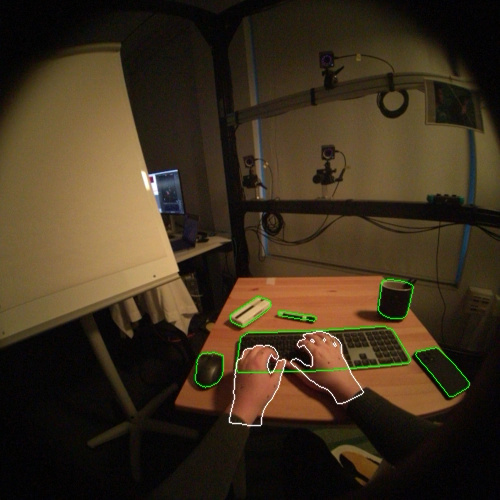} &
        \includegraphics[width=0.163\textwidth]{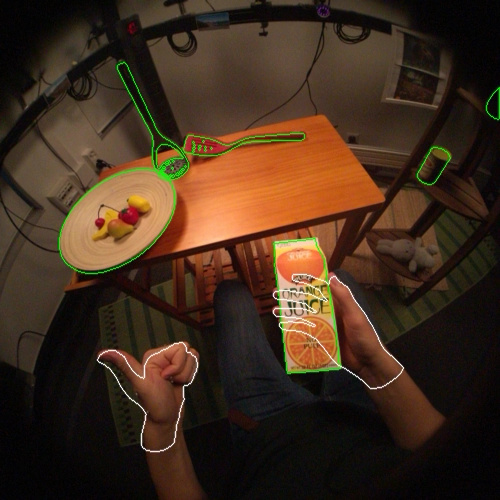} &
        \includegraphics[width=0.163\textwidth]{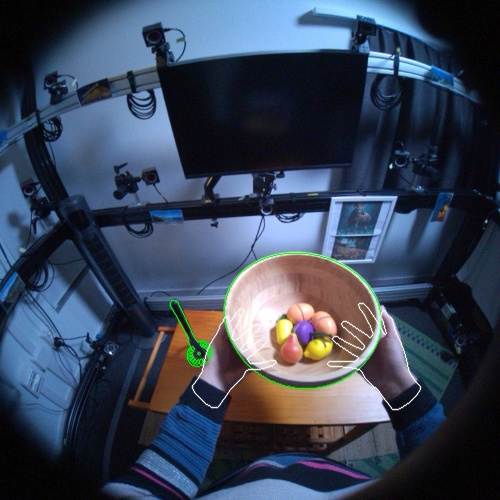} &
        \includegraphics[width=0.163\textwidth]{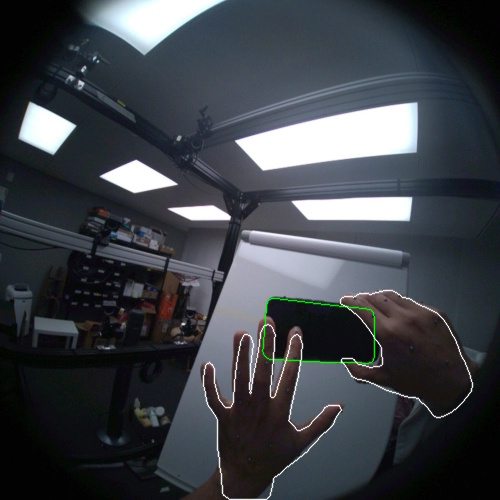} \\
        \includegraphics[width=0.163\textwidth]{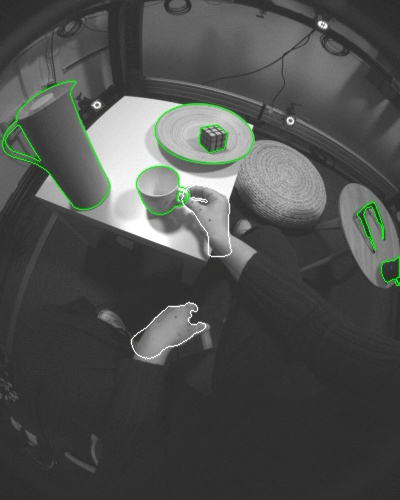} &
        \includegraphics[width=0.163\textwidth]{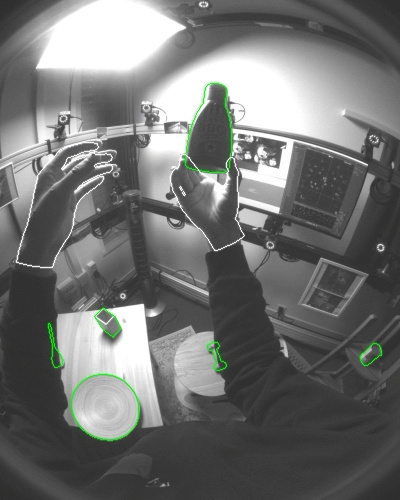} &
        \includegraphics[width=0.163\textwidth]{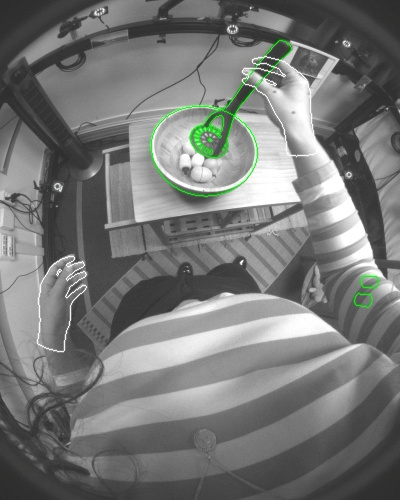} &
        \includegraphics[width=0.163\textwidth]{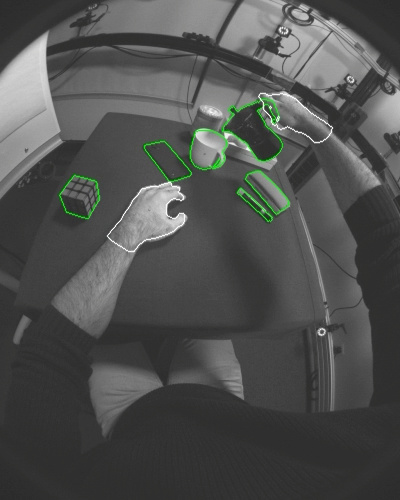} &
        \includegraphics[width=0.163\textwidth]{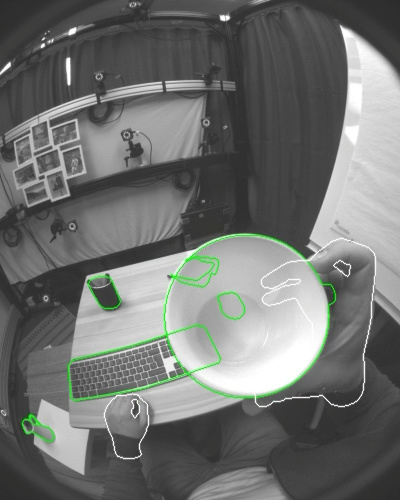} &
        \includegraphics[width=0.163\textwidth]{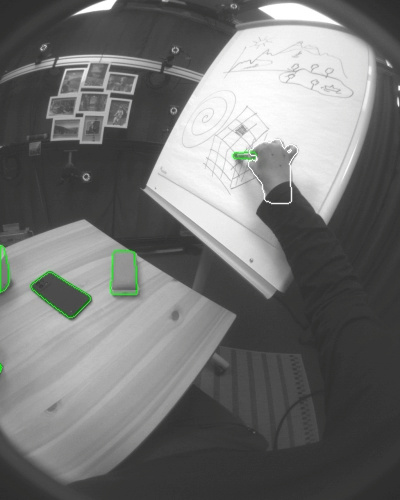}
    \end{tabular}
    \caption{\textbf{Sample images from Aria (top) and Quest 3 (bottom).} Aria recordings include one RGB and two monochrome image streams, while Quest 3 recordings include two monochrome streams (only images from one of the multi-view streams are shown). Contours of 3D models of hands and objects in the ground-truth poses are shown in white and green respectively.
    In addition to simple pick-up/observe/put-down actions, the subjects perform actions that are common in a kitchen, office, and living room.
    To increase diversity, the lighting, furniture, and decorations in the capture lab were regularly randomized.}
    \label{fig:sample_data}
    \vspace{1pt}
\end{figure*}

A vision-based system for automatic understanding of~hand-object interaction, which would be able to capture information about 3D motion, shape and contact of hands and objects, would be useful for a wide range of applications. For instance, such a system could enable transferring manual skills between users
by first capturing expert users performing a sequence of hand-object interactions (while assembling a piece of furniture, doing a tennis serve, \etc), and later using the captured information to guide less experienced users, \eg, via AR glasses. The skills could be similarly transferred from humans to robots.
Such a system could also help AI assistants better understand the context of a user's actions, or enable new input capabilities for AR/VR users. For example, it could turn any physical surface into a virtual keyboard or transform any pencil into a multi-functional magic wand. However, the accuracy and speed of existing methods for understanding hand-object~interaction are not sufficient to reliably support such applications.

To accelerate computer vision research on hand-object~interaction, we are publicly releasing HOT3D, an egocentric dataset
recorded using two recent head-mounted devices from Meta: Project Aria~\cite{engel2023project}, which is a research prototype of light-weight AI glasses, and Quest~3~\cite{Quest3}, a virtual-reality headset that has shipped millions of units.
The dataset offers over 833 minutes of egocentric video streams, including over 1.5M multi-view frames (3.7M+ images) and showing 19 subjects interacting with 33 diverse rigid objects. Besides a simple inspection scenario, where subjects pick up, observe, and put down the objects, the recordings show scenarios resembling typical actions in kitchen, office, and living room spaces. Hands and objects are annotated with accurate 3D poses collected using a passive marker-based motion-capture system. The dataset also includes 3D object models which were obtained by an in-house scanning-based 3D object reconstruction pipeline and include high-resolution geometry and PBR materials~\cite{mcdermott2018pbr}. 
Recordings from Aria additionally include 3D scene point clouds from SLAM and eye gaze signal. Sample images are in Fig.~\ref{fig:sample_data}.

The HOT3D dataset is primarily intended for the training and evaluation of hand and object tracking methods \emph{in 3D space} from \emph{localized, egocentric, multi-view video streams}, as opposed to monocular views or individual images.
Since images from all streams are synchronized with a hardware trigger (\ie, captured at the same timestamp), the dataset enables development of methods that leverage multi-view and/or temporal information.
While providing a testbed for CAD-based object tracking, the dataset also enables a CAD-free tracking setup by providing reference sequences showing different views at each object, which can be used to onboard the objects in a few-shot manner. Furthermore, the dataset can be used for tasks such as 3D object reconstruction and 2D detection or segmentation of hand-object interactions. We also encourage research
that leverages the eye gaze information from Aria, which can enable predicting the user's intent or efficient allocation of the computational budget via foveated sensing.

\begin{table*}[t!]
    \setlength{\tabcolsep}{4pt}
    \footnotesize
    \begin{center}
    \begin{tabularx}{1.0\linewidth}{@{\hspace{1mm}}Y c c c c c c c c c c c}
    \toprule
      Dataset
    & Images  %
    & Channels
    & Ego~/~exo %
    & HW trigger %
    & Headsets
    & Subjects %
    & Hands  %
    & Objects  %
    & ..per scene
    & Gaze %
    & Annotation  %
    \\
    \toprule
    \rowcolor{lightergray} HOT3D (ours) & 3.7M & RGB/gray  & 2--3~/~0 & \cmark & Aria, Quest 3 & 19 & Both & 33 & $\mytilde6$ & \cmark & Mocap \\
    ARTIC~\cite{fan2023arctic} & 2.1M & RGB  & 1~/~8 & \cmark & Helmet & 10 & Both & 11 & 1 & \xmark & Mocap \\
    HOI4D~\cite{liu2022hoi4d} & 2.4M & RGB-D  & 2~/~0 & \xmark & Helmet & 4 & Both & 800 & 1--12 & \xmark & \mbox{RGB-D\,+\,man.} \\
    HO-Cap~\cite{wang2024hocap} & 699K & RGB-D  & 1~/~9 & \xmark & HoloLens & 9 & Both & 64 & 4 & \xmark &  RGB-D\,+\,optim. \\
    DexYCB~\cite{chao2021dexycb} & 582K & RGB-D  & 0~/~8 & \cmark & -- & 10 & Single & 20 & 2--4 & \xmark & Manual \\
    HOGraspNet~\cite{2024graspnet} & 1.5M & RGB-D  & 0~/~4 & \cmark & -- & 99 & Single & 30 & 1 & \xmark & \mbox{RGB-D\,+\,optim.} \\
    H2O-3D\cite{hampali2022keypoint} & 75K & RGB-D  & 0~/~5 & \xmark & -- & 6 & Both & 10 & 1 & \xmark & \mbox{RGB-D\,+\,optim.} \\
    HO-3D~\cite{Hampali2020HO3D}& 78K & RGB  & 0~/~1 & -- & -- & 10 & Single & 10 & 1 & \xmark & \mbox{RGB-D\,+\,optim.} \\
    H2O~\cite{huang2020h2o} & 572K & RGB-D  & 1~/~4 & \cmark & Helmet & 4 & Both & 8 & 1 & \xmark & \mbox{RGB-D\,+\,optim.} \\
    ContactPose~\cite{brahmbhatt2020contactpose} & 2.9M & RGB-D  & 0~/~3 & \xmark & -- & 50 & Both & 25 & 1 & \xmark & RGB-D\,+\,mocap \\
    ObMan~\cite{hasson2019learning} & 147K & Synth.\ RGB  & 0~/~1 & -- & -- & 20 & Single & 2772 & 1 & \xmark & Synthetic \\
    \bottomrule
    \end{tabularx}
    \end{center}
    \caption{\textbf{Datasets with 3D hands and object annotations.} HOT3D is the first dataset to provide multi-view, hardware time-synced, egocentric videos captured with real headsets. It is the largest dataset in terms of total image count and provides high quality ground-truth annotations (comparable to~\cite{fan2023arctic}).}
    \label{tab:datasets_overview}
\end{table*}

In our experiments,
we primarily focus on demonstrating the effectiveness of \emph{multi-view egocentric} data for several popular tasks. This type of data has been largely unexplored despite being commonly available on current AR/VR devices. Our results show that multi-view methods for 3D hand tracking, model-based 6DoF object pose estimation, and 3D lifting of unknown in-hand objects significantly outperform their single-view variants.

\vspace{2.0ex}
\noindent
In summary, we make the following contributions:
\vspace{1.0ex}
\begin{enumerate}
\item \textbf{Publicly available HOT3D dataset:}
We collected and release the first large-scale dataset that offers (1)~multi-view egocentric video streams recorded with real headsets, (2)~high-quality pose annotations of the headset and multiple objects in each scene, as well as pose and shape annotations of both hands, (3)~non-trivial hand-object interaction scenarios with dynamic grasps, and (4)~3D object models with materials for physically based rendering.
HOT3D enables benchmarking methods for various 2D/3D tasks 
on understanding hand-object interaction.

\item \textbf{Strong baselines for tasks enabled by HOT3D:} We~developed simple yet powerful multi-view baseline methods for two tasks that are relevant for AR/VR and contextual AI applications: model-based 6DoF object pose estimation (extending FoundPose~\cite{ornek2023foundpose}), and 3D lifting of unknown in-hand objects (based on stereo matching of DINOv2~\cite{oquab2023dinov2} features).

\item \textbf{Demonstrated effectiveness of multi-view egocentric data:} Our experiments show that multi-view methods for 3D hand tracking, 6DoF object pose estimation, and 3D lifting of handheld objects clearly outperform their single-view counterparts.
This is an important result for research on power-efficient egocentric vision systems, which can typically afford a multi-view camera setup~\cite{engel2023project} but not, \eg, active depth sensors~\cite{ungureanu2020hololens}.

\end{enumerate}
\vspace{1.0ex}

\section{Related work} \label{sec:related_work}

The progress of research in computer vision has been strongly influenced by benchmark datasets~\cite{scharstein2002taxonomy,everingham2010pascal,russakovsky2015imagenet,geiger2012we,hodan2018bop} which enable comparing methods and understanding their limitations. In this section, we first review existing datasets with either hand or object pose annotations, and then focus on datasets that offer annotations of hands and hand-manipulated objects.

\customparagraph{Datasets with hands only.}
Vision-based 3D hand pose estimation and tracking has been
studied for many years, with the first methods focusing on custom datasets with monochrome images~\cite{rehg1994visual,heap1996towards}. Significant improvements
were later achieved on RGB-D images from datasets such as NYU~\cite{tompson2014real}, ICVL~\cite{tang2014latent}, MSRA~\cite{sun2015cascaded}, Tzionas~\etal \cite{tzionas2016capturing}, EgoDexter~\cite{mueller2017real}, or HANDS17~\cite{yuan20172017}. Recently, partly motivated by AR/VR use cases where depth sensors are often unavailable due to high power consumption, the research community has largely switched to RGB or monochrome images, working on datasets such as Stereo~\cite{zhang20163d}, InterHand2.6M~\cite{moon2020interhand2}, FreiHAND~\cite{zimmermann2019freihand}, UmeTrack~\cite{han2022umetrack}, AssemblyHands~\cite{ohkawa2023assemblyhands}, and datasets with pose annotations of both hands and objects reviewed below.

\customparagraph{Datasets with objects only.}
Research on 6DoF object pose estimation and tracking has followed a similar path, starting off with custom monochrome datasets~\cite{roberts1963machine,murase1995visual} and later largely switching to RGB-D datasets such as LM~\cite{hinterstoisser2013model}, YCB-V~\cite{xiang2017posecnn}, and T-LESS~\cite{hodan2017tless},
which are included in the BOP benchmark~\cite{hodan2018bop,hodan2020bop,sundermeyer2022bop,hodan2023bop}. The benchmark currently includes twelve datasets in a unified format, offering 3D object models and training and test RGB-D images annotated with 6DoF object poses. The 3D object models are created manually or using KinectFusion-like systems~\cite{newcombe2011kinectfusion} for 3D surface reconstruction.
The training images are real or synthetic (photo-realistically rendered with BlenderProc~\cite{denninger2019blenderproc,denninger2020blenderproc}) and all test images are real. Besides these instance-level datasets, the community also uses category-level RGB-D datasets such as Wild6D~\cite{fu2022category}, HouseCat6D~\cite{jung2022housecat6d}, and PhoCal\cite{wang2022phocal}. Recent methods started to focus again on estimating object pose from RGB-only images, using datasets such as OnePose~\cite{sun2022onepose} and HANDAL~\cite{guo2023handal}.

\customparagraph{Datasets with hands and objects.}
Many existing datasets include images of hands and objects (\eg, \cite{Fathi2011learning,Bullock2015yale,Bambach2015lending,shan2020understanding,zhang2022fine,Damen2022RESCALING,ragusa2023enigma51}),
but only provide annotations in the form of 2D bounding boxes, segmentation masks, or action labels. Some datasets for 3D hand pose estimation (\eg, \cite{mueller2017real,zimmermann2019freihand,ohkawa2023assemblyhands}) include images of hands interacting with objects, but do not provide 6DoF object pose annotations.

The first dataset with ground-truth poses of both hands and objects was created by Sridhar~\etal \cite{RealtimeHO_ECCV2016} and offers 3014 exocentric RGB-D images of a hand manipulating a cube, manually annotated with fingertip positions and 6DoF poses of the cube. To avoid the manual annotation, which is tedious and not scalable, the FHPA dataset~\cite{Garcia2018first} used magnetic sensors attached to one hand and objects, noticeably affecting their appearance. This dataset includes 105K egocentric RGB-D images with ground-truth poses of a single hand and 4 objects.
The ObMan dataset~\cite{hasson2019learning} resorted to synthesizing images of hands grasping objects, with the grasps generated by an algorithm from robotics.

HO-3D~\cite{Hampali2020HO3D} was the first dataset with real images annotated by an optimization procedure that leverages multi-view RGB-D image streams and is almost fully automatic. The dataset offers 78K images from several exocentric cameras, showing 10 subjects and 10 objects. A similar annotation procedure was used for several subsequent datasets~\cite{huang2020h2o,hampali2022keypoint,chao2021dexycb,liu2022hoi4d, bhatnagar22behave}.
H2O~\cite{huang2020h2o} includes 572K egocentric multi-view RGB-D images of 4 subjects manipulating 8 objects.
H2O-3D\cite{hampali2022keypoint} provides 75K exocentric RGB images of 6 subjects manipulating 10 YCB objects~\cite{calli2015ycb}. 
DexYCB~\cite{chao2021dexycb} consists of 1000 clips of 3 seconds with the total of 582K RGB-D images,
recorded from 8 exocentric views and showing 10 subjects picking up 20 YCB objects with near-static grasps. HOI4D~\cite{liu2022hoi4d} includes 2.4M egocentric RGB-D images from over 4000 video sequences showing 9 subjects interacting with 800 different objects from 16 categories in 610 different indoor environments.
Besides rigid objects, this dataset contains articulated objects, but focuses on simpler scenarios with a single hand and a single object, and only includes single-view video sequences.
An RGB-D optimization procedure was also used in ContactPose~\cite{brahmbhatt2020contactpose} along with the information from thermal cameras for accurately annotating hand poses, while the object poses were annotated using optical markers. ContactPose includes 2.9M RGB-D images of 50 subjects grasping 25 household objects, however, the grasps are static, background green and all objects are blue (3D printed), which makes the images less realistic.

Similar to HOT3D, a marker-based motion-capture system was used to collect ground-truth poses of hands and objects in the recent ARCTIC dataset~\cite{fan2023arctic}. This dataset includes 2.1M RGB images showing 10 subjects interacting with 11 articulated objects. The images were captured at 233K timestamps from 9 views, only one of which is egocentric -- recorded with a mock-up of an egocentric device (a camera mounted on a helmet).

Besides HOT3D, HO-Cap~\cite{wang2024hocap} is the only other dataset that is recorded with a real headset (Microsoft HoloLens) and provides 3D annotations of both hands and multiple objects, although the annotations were obtained using RGB-D cameras (instead of a precise motion-capture lab) and are therefore of a lower accuracy.

\section{HOT3D dataset} \label{sec:dataset}

\noindent\textbf{833 minutes of recordings.}
HOT3D includes egocentric,~multi-view, synchronized data streams recorded with
Aria~\cite{engel2023project} and Quest 3~\cite{Quest3}. Image streams are recorded at 30 fps and contain 1.5M+ multi-view frames consisting of 3.7M+ images. Each Aria frame consists of one RGB 1408$\times$1408 image and two monochrome 640$\times$480 images. Each Quest~3 frame consists of two monochrome 1280$\times$1024 images. Intrinsic
parameters and camera-to-world transformations are available for all images. Aria recordings also include 3D scene point clouds (from SLAM) and eye gaze signal.
See appendix for details.

\begin{figure}[t!]
  \centering
  \includegraphics[width=0.86\linewidth]{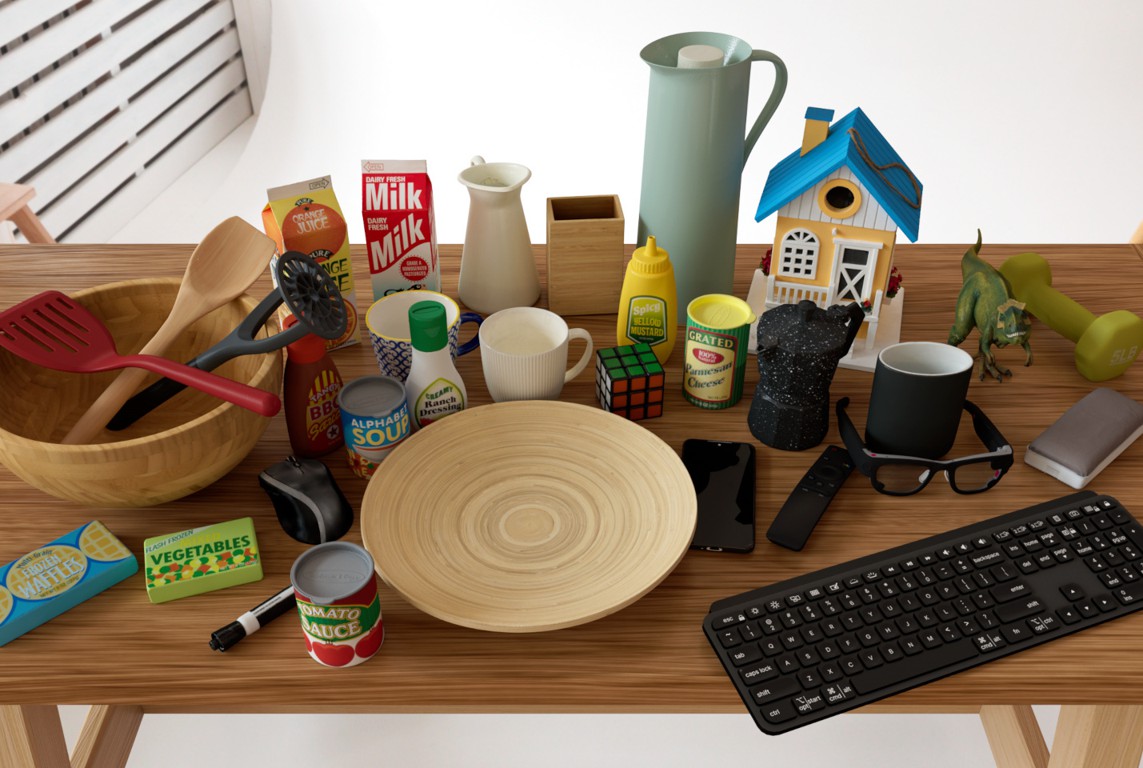}

   \caption{\textbf{High-quality 3D mesh models.} This image shows a rendering of the 33 object models, demonstrating their quality. The models were obtained by an in-house scanning-based 3D object reconstruction pipeline and include PBR materials, which enable rendering of photo-realistic training images for methods that require it. The collection includes household and office objects of diverse appearance, size, and affordances.
   }
   \label{fig:models}
   \vspace{-5pt}
\end{figure}

\customparagraph{3D mesh models of 33 objects.} The models were obtained using an in-house scanning-based 3D object reconstruction pipeline, which provides high-resolution geometry and
PBR~\cite{mcdermott2018pbr} materials.
These materials include metallic, roughness, and normal maps and enable rendering of photo-realistic training images~\cite{hodan2019photorealistic,hodan2020bop}. The object collection includes household and office objects of diverse appearance, size, and affordances (Fig.~\ref{fig:models}).

\customparagraph{19 diverse subjects.} To ensure diversity, we recruited 19~participants with different hand appearances and shapes. Hands of each participant were scanned by a custom 3D hand scanner and are provided in the UmeTrack~\cite{han2022umetrack} and MANO~\cite{mano:romero:tog17} formats.

\customparagraph{4 everyday scenarios.}
In addition to a simple inspection scenario, where subjects pick up, observe, and put down objects, subjects were asked to perform typical actions in a kitchen, office, and living room. All scenarios were captured in the same lab equipped with scenario-specific furniture. In each recording, subjects were asked to interact with  up to 6 objects. To enhance diversity within the dataset, we regularly randomized various aspects such as lighting, 
furniture placement, and decorative elements. The resulting dataset consists of 425 recordings, with 198 captured with Aria and 226 with Quest 3. Each recording is around 2 minutes long.

\begin{figure}[t!]
  \centering
  \includegraphics[width=0.86\linewidth]{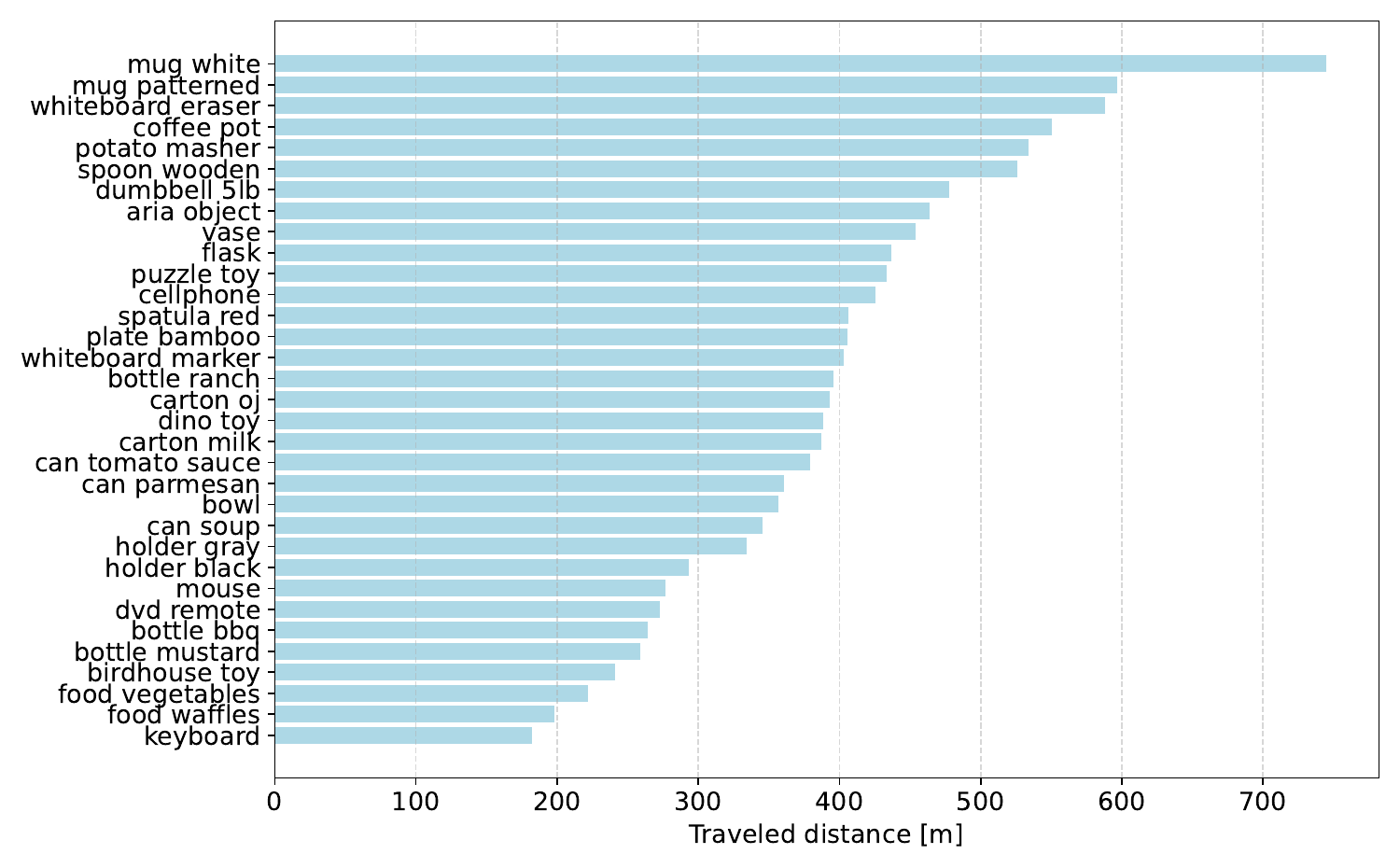}
  \vspace{-2pt}
   \caption{\textbf{Distances traveled by HOT3D objects.} In total, subjects moved the 33 objects over 13\,km. While objects like the keyboard and waffles were mostly resting, the white mug is a true explorer.
   }
   \label{fig:traveled_distances}
\end{figure}

\customparagraph{Ground-truth annotations.} Recordings are annotated with per-frame ground-truth poses of hands and objects obtained in a motion-capture lab shown in Fig.~\ref{fig:rig} and described in the supplement.
Object and wrist poses are represented as 3D rigid transformations from the 3D model space to the scene space, and hand poses are represented in the UmeTrack~\cite{han2022umetrack} and MANO~\cite{mano:romero:tog17} formats (UmeTrack is more accurate while MANO is more standard).
Annotations in some frames may be missing or be of a lower quality.
Out of 1.5M frames included in the dataset, 1.16M frames are fully annotated (\ie, ground-truth poses of all hands and objects are available) and passed our visual inspection (manually flagging frames were rendering of hand and object models in the ground-truth poses is not closely aligned with the observed image).
We release all 1.5M frames, which may be useful for unsupervised training, and provide a mask of the valid 1.16M frames.
See Fig.~\ref{fig:traveled_distances}
and the appendix for statistics of the ground-truth object poses.

\customparagraph{Training and test splits.} The training split of HOT3D includes recordings of 13 subjects (1M multi-view frames), and the test split includes recordings of the remaining 6 subjects (0.5M multi-view frames). Ground-truth pose annotations are publicly released only for the training split. Ground-truth annotations for the test split are accessible only by dedicated evaluation servers.

\customparagraph{Curated clips (HOT3D-Clips).} To facilitate benchmarking of various tracking and pose estimation methods, we also release 3832 curated clips extracted from the full recordings (2804 clips come from the training and 1028 from the test split; 1983 from Aria and 1849 from Quest 3). Each clip has 150 frames (5 seconds) which all passed several quality-assurance tests (verifying that all hand and object annotations are present, at least one hand and one object are visible, discarding overexposed frames) and our visual inspection mentioned earlier.

\customparagraph{Object onboarding sequences.} 
To enable benchmarking~model-free object tracking methods~\cite{sun2022onepose}, which learn new objects from reference images, and 3D object reconstruction methods~\cite{mildenhall2021nerf}, HOT3D includes two types of onboarding sequences which show all possible views of each object: (1) sequences showing a static object on a desk, when the object is standing upright and upside-down, and (2) sequences showing an object manipulated by hands.
The static onboarding setup is suitable for NeRF-like reconstruction methods~\cite{mildenhall2021nerf}, while the latter is more practical for AR/VR applications yet more challenging~\cite{hampali2023hand}.
The ground-truth object poses are provided for all frames of the static sequences, but only for the first frame of the dynamic sequences. This is to simulate real-world settings, where the poses can be easily obtained by SfM~\cite{schonberger2016structure} in the static setup, but are challenging to obtain in the dynamic setup.
The ground-truth pose for the first frame of dynamic sequences is provided to define the canonical object space, which is necessary for evaluating 6DoF object tracking.

\begin{figure}[t!]
  \centering
   \includegraphics[width=0.80\linewidth]{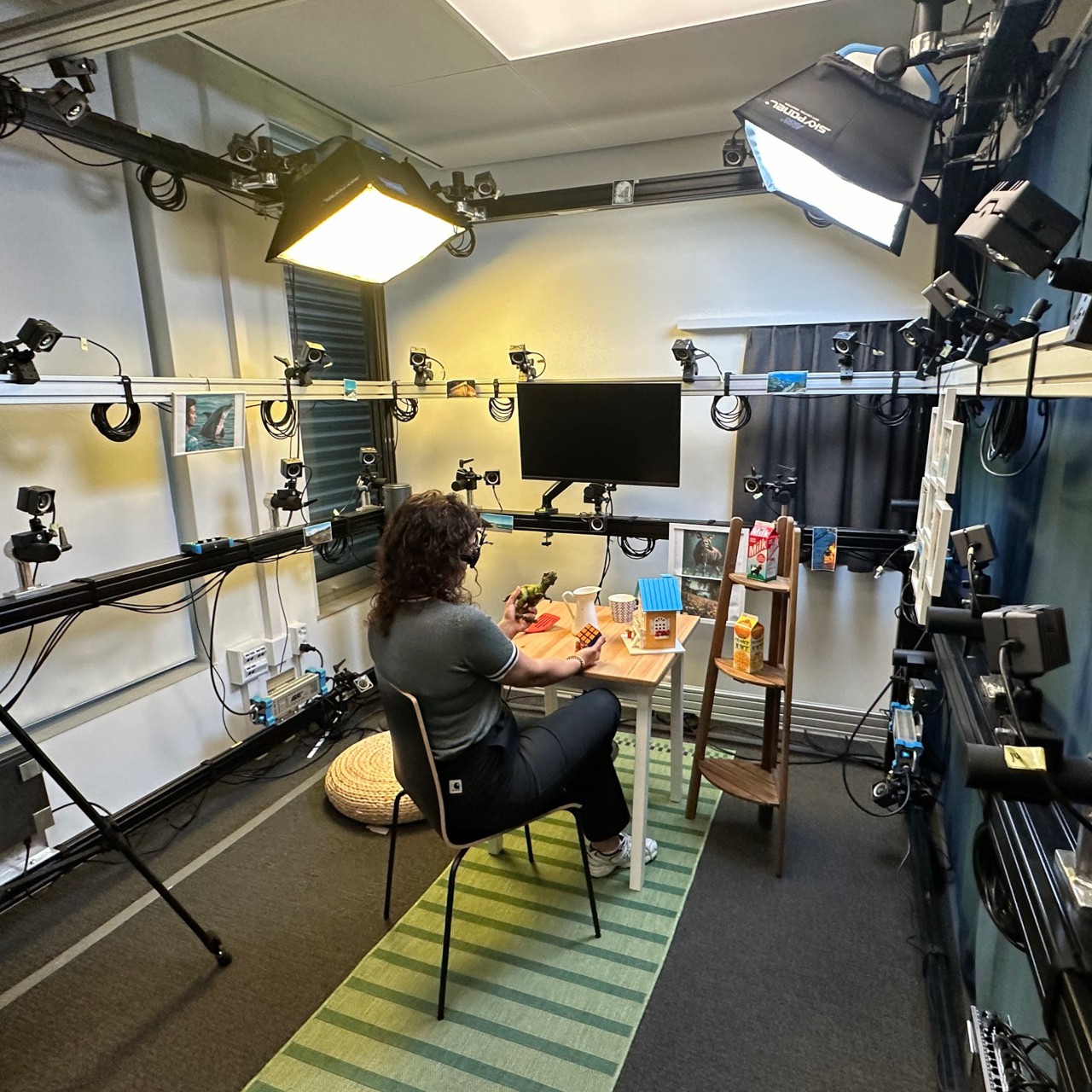}
   \caption{
   \textbf{Motion-capture lab.} 
   The HOT3D dataset was collected using a motion-capture rig equipped with a few dozens of infrared exocentric OptiTrack cameras and light diffuser panels for illumination variability.
   }
   \label{fig:rig}
\end{figure}

\section{Experiments} \label{sec:experiments}
Wearable headsets often feature multiple cameras, which makes them naturally suited for developing multi-view 3D vision methods.
In this section, we demonstrate that multi-view methods outperform single-view methods for several popular egocentric tasks. First, we compare the single-view and multi-view versions of the UmeTrack~\cite{han2022umetrack} method for 3D hand tracking. Second, we extend the FoundPose~\cite{ornek2023foundpose} method for model-based 6DoF object pose estimation to multiple views and evaluate against the original version. Third, we develop a method for 3D lifting of unknown in-hand objects by stereo matching of DINOv2~\cite{oquab2023dinov2} features, which we compare against a single-view method similar to OSNOM~\cite{plizzari2024spatial}.
Note that all 3D predictions in our experiments were expressed in the headset space but could be transformed to the world space as the camera-to-world transformation is available (from on-headset SLAM).

\subsection{3D hand pose tracking}

\noindent\textbf{Experimental setup.} Given the hand shape (3D hand skeleton in the canonical pose) and ground-truth 2D bounding boxes of visible hands, the task is to estimate the hand poses (3D locations of skeleton joints) in every frame of an input sequence.
We train the UmeTrack~\cite{han2022umetrack} hand tracker on three variants of training data: (1)~training sequences from the UmeTrack dataset, (2)~HOT3D training sequences recorded with Quest~3, and (3)~the combination of the two. The UmeTrack dataset was recorded with the Quest 2 headset, which has the same but differently arranged cameras compared to Quest 3, and includes 1397 real and 1397 synthetic sequences, each recorded at 30\,fps for 15 seconds. The sequences depict single-hand motions and hand-hand interactions performed by 53 participants, but do not include any hand-object interactions.
All three UmeTrack models were trained on two-view image streams with one of the views randomly masked out (as in~\cite{han2022umetrack}). The masking increases the tracking robustness and encourages the models not to rely on both views, which enables a fair comparison of their single- and two-view modes. 
We evaluate the models on all frames of test UmeTrack sequences and all frames of test HOT3D clips from Quest 3. The accuracy of the predicted 3D locations is measured by the Mean Keypoint Position Error (MPKE)~\cite{han2022umetrack}.

\customparagraph{Results (Tab.~\ref{tab:hand_pose}).} When trained on the UmeTrack dataset, the hand tracker in the single-view mode performs poorly on HOT3D (24.2 MKPE on HOT3D \vs 13.6 on UmeTrack). Similarly, when trained on HOT3D, the single-view tracker performs poorly on UmeTrack (23.7 MKPE on UmeTrack \vs 18.0 on HOT3D). The main reason of these accuracy gaps is that hand-object interactions are present only in HOT3D while hand-hand interactions only in the UmeTrack dataset.
The accuracy drop is even larger when the tracker, still trained only on one of the datasets, is evaluated in the two-view mode. This is because the datasets are recorded with different headsets (Quest 2 \vs Quest 3) and the tracker overfits to the camera configuration seen at training. The domain gap between the two datasets is effectively closed when the tracker is trained on both, achieving 13.4 MKPE on UmeTrack and 15.4 on HOT3D in the single-view mode, and a significant $41\%$ improvement (9.5 MKPE on UmeTrack and 10.9 on HOT3D) in the two-view mode.

\begin{table}[t!]
    \setlength{\tabcolsep}{3.5pt}
    \small
    \begin{center}
    \begin{tabularx}{1.0\linewidth}{c c Y Y}
    \toprule
     & & MKPE~on & MKPE~on \\
    Training dataset & Views & UmeTrack $\downarrow$ & HOT3D $\downarrow$\\
    \toprule
    UmeTrack~\cite{han2022umetrack} & 1 & 13.6 & 24.2 \\
    UmeTrack~\cite{han2022umetrack} & 2 & \phantom{0}9.7 & 25.6 \\
    \midrule
    HOT3D-Quest3 & 1 & 23.7 & 18.0 \\
    HOT3D-Quest3 & 2 & 30.3 & 13.1 \\
    \midrule
    UmeTrack\;+\;HOT3D-Quest3 & 1 & 13.4 & 15.4 \\
    UmeTrack\;+\;HOT3D-Quest3 & 2 & \phantom{0}\textbf{9.5} & \textbf{10.9} \\
    \bottomrule
    \end{tabularx}
    \end{center}
    \caption{\textbf{3D hand pose tracking by UmeTrack~\cite{han2022umetrack}.}
    Reported~is the Mean Keypoint Position Error (MKPE, in mm) achieved on the UmeTrack and HOT3D-Quest3 datasets by single-view and two-view variants of the UmeTrack hand tracker, which was trained on training splits of either of the datasets or on their combination.
    }
    \label{tab:hand_pose}
\end{table}

\begin{figure}[t!]
    \centering
    \footnotesize
    \setlength{\tabcolsep}{1pt} %
    \renewcommand{\arraystretch}{0.6} %
    \begin{tabular}{ccccc}
        Input & GT skeleton & GT mesh & Pred.\ skeleton & Pred.\ mesh \vspace{2pt} \\
        \includegraphics[width=0.19\linewidth]{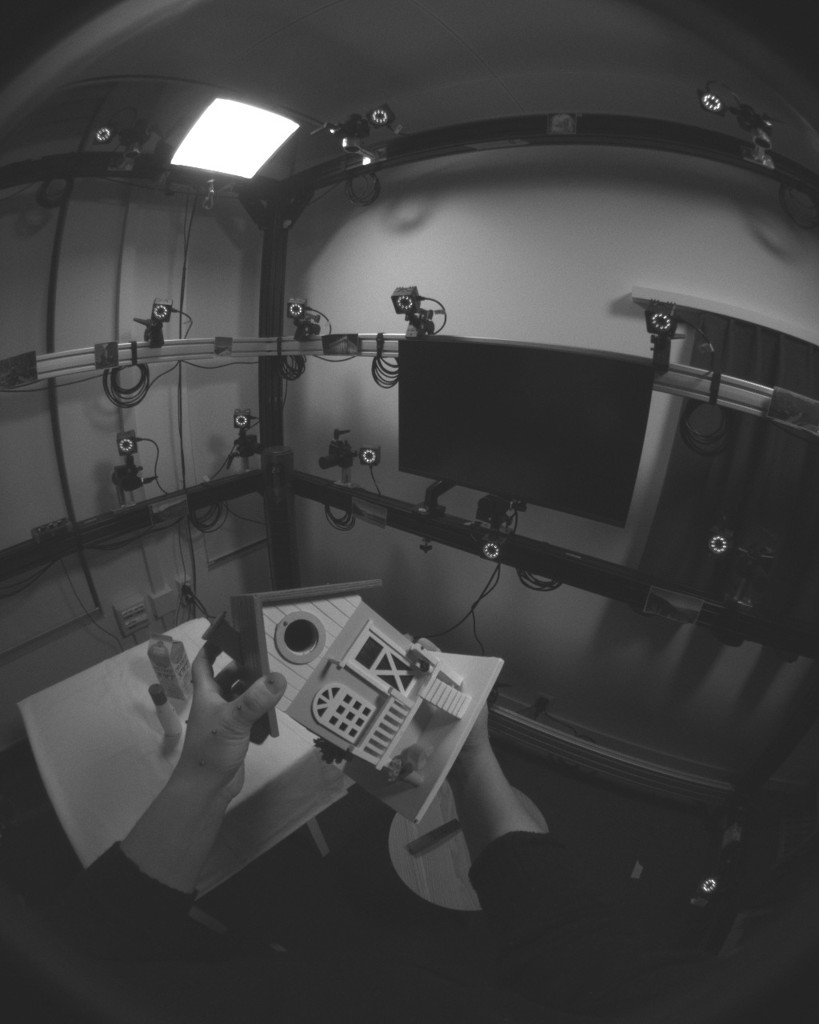} &
        \includegraphics[width=0.19\linewidth]{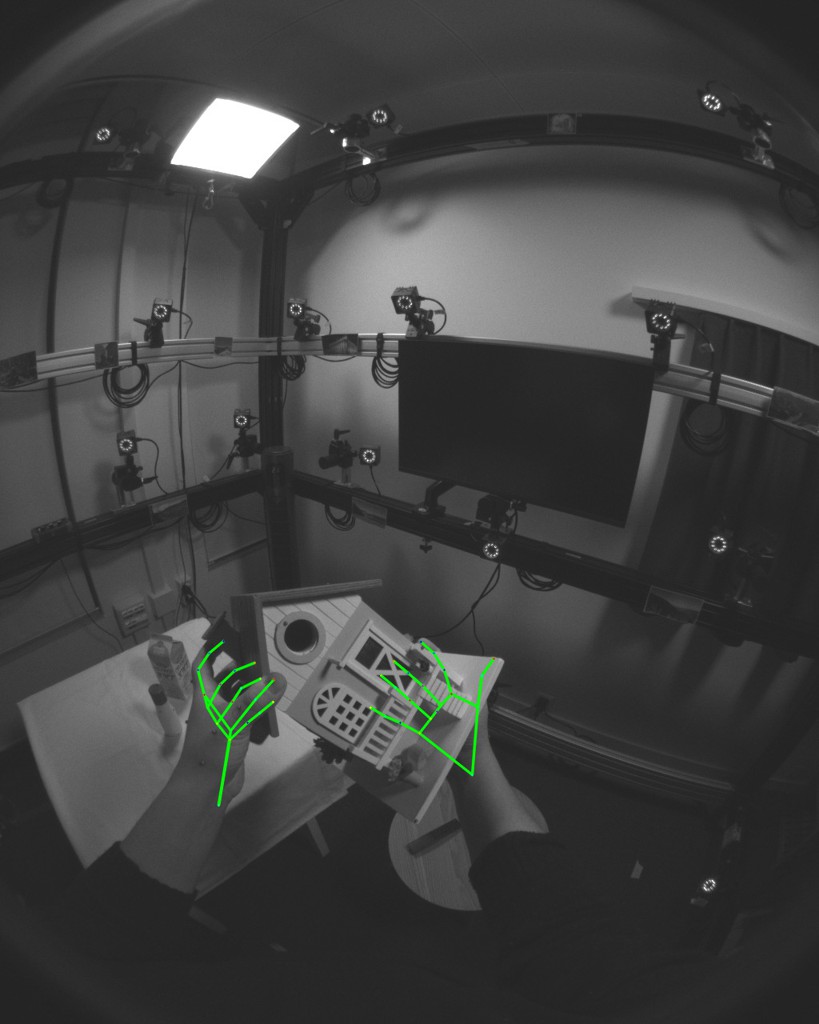} &
        \includegraphics[width=0.19\linewidth]{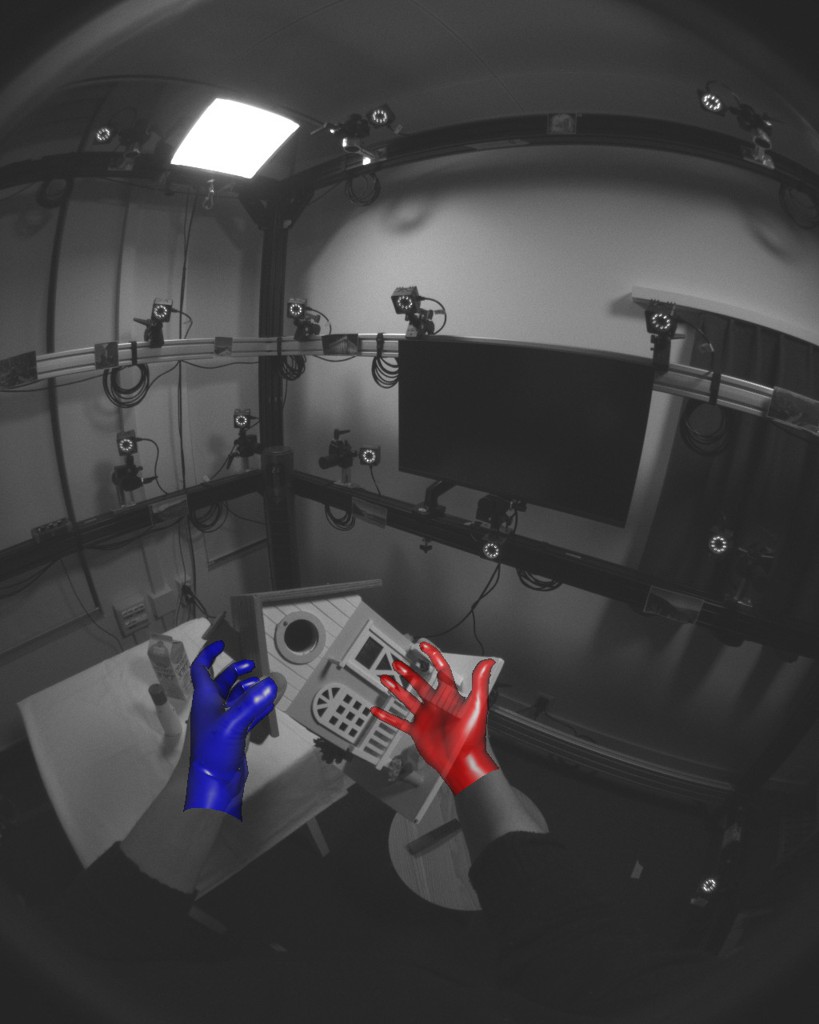} &
        \includegraphics[width=0.19\linewidth]{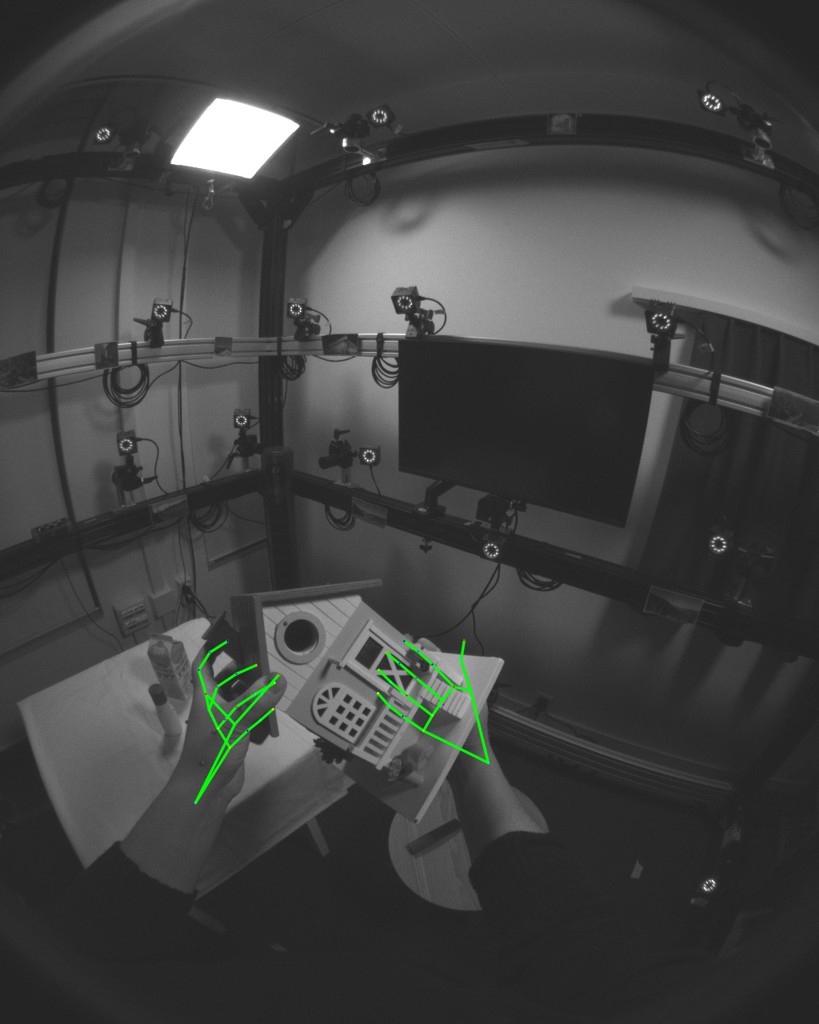} &
        \includegraphics[width=0.19\linewidth]{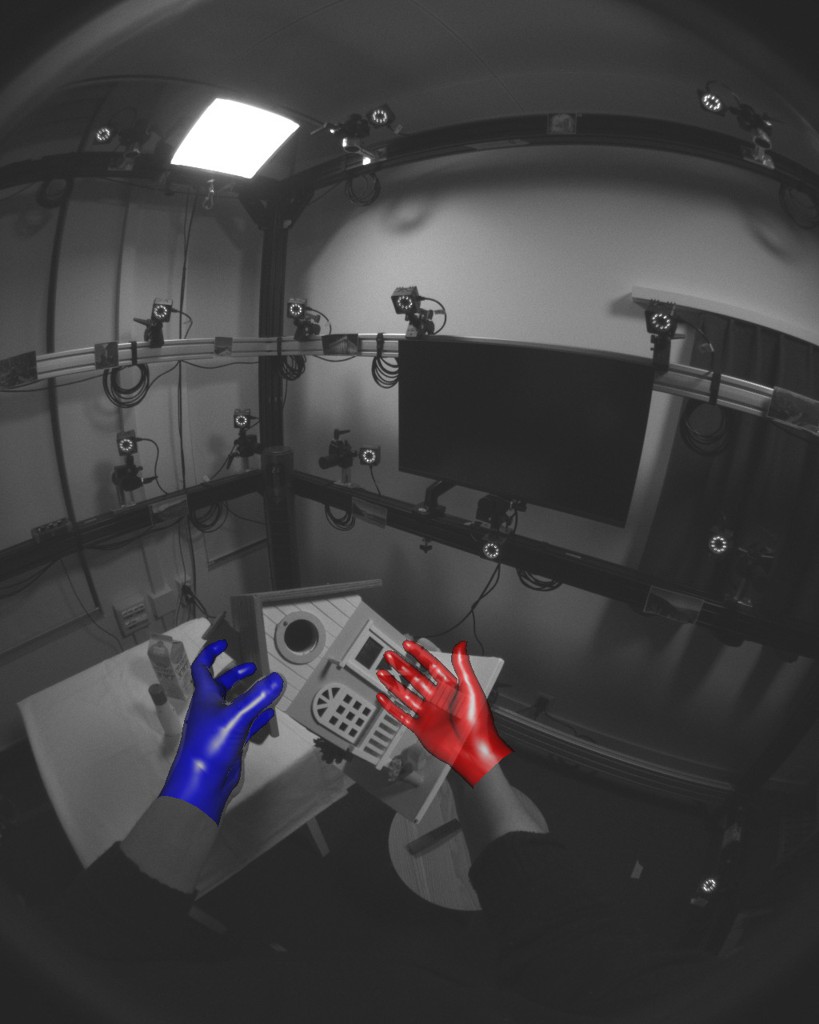} \\
        \includegraphics[width=0.19\linewidth]{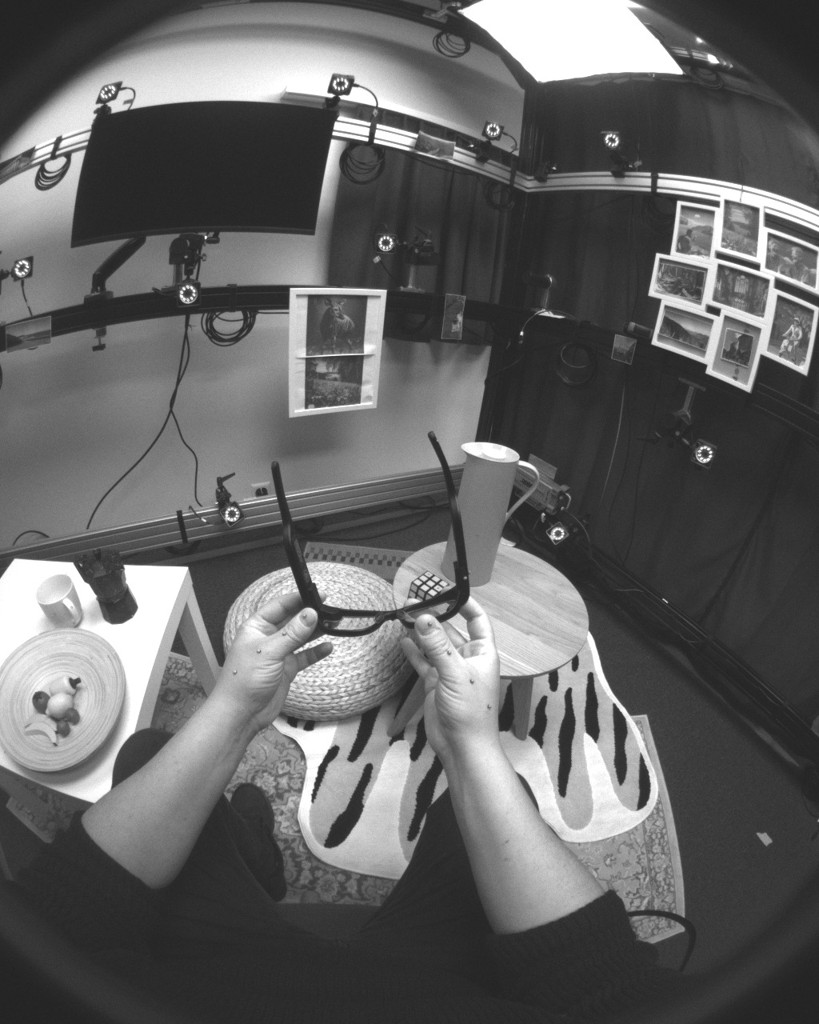} &
        \includegraphics[width=0.19\linewidth]{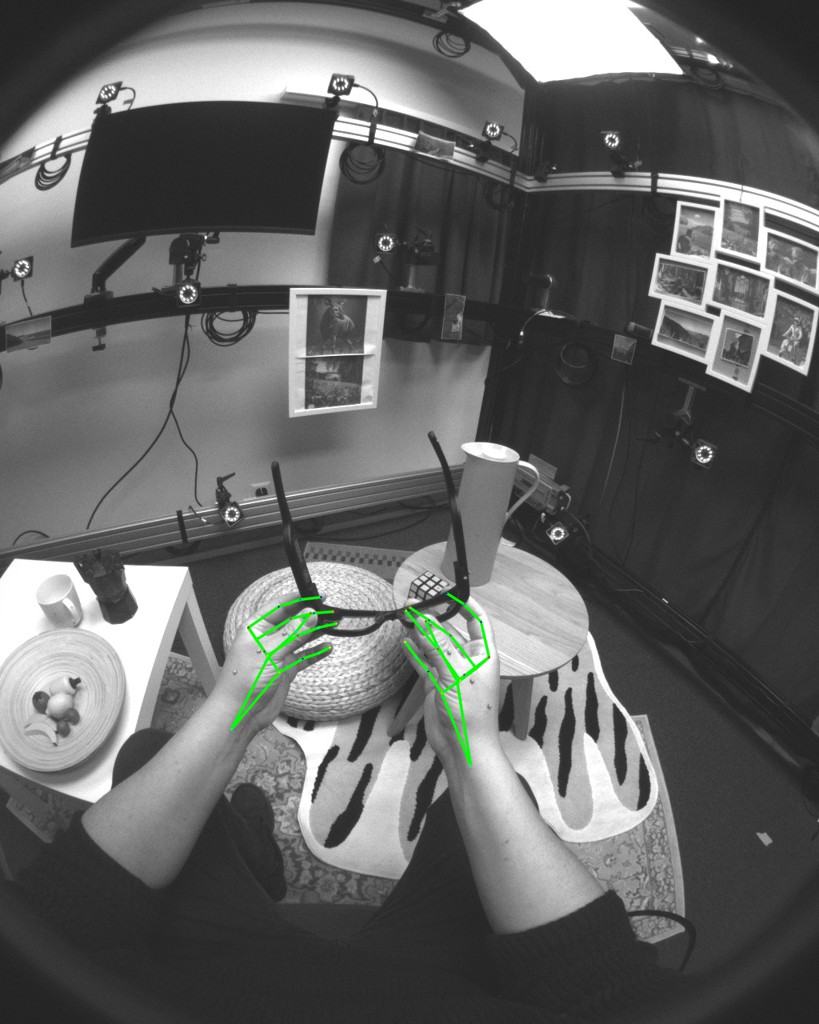} &
        \includegraphics[width=0.19\linewidth]{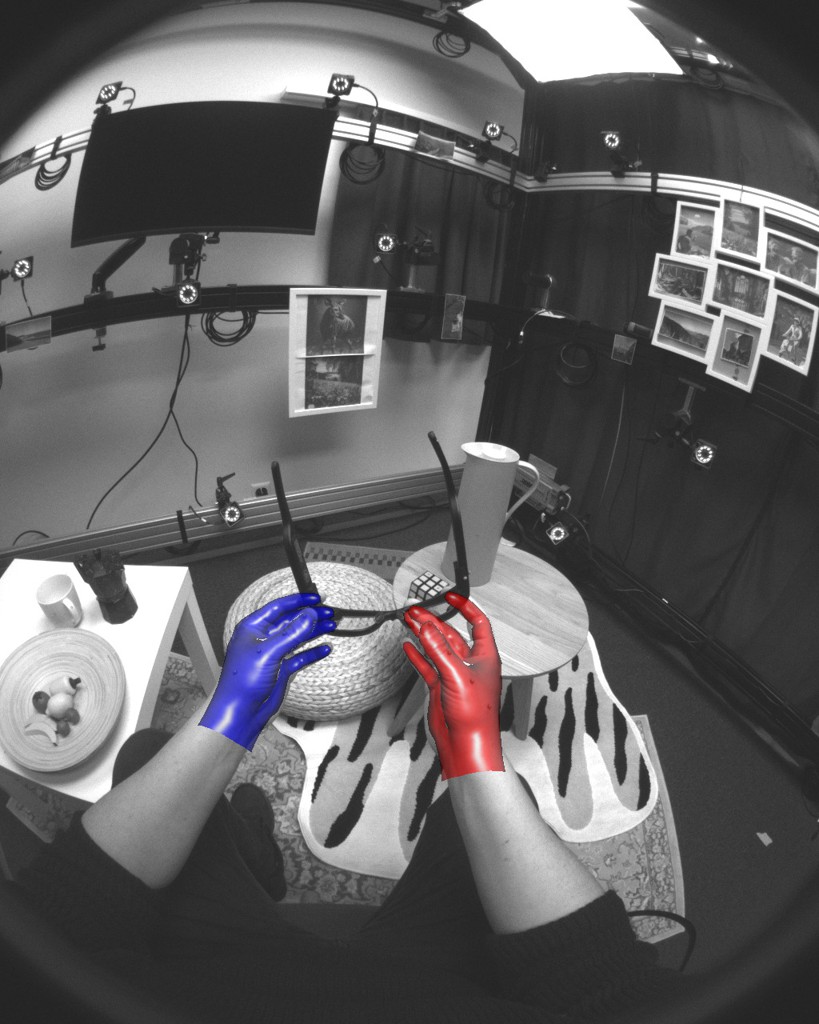} &
        \includegraphics[width=0.19\linewidth]{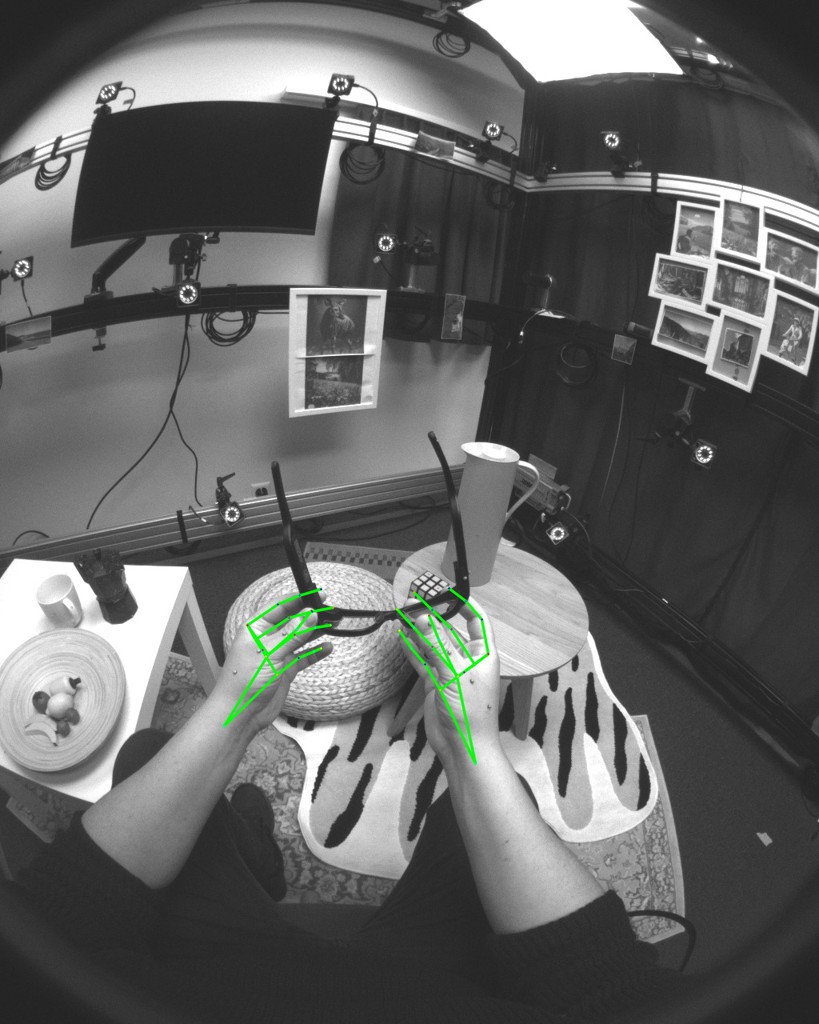} &
        \includegraphics[width=0.19\linewidth]{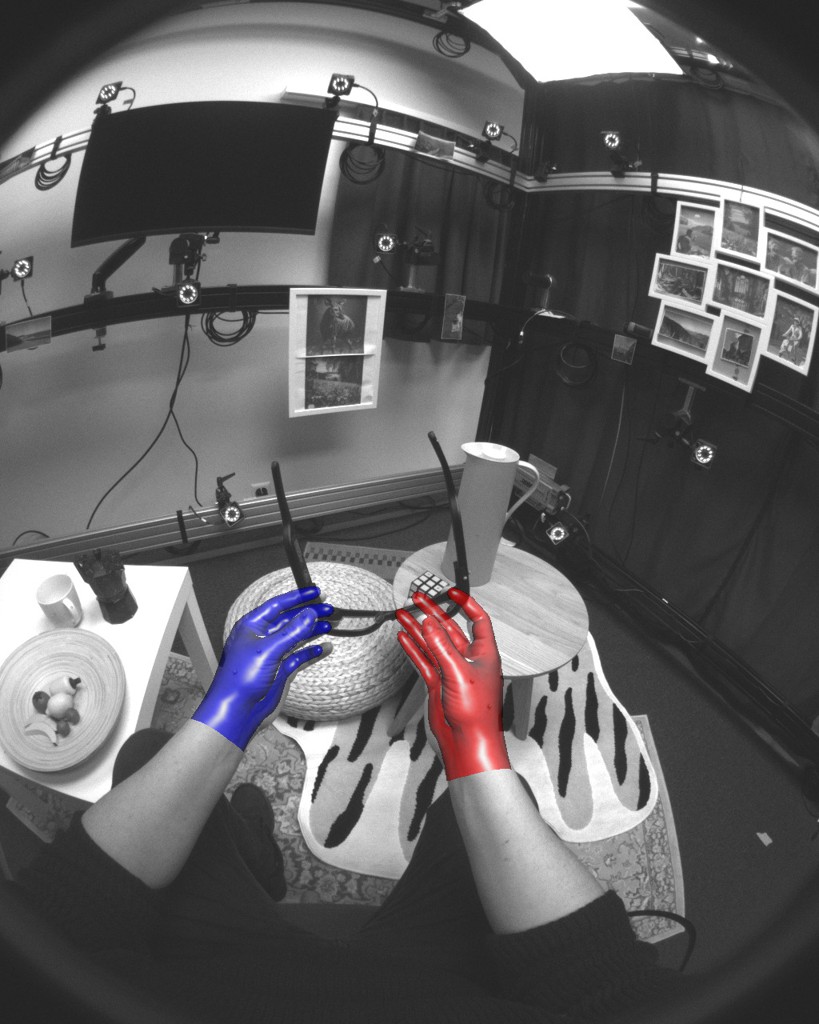} \\
        \includegraphics[width=0.19\linewidth]{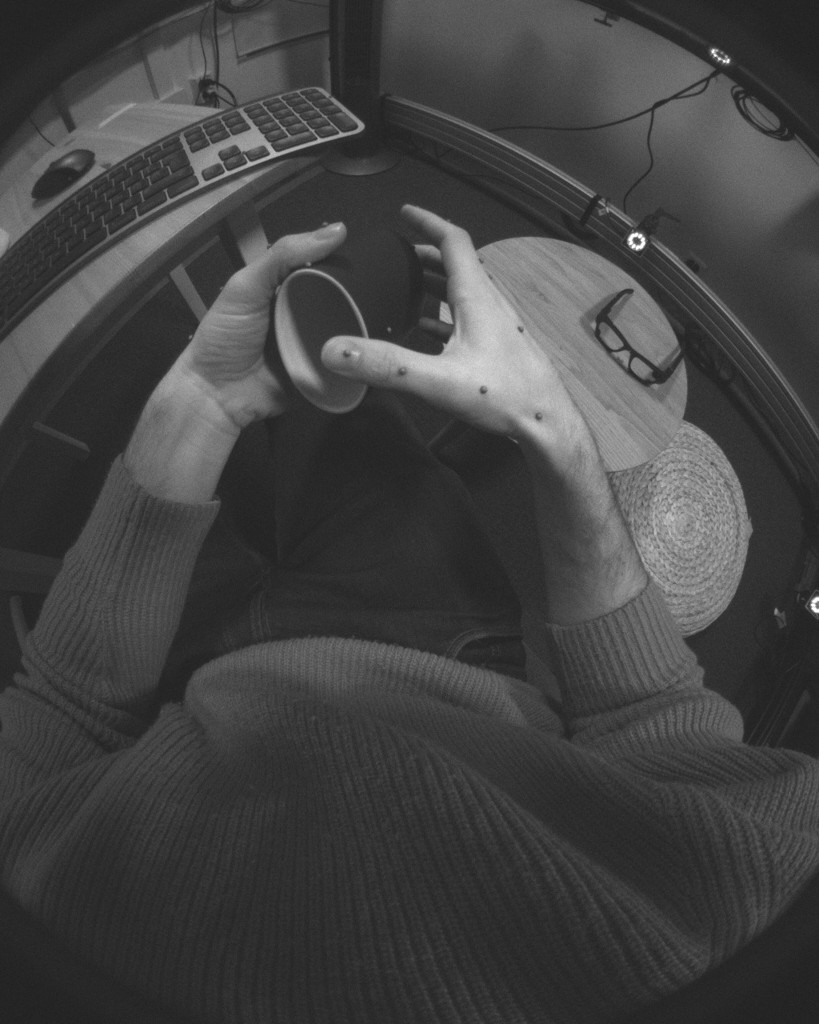} &
        \includegraphics[width=0.19\linewidth]{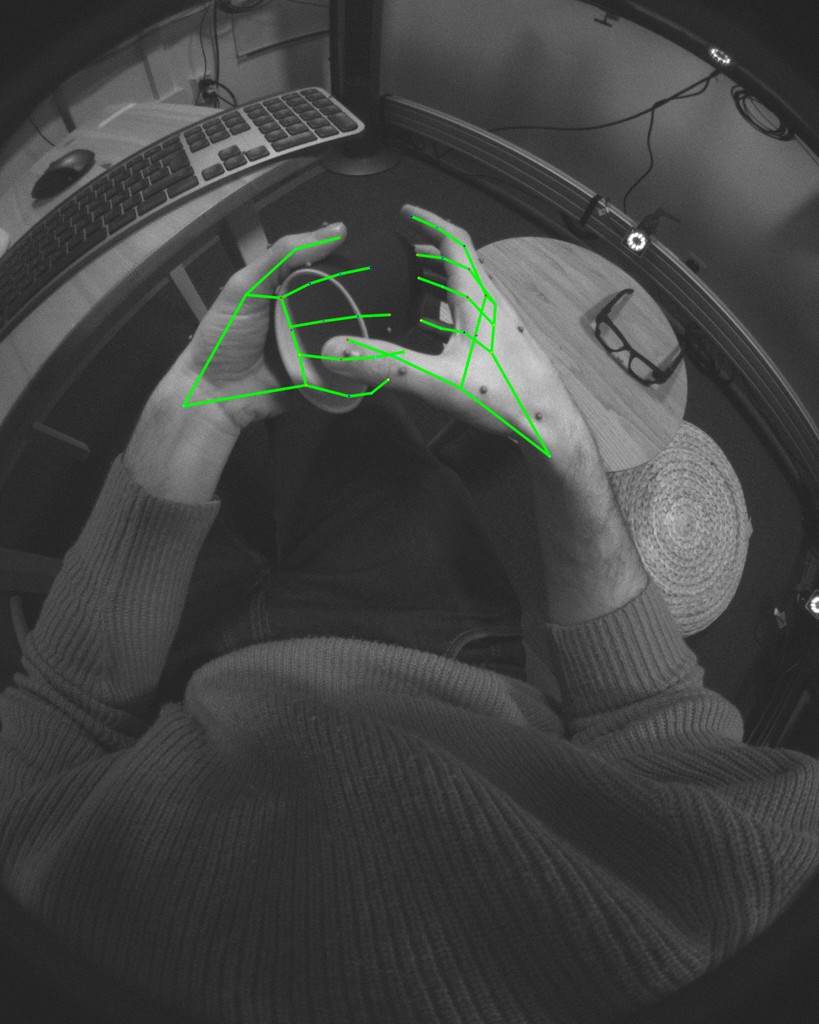} &
        \includegraphics[width=0.19\linewidth]{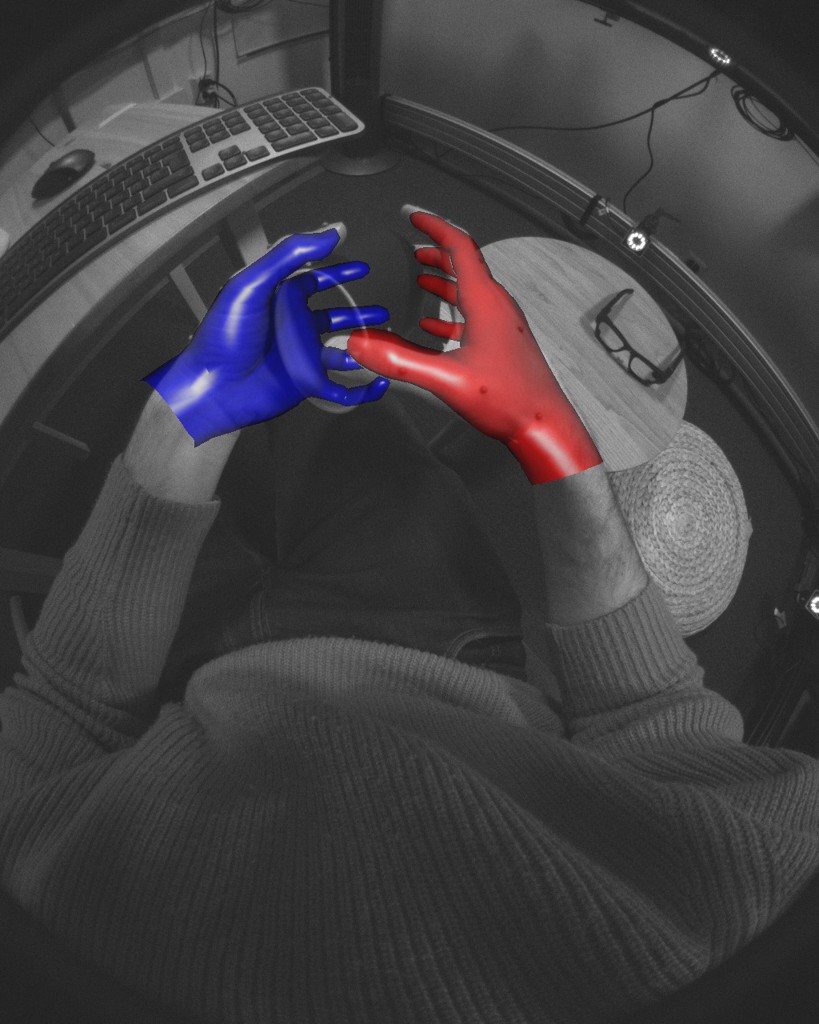} &
        \includegraphics[width=0.19\linewidth]{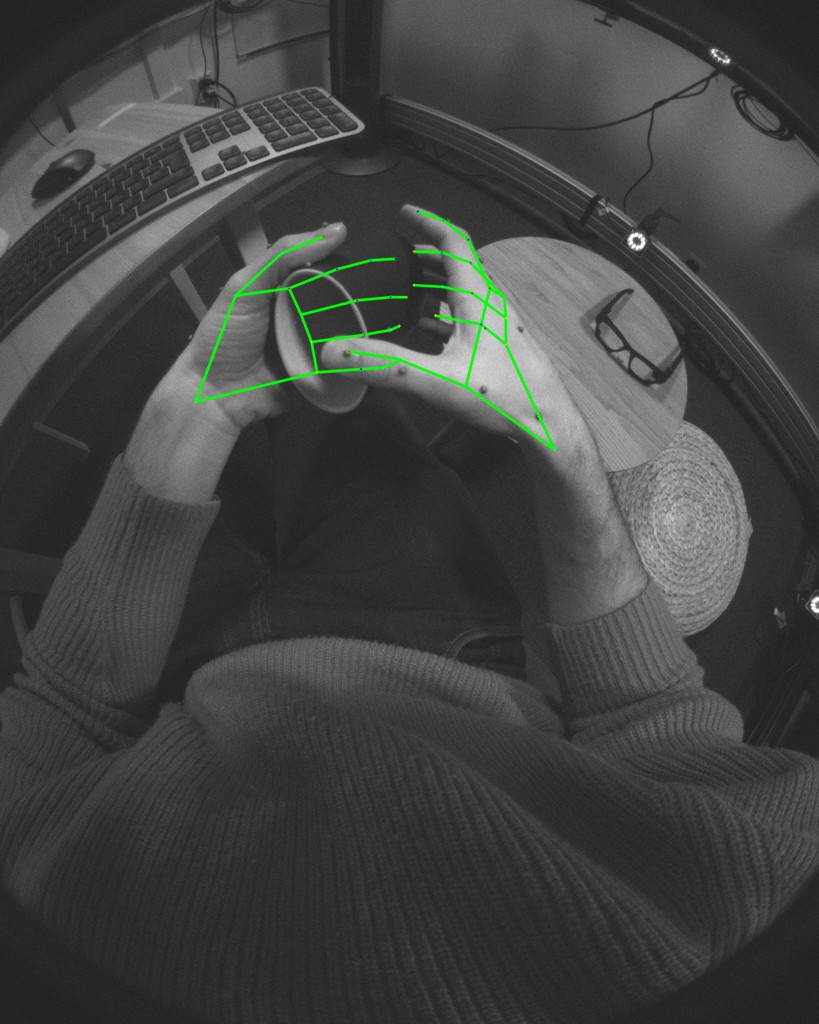} &
        \includegraphics[width=0.19\linewidth]{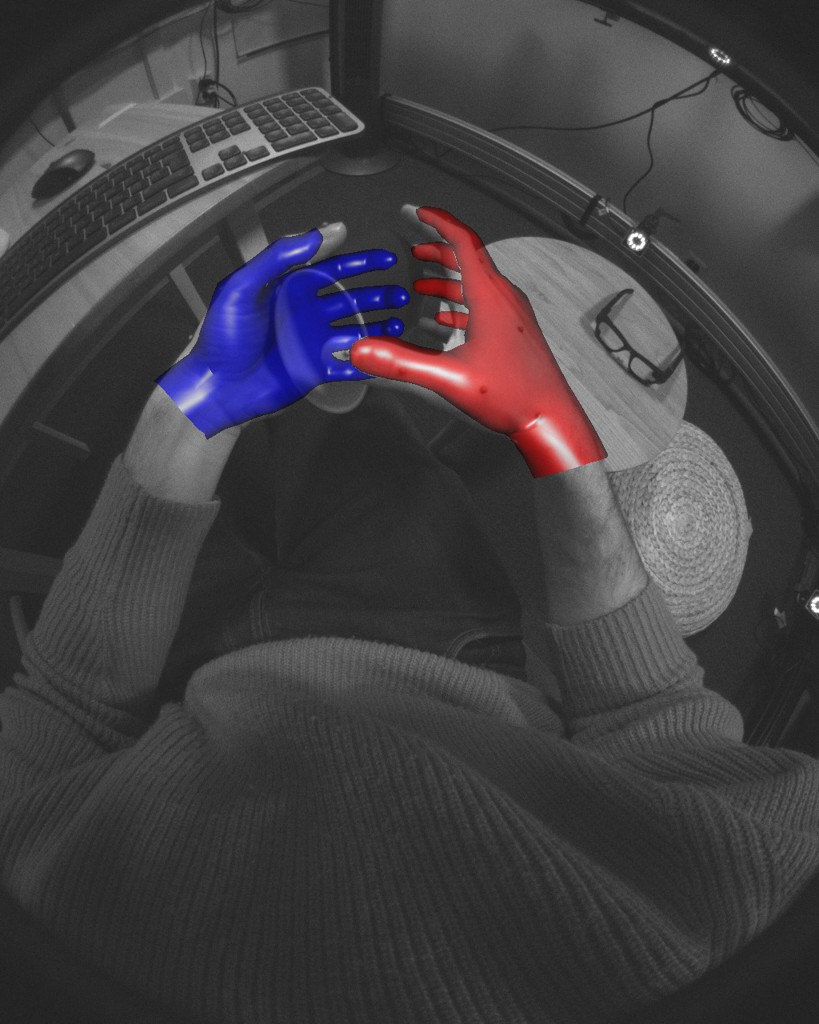} \\
        \includegraphics[width=0.19\linewidth]{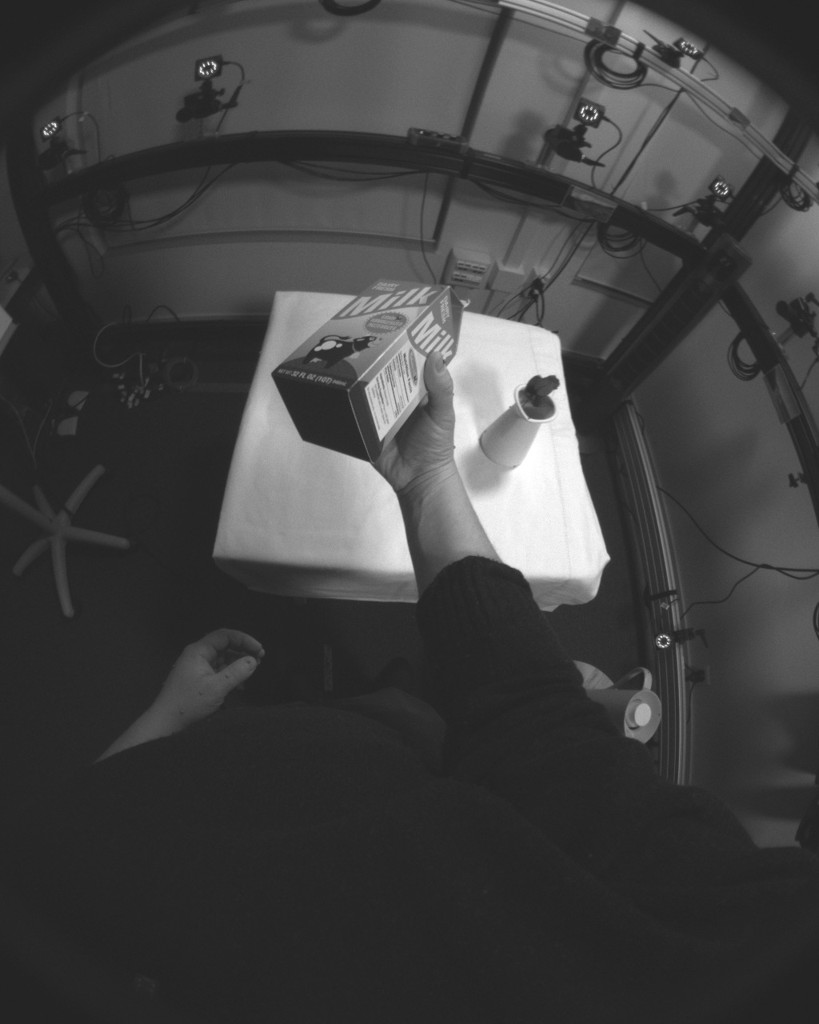} &
        \includegraphics[width=0.19\linewidth]{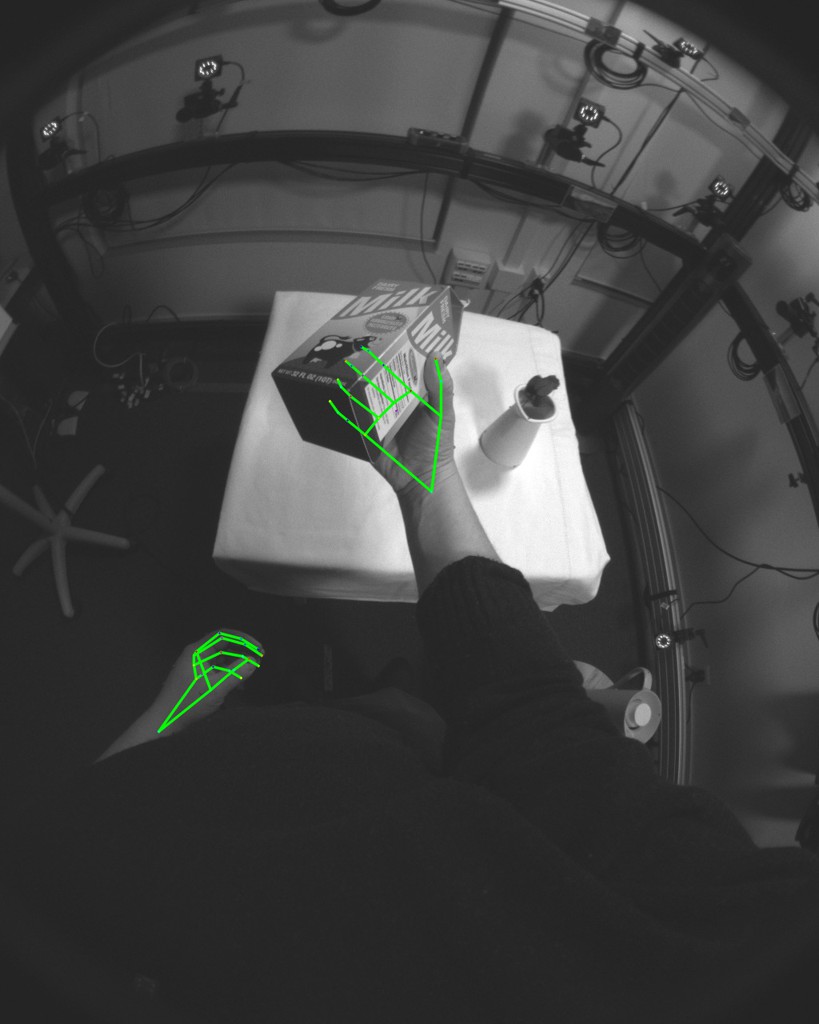} &
        \includegraphics[width=0.19\linewidth]{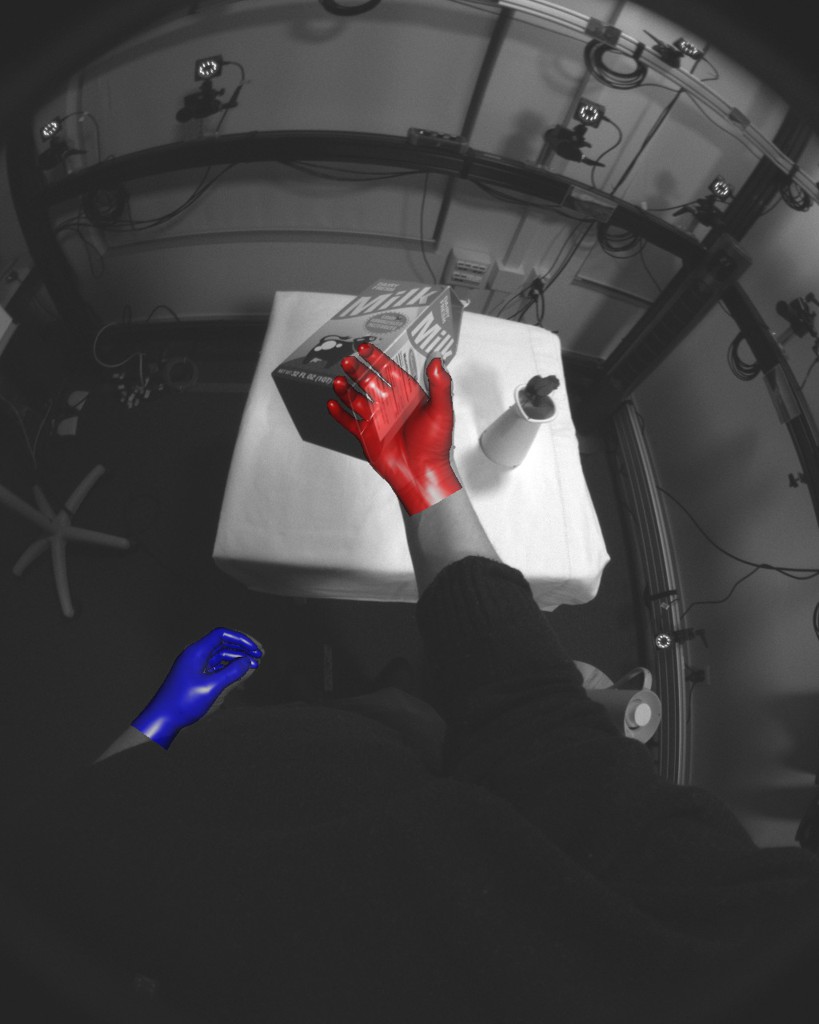} &
        \includegraphics[width=0.19\linewidth]{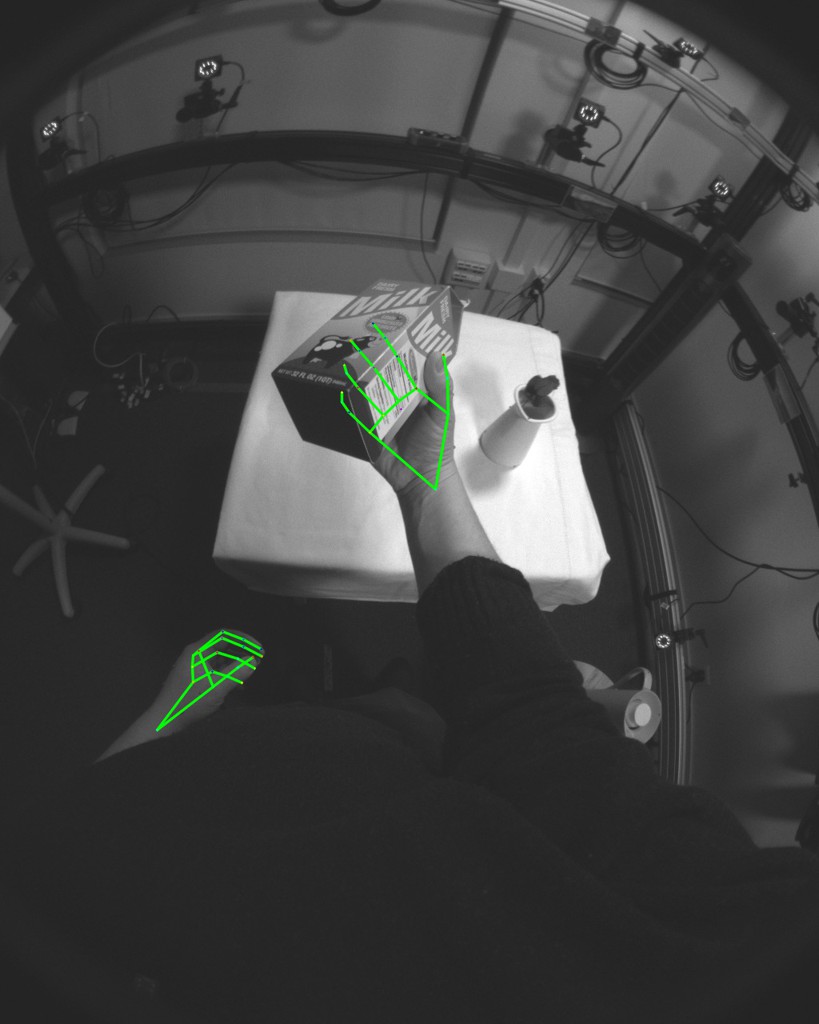} &
        \includegraphics[width=0.19\linewidth]{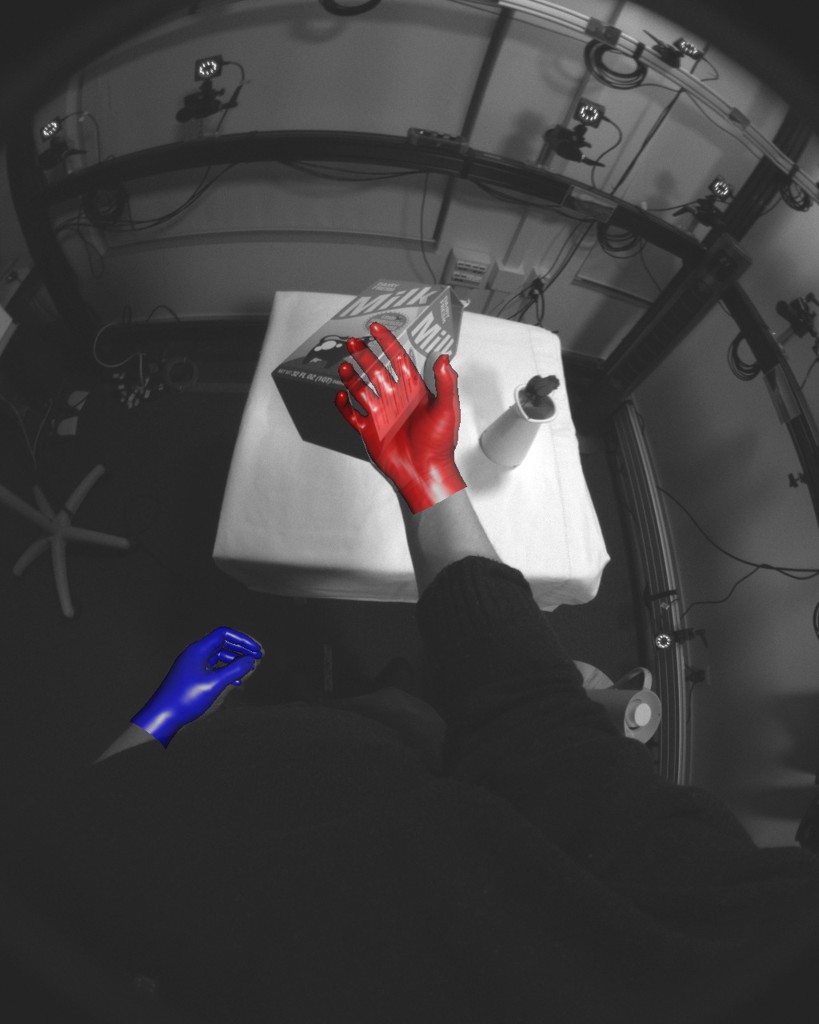} \\
    \end{tabular}
   \caption{
       \textbf{Example 3D hand pose tracking results by UmeTrack~\cite{han2022umetrack}.} 
       Shown are hand skeletons and meshes of the UmeTrack hand model in the ground-truth and estimated poses.
       The UmeTrack hand tracker was provided the ground-truth hand skeleton and was tasked to estimate 3D locations of the skeleton joints.
       } 
    \label{fig:umetrack}
\end{figure}

\subsection{6DoF object pose estimation}

\noindent\textbf{Experimental setup.} In this experiment we focus on training-free 6DoF object pose estimation, where objects are onboarded during a short stage using only their CAD models~\cite{hodan2023bop}. We evaluate FoundPose~\cite{ornek2023foundpose}, a recent open-source method that achieves state-of-the-art results in refinement-free pose estimation from a single RGB image, and its extension to multi-view input that we propose below. We evaluate refinement-free versions of these methods
on every 30th frame of the test HOT3D clips from both Aria and Quest 3.
The input of the methods is a single/multi-view frame with ground-truth 2D segmentation masks of visible objects.
The accuracy of the estimated poses is measured by a recall rate defined as the fraction of samples for which a correct pose was estimated. A pose estimate is considered correct if its symmetry-aware translational and rotational errors are below a threshold.

\begin{table}[t!]
    \setlength{\tabcolsep}{3.5pt}
    \small
    \begin{center}
    \begin{tabularx}{1.0\linewidth}{c c Y Y Y}
    \toprule
    & & \multicolumn{3}{c}{Recall [\%] $\uparrow$} \\
    \cmidrule{3-5}
     Test dataset & Views & 5\,cm, 5\textdegree & 10\,cm, 10\textdegree & 20\,cm, 20\textdegree \\
    \toprule
    HOT3D-Aria & 1 & 25.2 & 41.7 & 54.5 \\
    HOT3D-Aria & 3 & \textbf{33.8} & \textbf{52.9} & \textbf{66.2} \\
    \midrule
    HOT3D-Quest3 & 1 & 28.9 & 46.6 & 58.9 \\
    HOT3D-Quest3 & 2 & \textbf{36.9} & \textbf{55.9} & \textbf{66.4} \\
    \bottomrule
    \end{tabularx}
    \end{center}
    \caption{\textbf{6DoF object pose estimation by FoundPose~\cite{ornek2023foundpose}.}
    The proposed multi-view extension of FoundPose is compared with the original single-view version in recall rates on test images from both headsets.
    The recall rate is defined as the fraction of pose estimates whose translational and rotational errors are below a specified threshold.
    }
    \label{tab:object_pose}
\end{table}

\begin{figure}[t!]
    \centering
    \footnotesize
    \setlength{\tabcolsep}{1pt} %
    \renewcommand{\arraystretch}{0.6} %
    \begin{tabular}{ccc}
        RGB & Monochrome 1 & Monochrome 2 \vspace{2pt} \\
        \includegraphics[width=0.325\linewidth]{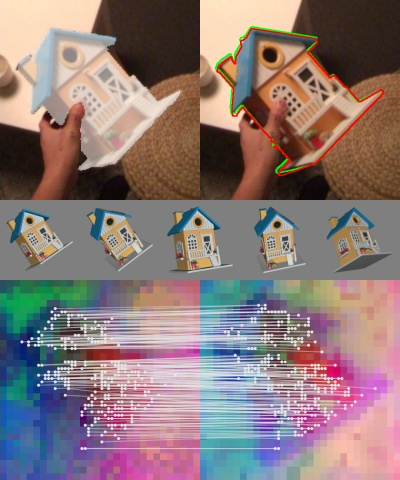} &
        \includegraphics[width=0.325\linewidth]{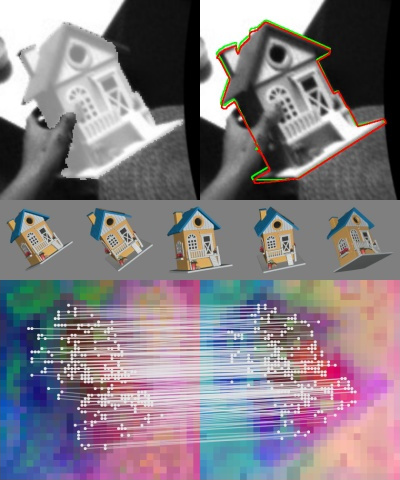} &
        \includegraphics[width=0.325\linewidth]{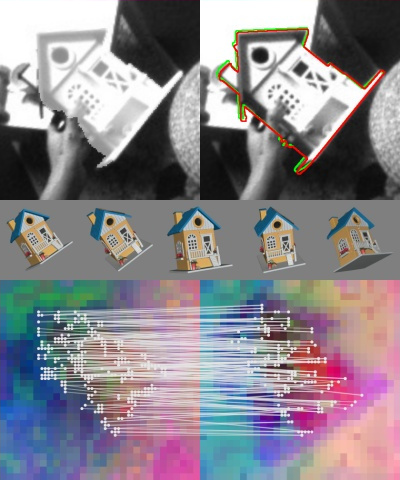} \\
        \vspace{-0.4ex} & & \\
        \includegraphics[width=0.325\linewidth]{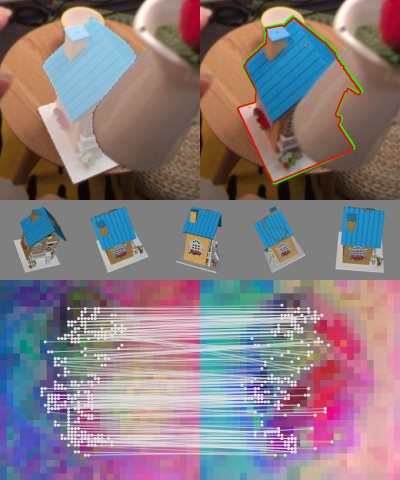} &
        \includegraphics[width=0.325\linewidth]{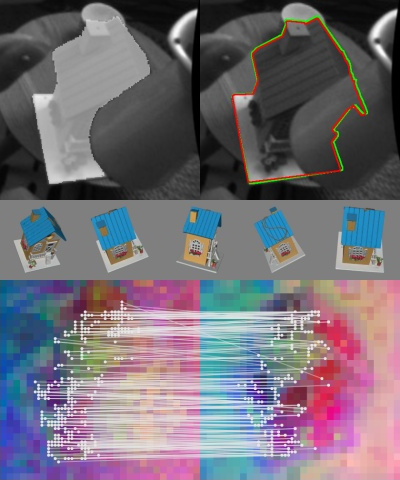} &
        \includegraphics[width=0.325\linewidth]{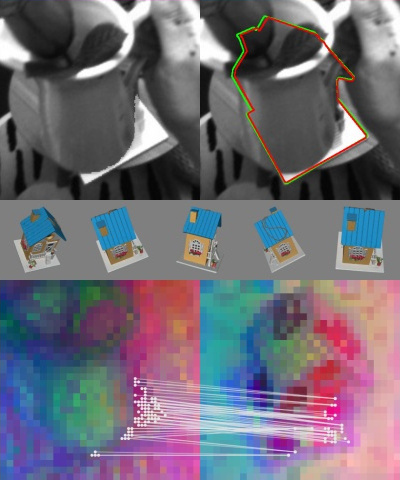}
    \end{tabular}
   \caption{
       \textbf{Example 6DoF pose estimation results by FoundPose~\cite{ornek2023foundpose}.}
        Each row shows synchronized views of the same object from three Aria cameras. Our multi-view extension of FoundPose estimates the object pose from 2D-3D correspondences established between all available views and retrieved RGB-D templates. Thanks to the multi-view input, the method is able to estimate poses of heavily occluded objects (bottom row).
        Contours of 3D object models in the ground-truth and the estimated poses are shown in green and red respectively (top right corners). FoundPose was provided the ground-truth object mask as input (top left corners).
       } 
\end{figure}

\customparagraph{FoundPose~\cite{ornek2023foundpose} and its multi-view extension.} During a short onboarding stage, FoundPose renders RGB-D templates showing the object model in different orientations, extracts DINOv2 patch features from the RGB channels, and registers the features in 3D using the depth channel. At inference time, FoundPose crops the RGB query image around the object mask and retrieves a small set of most similar templates using a DINOv2-based bag-of-words approach. For each retrieved template, a pose hypothesis is generated by P\emph{n}P-RANSAC from 2D-3D correspondences established by matching DINOv2 patch features of the image crop and the template. Finally, the pose hypothesis with the highest number of inlier correspondences is selected as the pose estimate.

To study the effect of multi- \vs single-view input for this task, we propose a straightforward multi-view extension of the original FoundPose method. The extended version relies on the same onboarding stage, but at inference time crops the object in all available views, retrieves a handful of templates with the highest sum of per-view template scores, establishes multi-view 2D-3D correspondences between each of the templates and all views, and solves for the pose by solving the generalized P\emph{n}P problem~\cite{kukelova2016efficient}.

\customparagraph{Results (Tab.~\ref{tab:object_pose}).} Our straightforward multi-view extension of FoundPose significantly outperforms the original single-view version, achieving 8--12\% higher recall rates (13--34\% relative improvement) on data from both headsets. Besides introducing additional constraints for 3D reasoning, the multi-view input offers more opportunities to observe objects that may be heavily or fully occluded in a single view. It is noteworthy that FoundPose performs well on the two-view monochrome setup from Quest3 and also the three-view setup from Aria, where one view is captured by a high-resolution RGB camera and the other two by lower-resolution monochrome cameras. We attribute this ability to generalize across different sensors primarily to DINOv2, which serves as a backbone for FoundPose.

\subsection{2D segmentation of in-hand objects}

\noindent\textbf{Experimental setup.} Given an input image, the task is to predict a binary 2D mask of objects manipulated by hands. Besides serving as a prerequisite for 3D lifting of in-hand objects (Sec.~\ref{sec:exp_lifting}), this task is useful for downstream applications such as hand state classification or video activity recognition~\cite{zhang2022fine}. We evaluate three methods, including the off-the-shelf EgoHOS~\cite{zhang2022fine} and two variants of Mask R-CNN~\cite{he2017mask}, trained on a proprietary dataset of 400K RGB Aria images annotated with 52K masks of in-hand objects. We trained one model directly on the RGB image channels (denoted as MRCNN in Tab.~\ref{tab:2d_seg}), and one on the depth channel predicted by Depth Anything V2~\cite{depth_anything_v2} (denoted as MRCNN-DA). We evaluate these methods on every 30th frame of the training and test HOT3D clips from both Aria and Quest3 ($\sim$19K frames). Objects are considered to be in-hand if the minimum distance between object and hand mesh vertices in their ground-truth poses is below a threshold of 1\,cm and the object is moving with a velocity larger than 1\,cm/s. Masks of such objects are used as the ground truth for this task.
The accuracy of predicted masks is measured by mean Intersection over Union (mIoU), as in~\cite{zhang2022fine}.

\customparagraph{Results (Tab.~\ref{tab:2d_seg}).} The EgoHOS~\cite{zhang2022fine} model exhibits a noticeable decline in accuracy on HOT3D compared to its performance on the EgoHOS dataset, which we attribute to the domain gap between the datasets.
This is particularly evident in the $\sim50\%$ lower accuracy on Quest 3 monochrome images.
MRCNN-DA is the top performing method,
outperforming EgoHOS on Aria and Quest 3 frames by 30\% and 65\%
respectively. The incorporation of predicted depth maps enables more accurate disambiguation of foreground in-hand objects from the background, resulting in improved segmentation masks (Fig. \ref{fig:inhandobj_2dseg}).

\begin{table}[t!]
	\setlength{\tabcolsep}{3.5pt}
	\small
	\begin{center}
		\begin{tabularx}{1.0\linewidth}{c c Y Y Y Y}
			\toprule
			    & & \multicolumn{4}{c}{Object in hand (mIoU $\uparrow$):} \\
			\cmidrule(lr){3-6}
			Method & Test dataset & Either & Left & Right & Both \\
			\toprule
			EgoHOS~\cite{zhang2022fine} & EgoHOS & -- & 62.2 & 44.4 & 52.8 \\
			\midrule
			EgoHOS~\cite{zhang2022fine} & HOT3D-Aria & 42.6 & 21.0 & 26.3 & 32.5 \\
			MRCNN & HOT3D-Aria & 47.1 & -- & -- & -- \\
			MRCNN-DA & HOT3D-Aria & \textbf{55.2} & -- & -- & -- \\
                \midrule
                EgoHOS~\cite{zhang2022fine} & HOT3D-Quest3 & 33.1 & 13.5 & 14.4 & 24.8  \\
                MRCNN & HOT3D-Quest3 & 37.8 & -- & -- & -- \\
                MRCNN-DA & HOT3D-Quest3 & \textbf{54.7} & -- & -- & -- \\
			\bottomrule
	\end{tabularx}
	\end{center}
	\caption{\textbf{2D segmentation of in-hand objects.} EgoHOS~\cite{zhang2022fine} trained on the EgoHOS dataset is compared with our baselines based on Mask R-CNN~\cite{he2017mask} and trained on our in-house dataset of images from Aria. We observe a large accuracy drop of the EgoHOS model on HOT3D.}
	\label{tab:2d_seg}
\end{table}

\begin{figure}[t]
    \centering
    \setlength{\tabcolsep}{1pt} %
    \renewcommand{\arraystretch}{0.6} %
    \begin{tabular}{cccc}
        \includegraphics[width=0.243\linewidth]{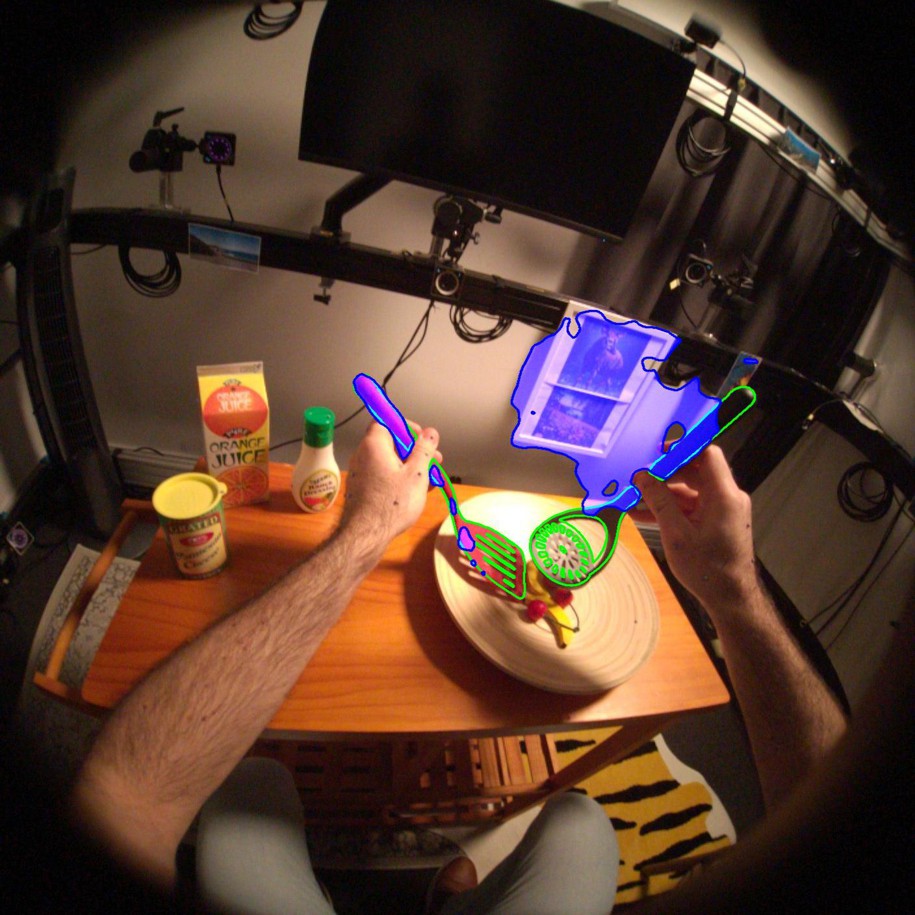} &
        \includegraphics[width=0.243\linewidth]{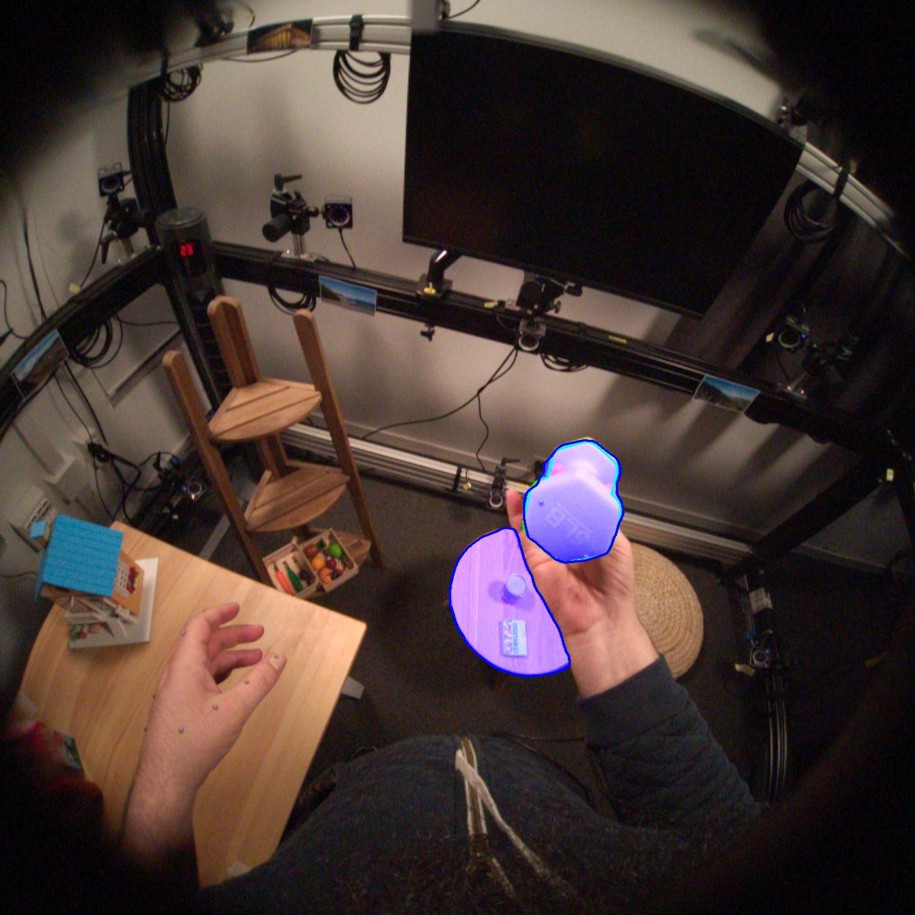} &
        \includegraphics[width=0.243\linewidth]{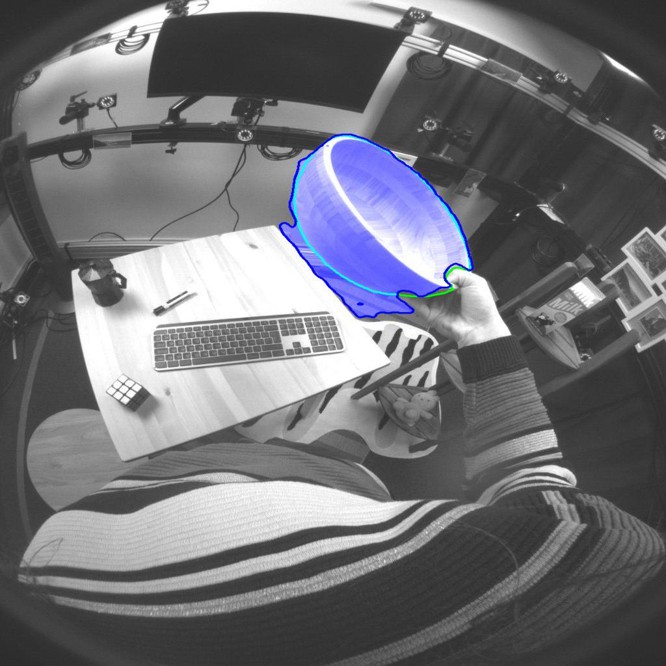} &
        \includegraphics[width=0.243\linewidth]{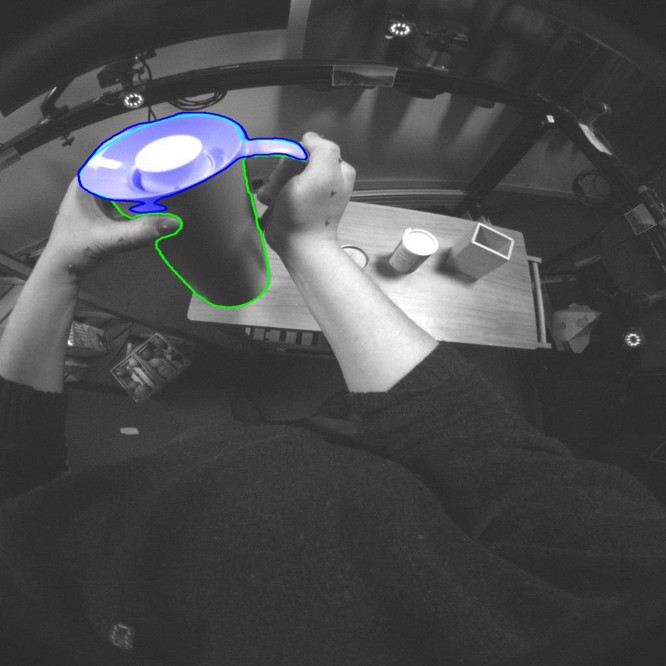} \\
        \includegraphics[width=0.243\linewidth]{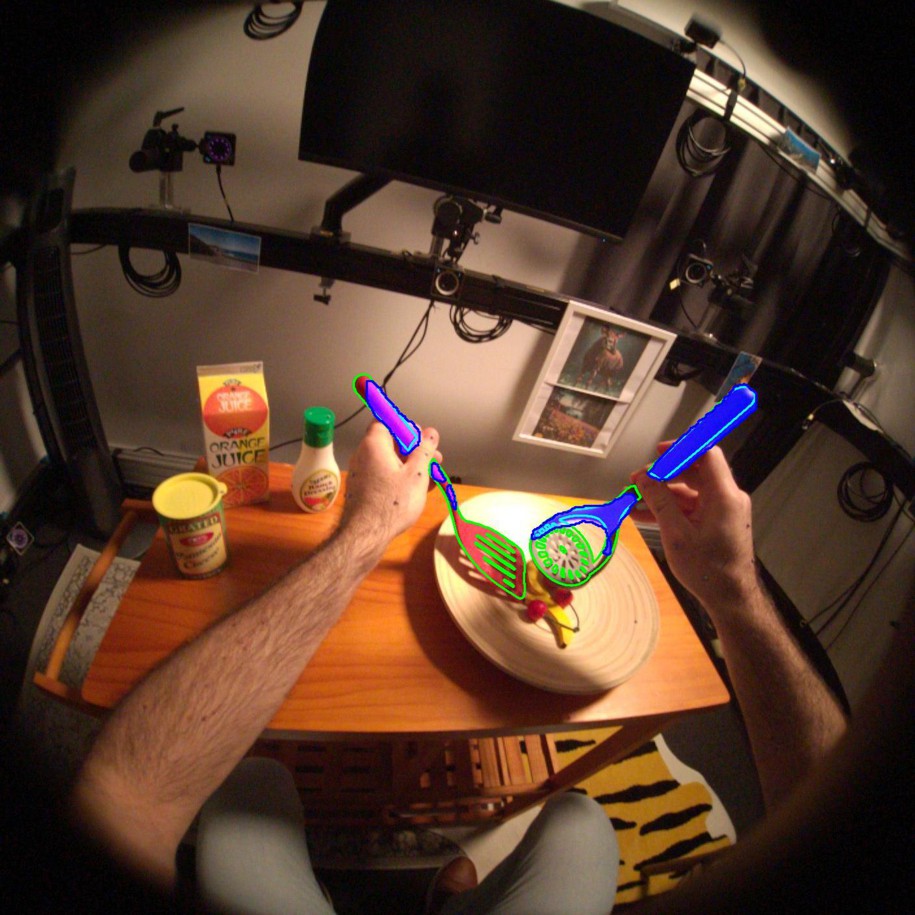} &
        \includegraphics[width=0.243\linewidth]{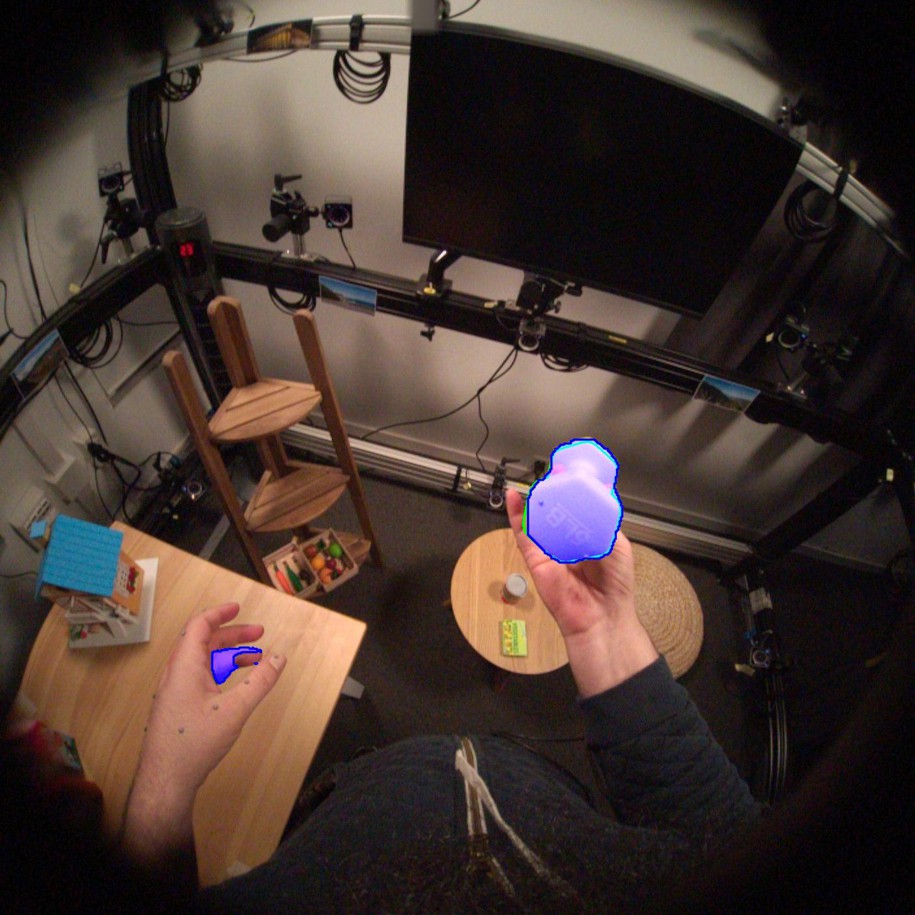} &
        \includegraphics[width=0.243\linewidth]{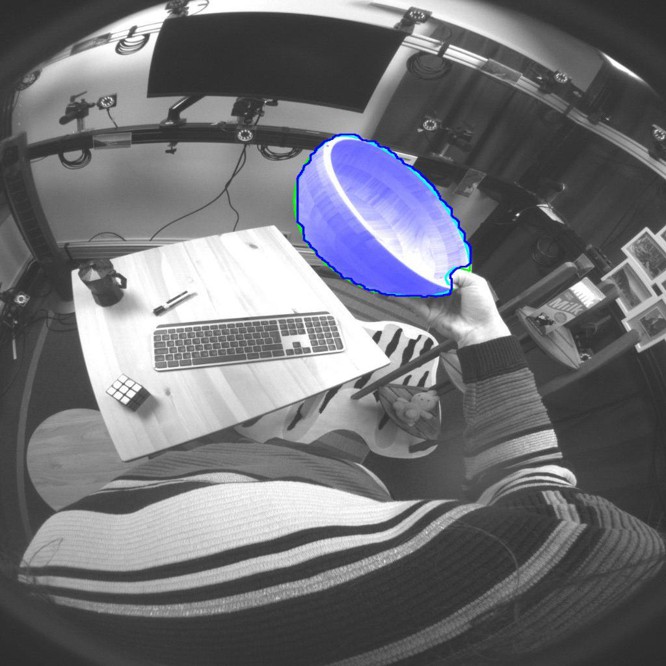} &
        \includegraphics[width=0.243\linewidth]{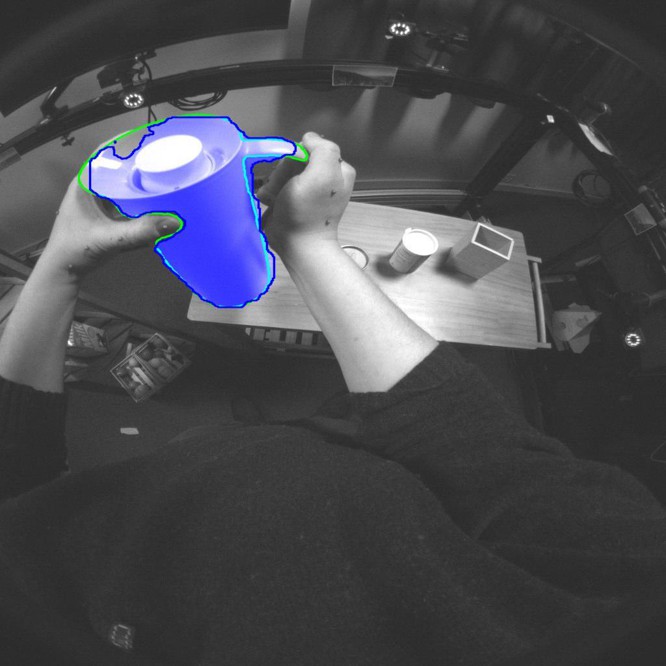} \\
        \vspace{-0.7ex} & & & \\
        \includegraphics[width=0.243\linewidth]{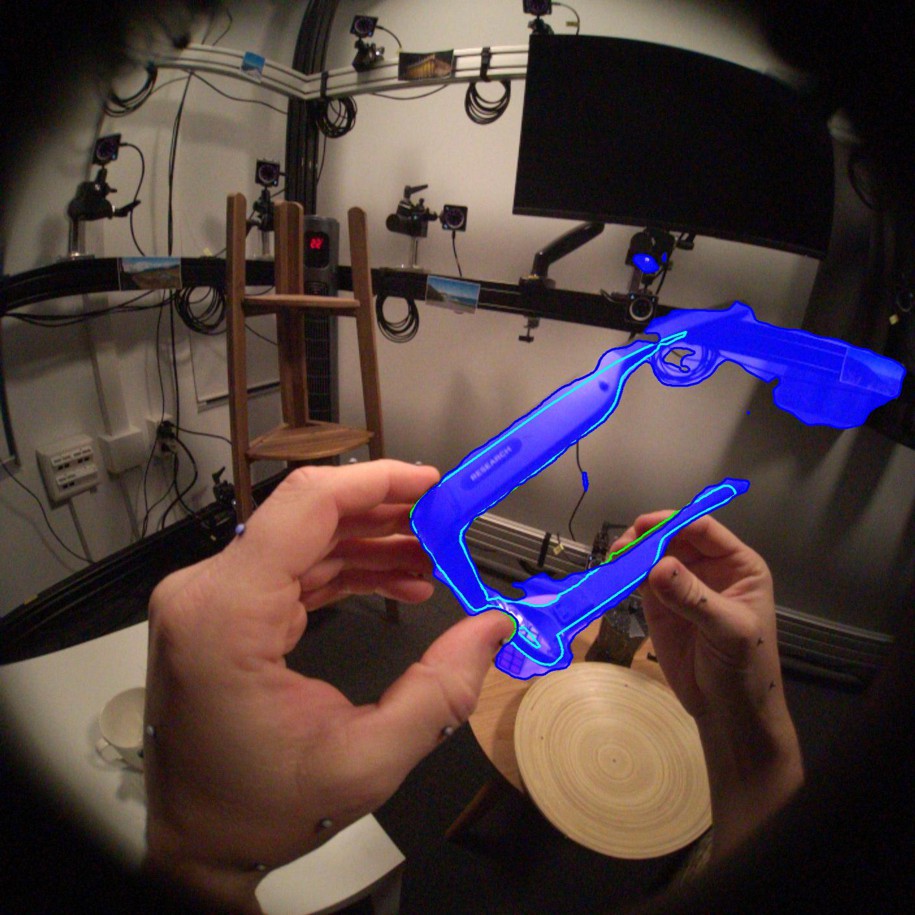} &
        \includegraphics[width=0.243\linewidth]{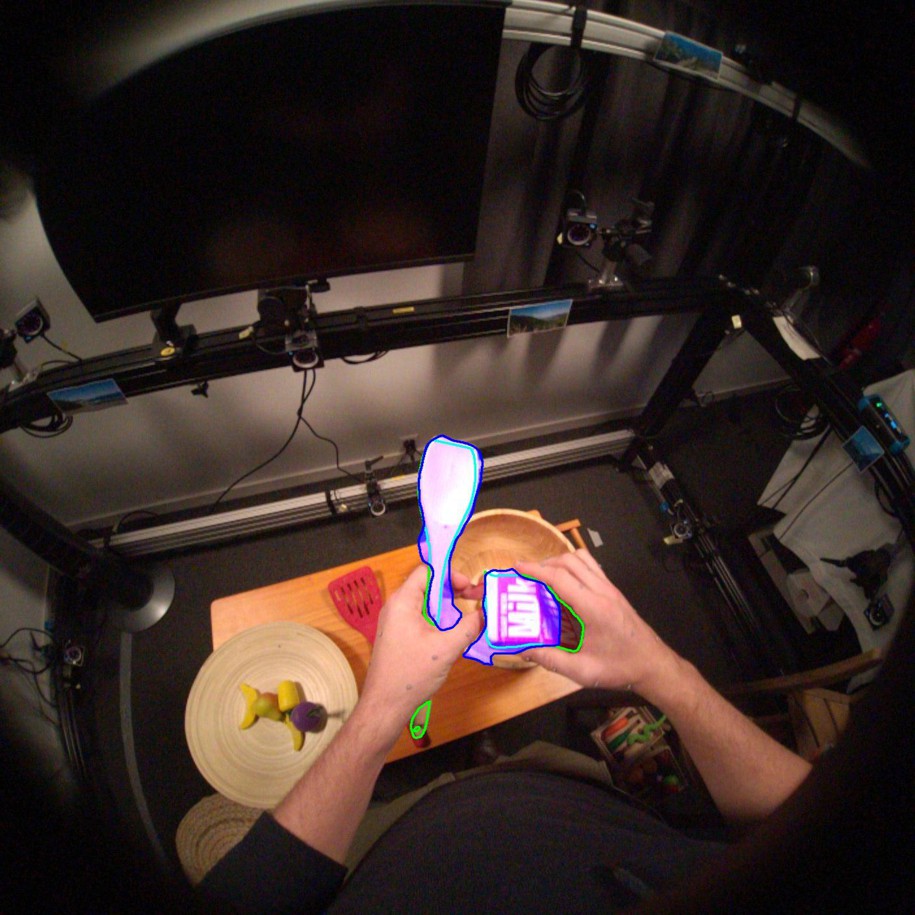} &
        \includegraphics[width=0.243\linewidth]{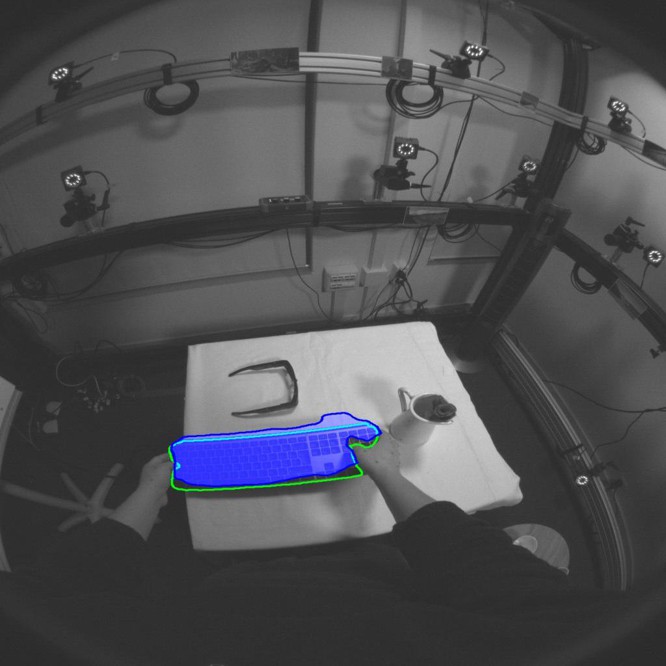} &
        \includegraphics[width=0.243\linewidth]{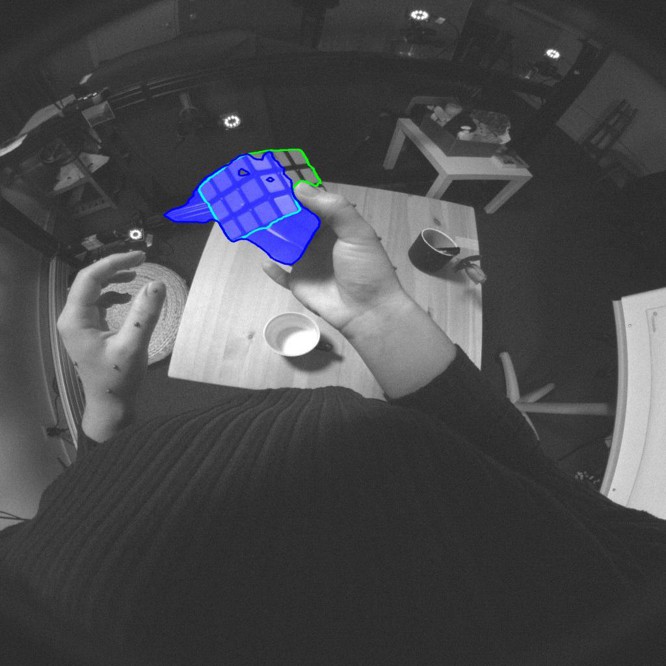} \\
        \includegraphics[width=0.243\linewidth]{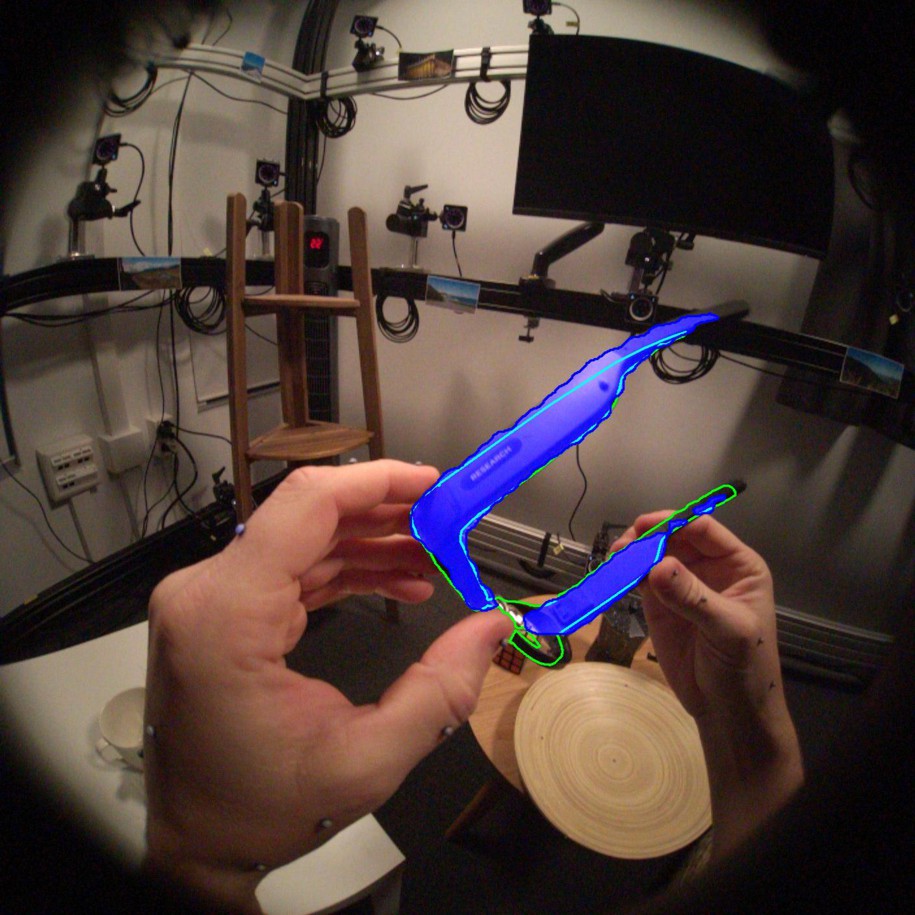} &
        \includegraphics[width=0.243\linewidth]{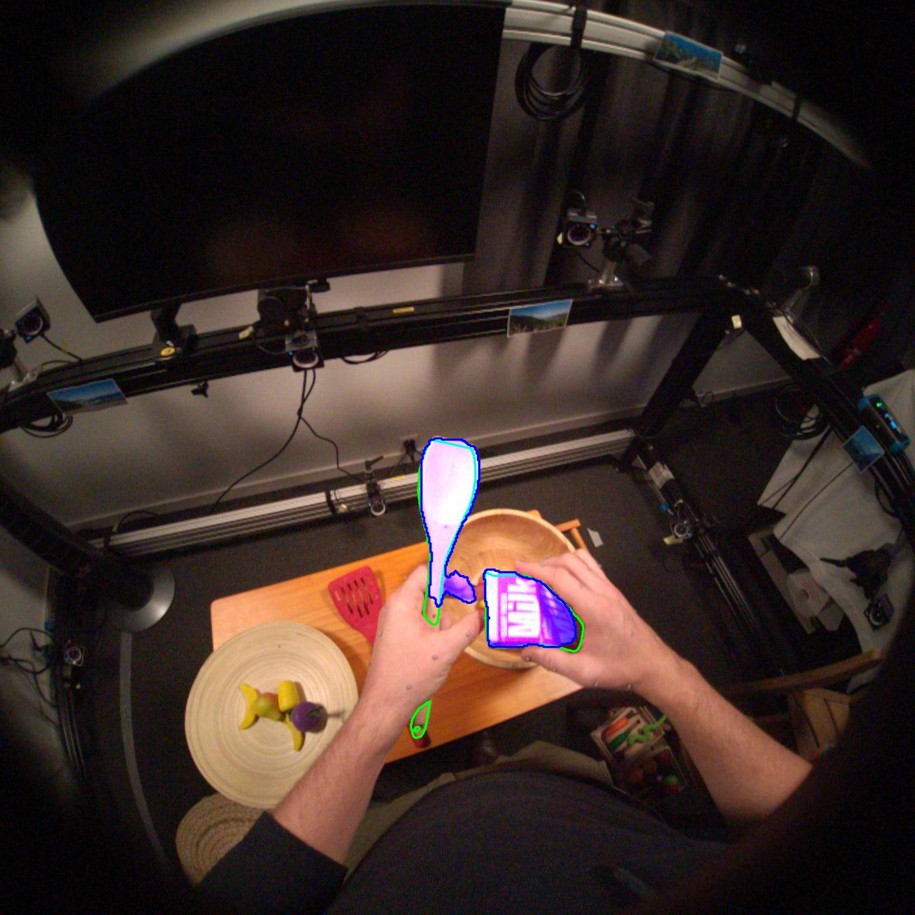} &
        \includegraphics[width=0.243\linewidth]{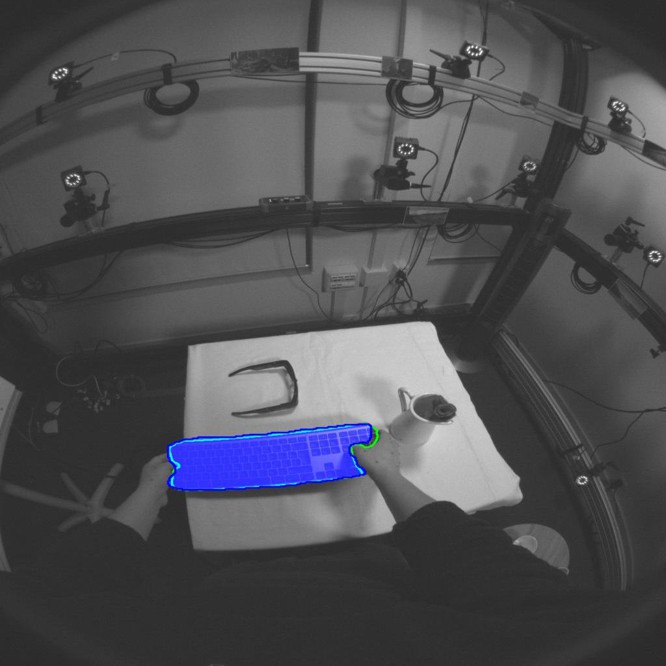} &
        \includegraphics[width=0.243\linewidth]{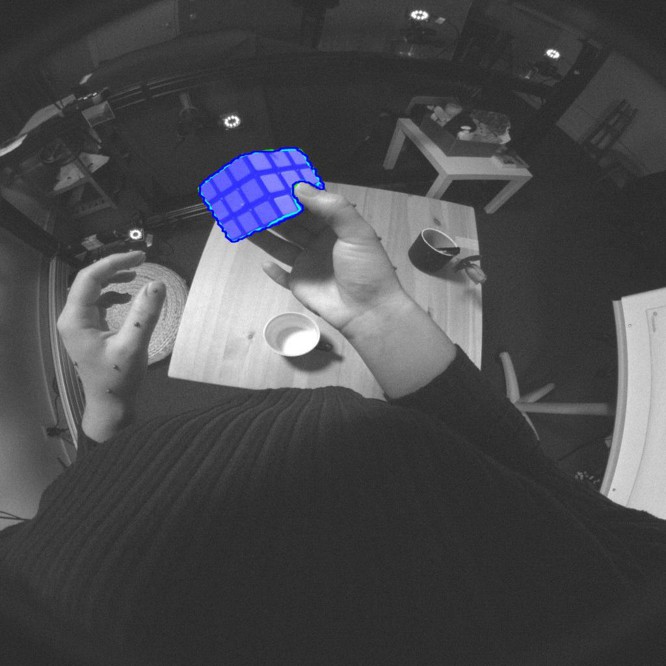}

    \end{tabular}
    \caption{
     \textbf{Example results of 2D segmentation of in-hand objects.} Masks predicted by EgoHOS~\cite{zhang2022fine} (1st and 3rd row) are compared with masks predicted by our MRCNN-DA (2nd and 4th row). Predicted masks are shown in blue, and the contour of ground-truth masks in green.
    }    
    \label{fig:inhandobj_2dseg}
\end{figure}

\setlength{\dashlinedash}{1pt}
\setlength{\dashlinegap}{1pt}
\setlength{\arrayrulewidth}{0.2pt}
\begin{table}[t!]
	\setlength{\tabcolsep}{3.5pt}
	\small
	\begin{center}
		\begin{tabularx}{1.0\linewidth}{c c c Y Y Y Y}
			\toprule
                & & & \multicolumn{4}{c}{Recall [\%] $\uparrow$} \\
                \cmidrule{4-7}
                Method & Test dataset & Views & 5\,cm & 10\,cm & 20\,cm & 30\,cm \\
                \toprule
                HandProxy & HOT3D-Aria & -- & 0.5 & 13.5 & 90.6 & 98.4 \\
			\midrule
                \multicolumn{7}{l}{{\footnotesize Using ground-truth 2D segmentation masks:}}\vspace{1.7pt} \\
			  MonoDepth & HOT3D-Aria & 1 & 14.3 & 30.2 & 53.6 & 69.9 \\
			StereoMatch\ & HOT3D-Aria & 3 & \textbf{64.4} & \textbf{86.2} & \textbf{95.5} & \textbf{96.9} \\
                \hdashline
                StereoMatch\ & HOT3D-Quest3 & 2 & 76.4 & 96.8 & 99.1 & 99.2 \\
			\midrule
                \multicolumn{7}{l}{{\footnotesize Using 2D segmentation masks predicted by MRCNN-DA:}}\vspace{1.7pt} \\
			MonoDepth & HOT3D-Aria & 1 & 11.1 & 23.3 & 43.7 & 58.2 \\
                StereoMatch\ & HOT3D-Aria & 3 & \textbf{42.6} & \textbf{56.4} & \textbf{63.6} & \textbf{66.0} \\
                \hdashline
                StereoMatch\ & HOT3D-Quest3 & 2 & 59.1 & 75.3 & 80.4 & 81.3 \\
			\bottomrule
		\end{tabularx}
	\end{center}
	\caption{\textbf{3D lifting of in-hand objects.} Shown are recall rates of our three baseline methods for several thresholds of correctness (a predicted 3D location is considered correct if its distance from the ground-truth location is below a threshold). The multi-view method (StereoMatch) clearly outperforms the other two methods at stricter threshold levels.
    }
	\label{tab:inhandobj_lift3d}
\end{table}

\begin{figure}[t!]
    \centering
    \setlength{\tabcolsep}{1pt} %
    \renewcommand{\arraystretch}{0.6} %
    \begin{tabular}{cc}
        \includegraphics[width=0.495\linewidth]{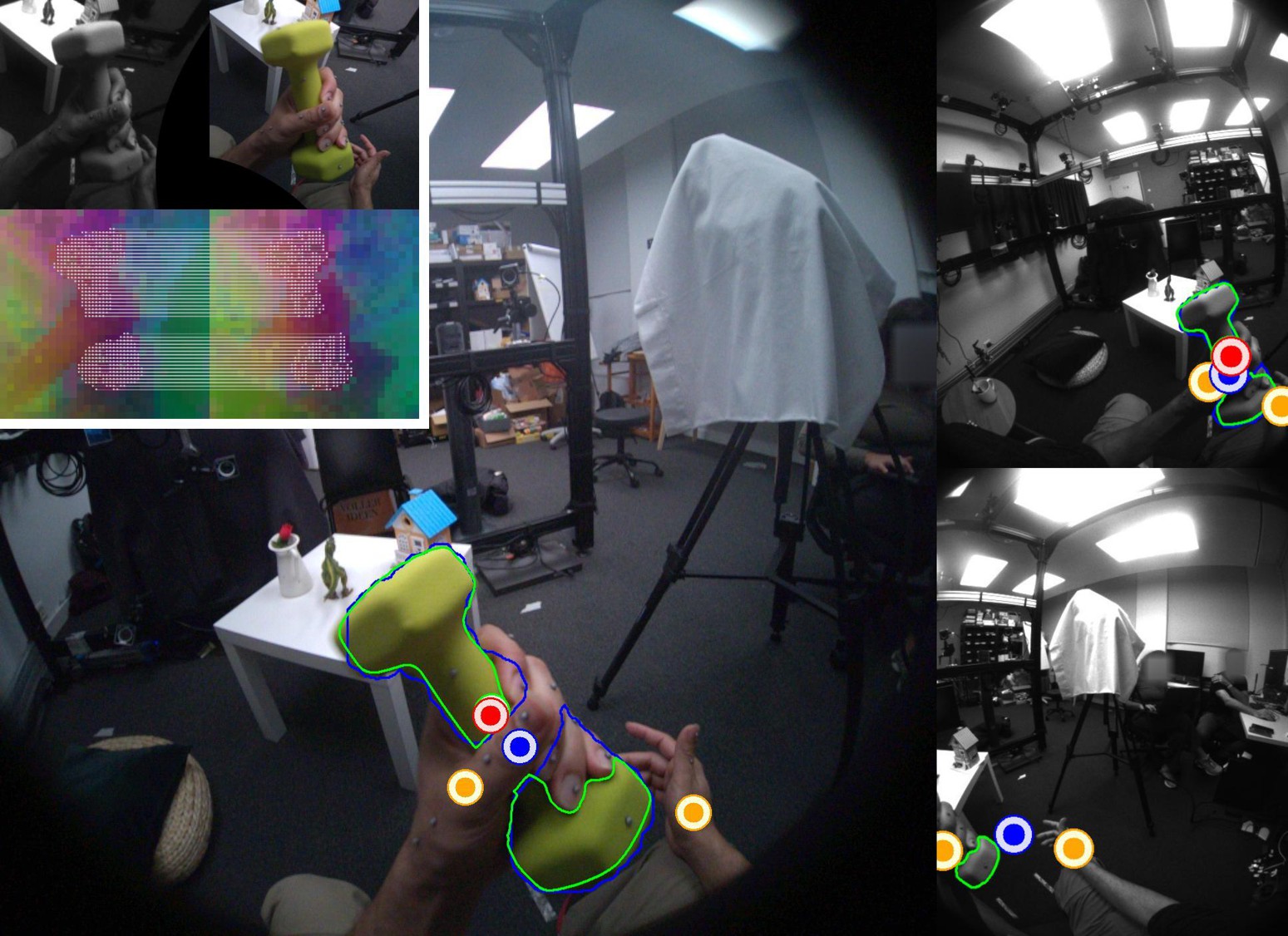} &
        \includegraphics[width=0.495\linewidth]{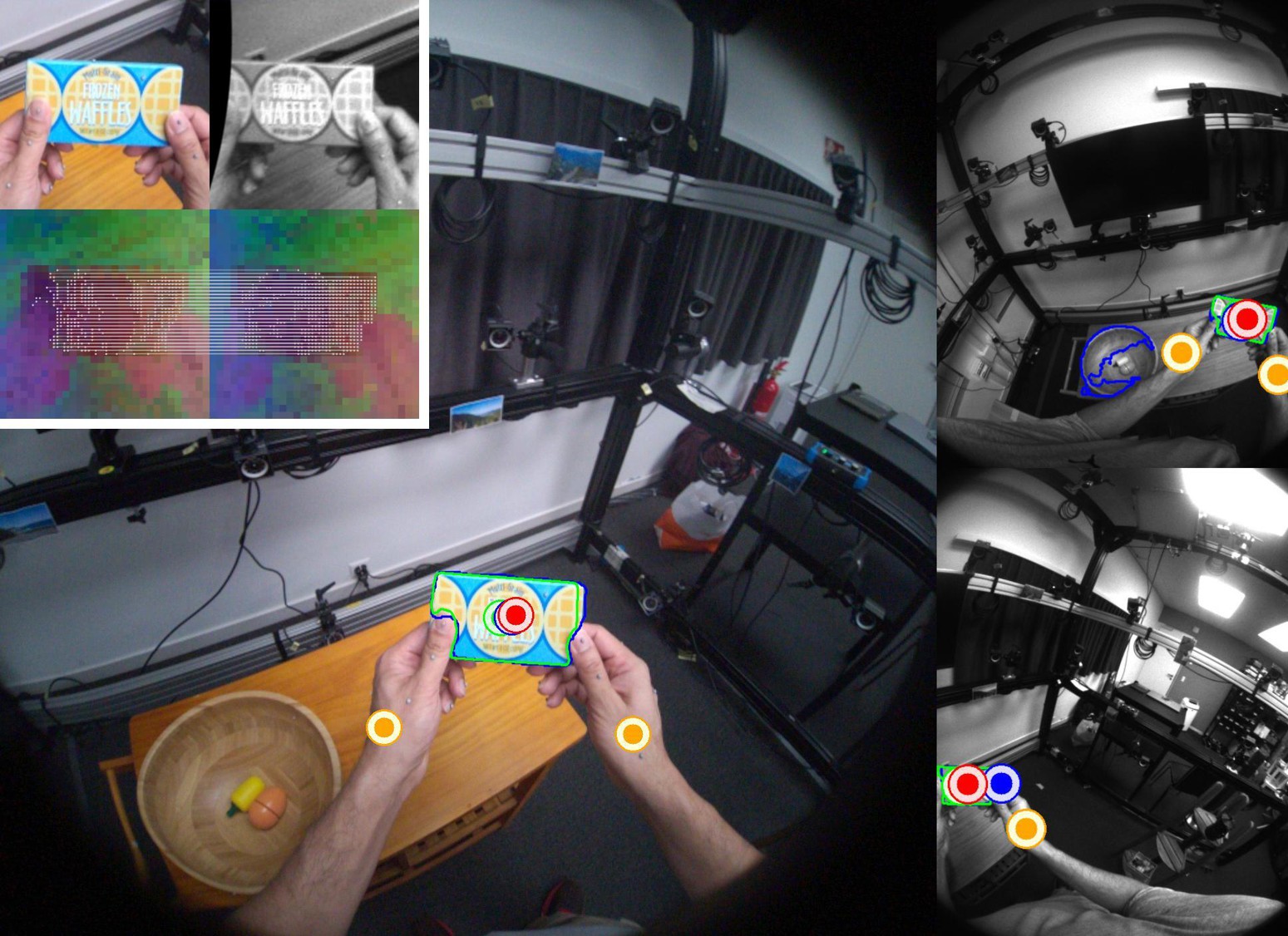} \\
        \includegraphics[width=0.495\linewidth]{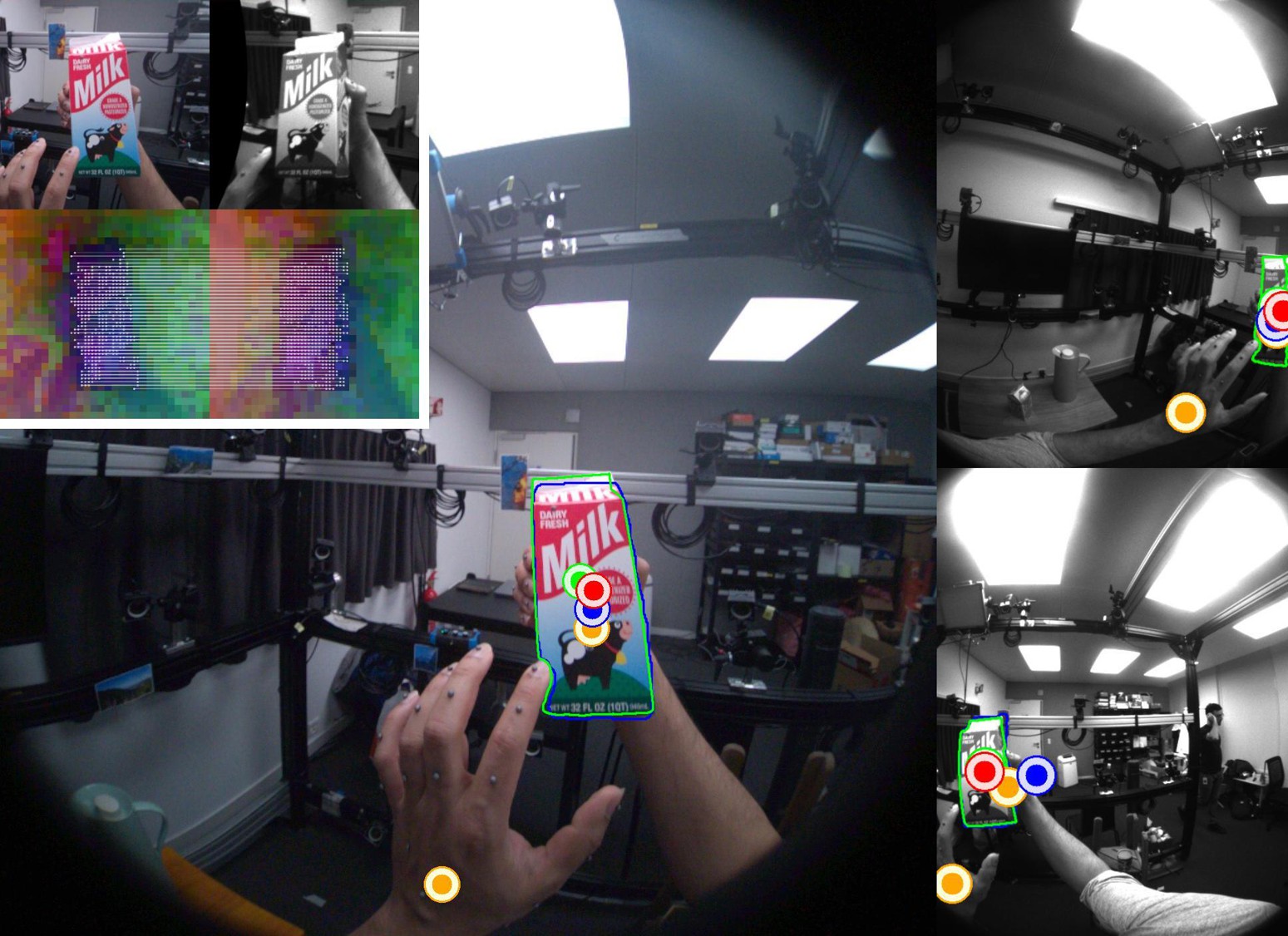} &
        \includegraphics[width=0.495\linewidth]{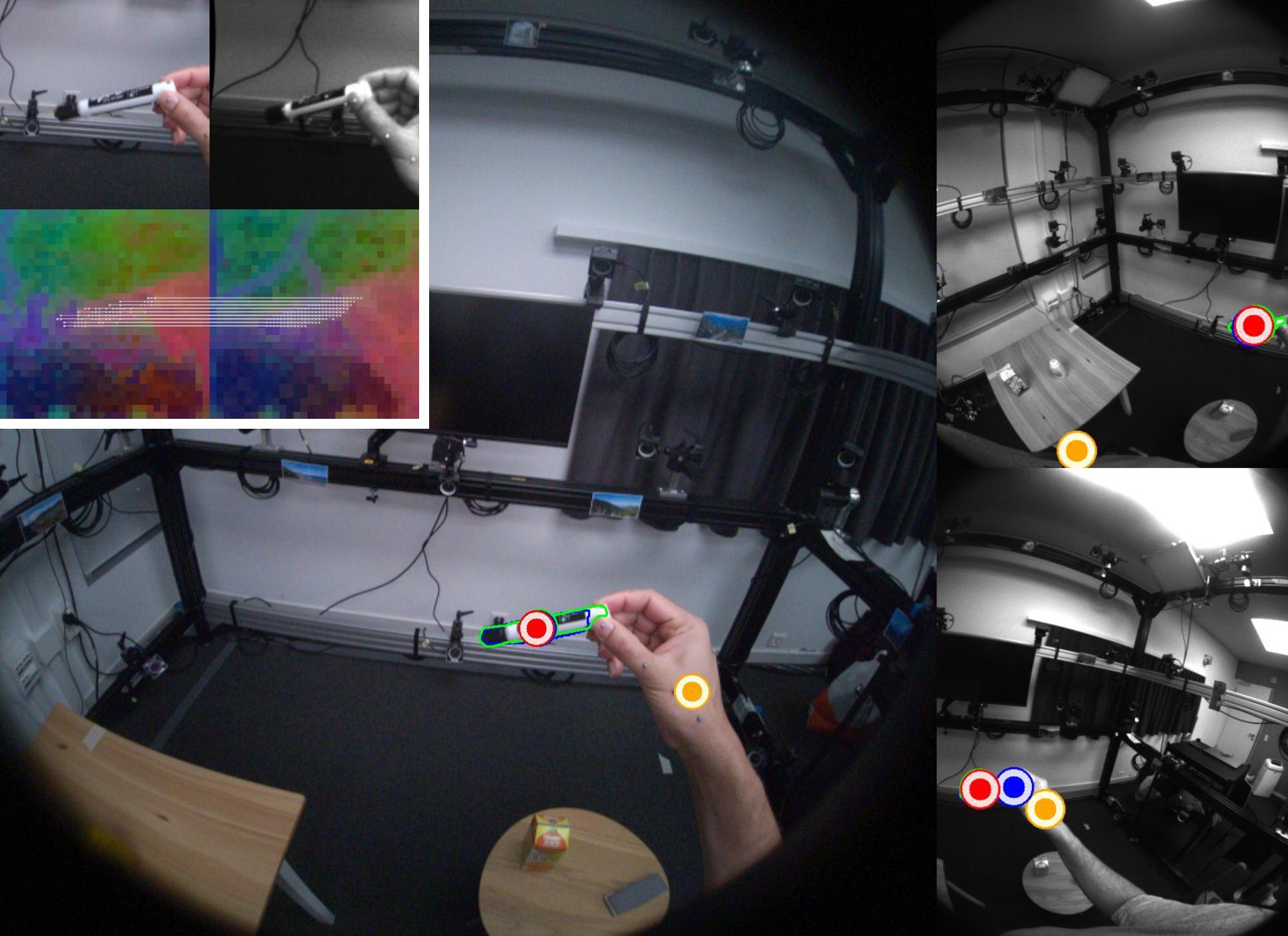} \\
        \includegraphics[width=0.495\linewidth]{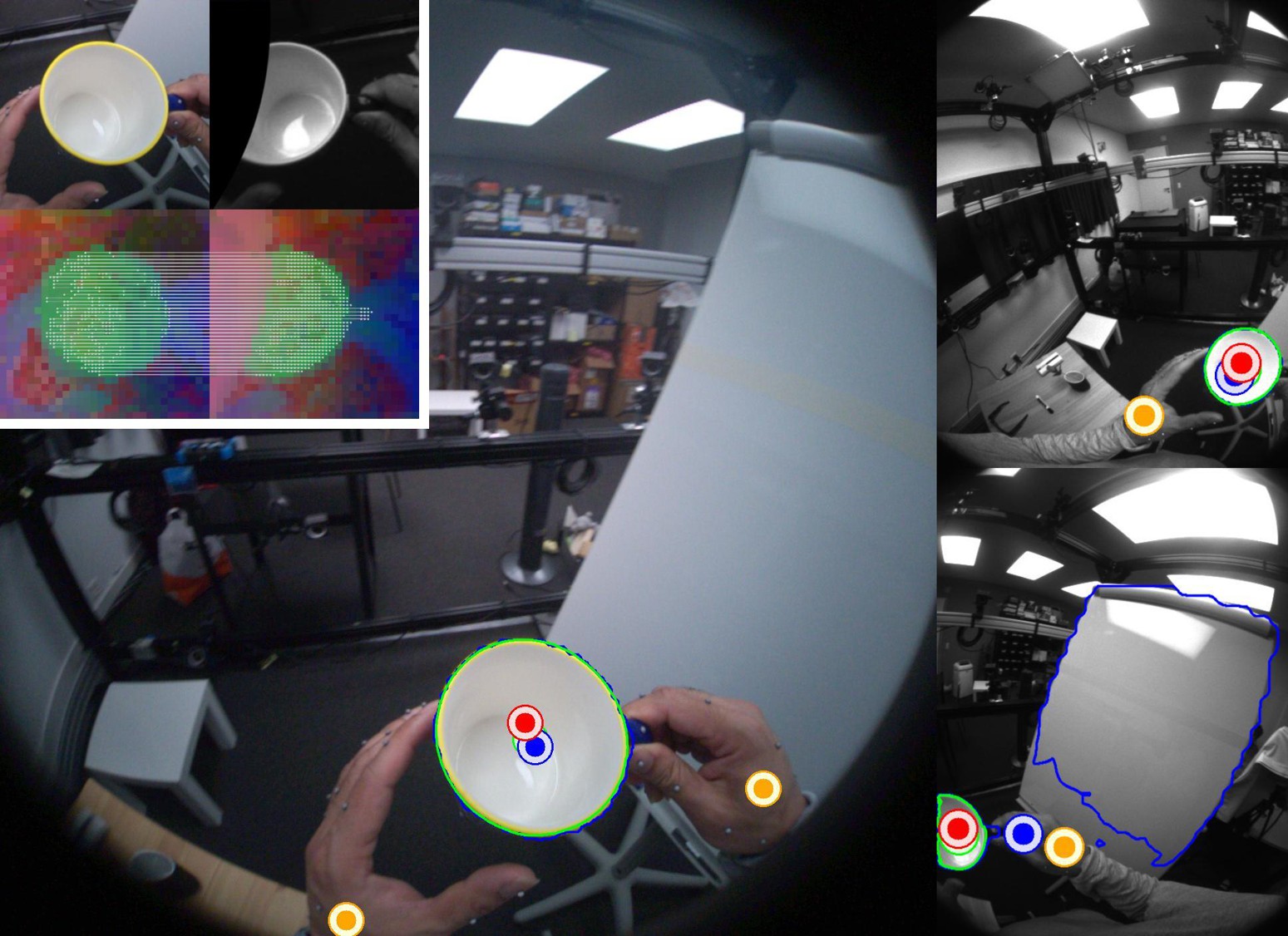} &
        \includegraphics[width=0.495\linewidth]{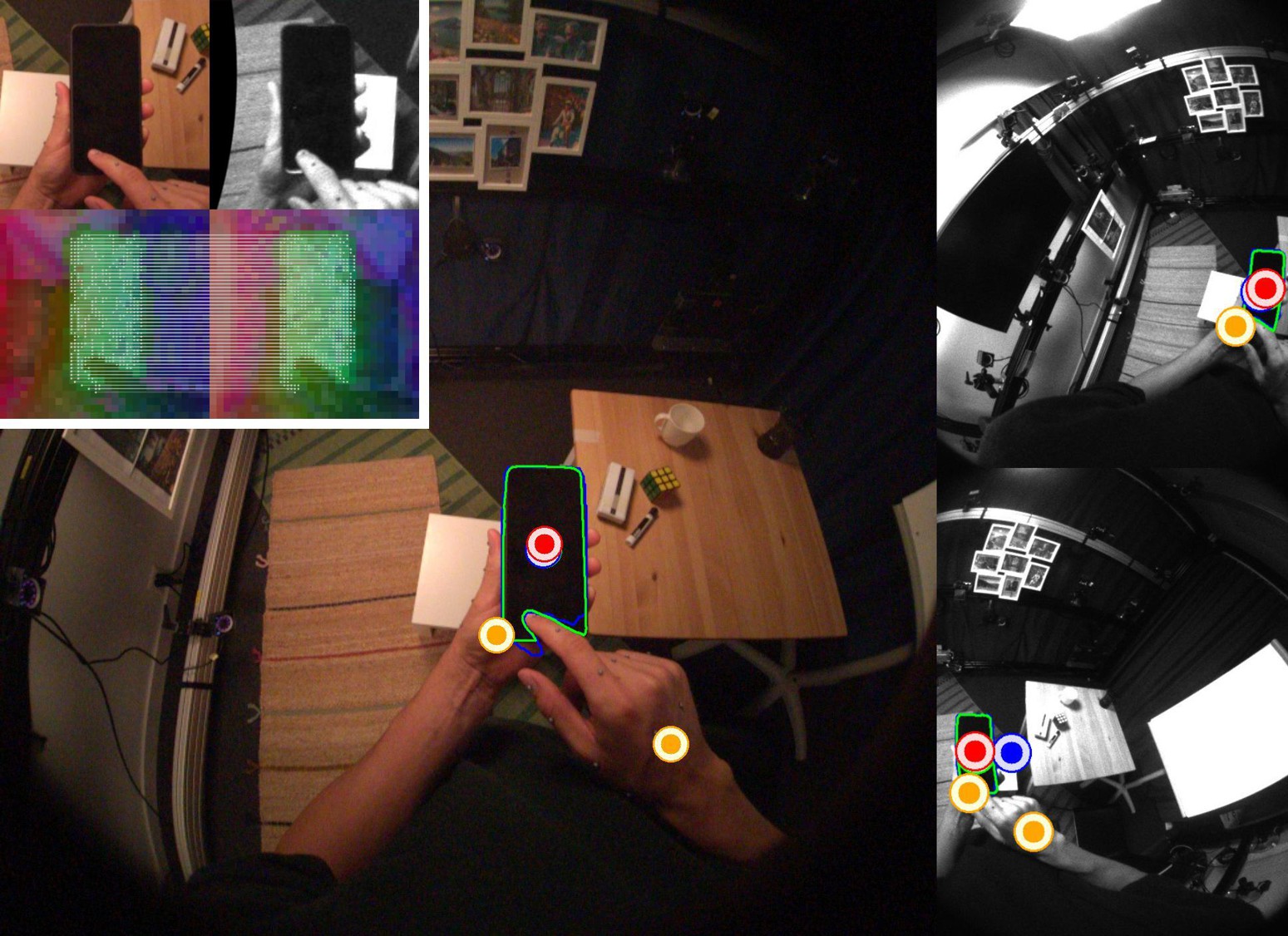} \\
    \end{tabular}
    \caption{
     \textbf{Example results of 3D lifting of in-hand objects.}~Ground-truth 3D object locations are shown in green, predictions from StereoMatch in red, MonoDepth in blue, and HandProxy in orange. The locations are projected to the three Aria views. In each example, the RGB image is on the left, monochrome images on the right, and DINOv2 stereo matching between the RGB and a monochrome image in the top left.
     }
    \label{fig:inhandobj_lift3d}    
\end{figure}

\subsection{3D lifting of in-hand objects} \label{sec:exp_lifting}

\noindent\textbf{Experimental setup.}
Finally, we focus on the task of 3D lifting of unknown in-hand objects, which is useful for object indexing and long-term tracking~\cite{plizzari2024spatial}. Given per-view 2D segmentation masks of an in-hand object, the goal is to estimate the 3D object location.
We developed and compare the performance of three methods described below.
The accuracy of the estimated locations is measured by a recall rate defined as the fraction of samples for which a correct location was estimated. A location is considered correct if its offset from the ground-truth location, defined by the center of the 3D object bounding box, is below a threshold.

\customparagraph{Using hand as the object proxy (HandProxy).}
As a simple baseline, we use the ground-truth 3D palm center as the 3D object location. We define the palm center as the middle point between the wrist joint and the first joint of the middle finger. If both hands are visible, we use the 3D palm location whose 2D projection is closer to the 2D object mask centroid. The goal of this baseline is to evaluate whether a specialized solution is necessary for this task or whether an accurate 3D hand tracker could provide a sufficient estimate.

\customparagraph{Lifting by monocular depth estimation (MonoDepth).} Our second approach is inspired by~\cite{plizzari2024spatial} and relies on monocular depth prediction and sparse SLAM observations. Specifically, we predict a monocular depth map by applying Depth Anything V2~\cite{depth_anything_v2} to a rectilinear input image (warped from the fisheye camera to a pinhole camera), align it to the 3D point cloud from SLAM by a scale-shift transformation~\cite{plizzari2024spatial}, and finally calculate the 3D object location as the median of 3D points given by the aligned depth values collected from the object mask. Since MonoDepth requires SLAM observations that are not available for Quest 3, we evaluate this method only on recordings from Aria (using RGB images).

\customparagraph{Lifting by stereo matching (StereoMatch).} Our third approach for 3D lifting of in-hand objects is based on matching DINOv2 features in stereo images. Given 2D object segmentation masks in two views, we first construct a stereo crop pair such as the crops closely surround the given masks, both have a resolution of 420$\times$420 pixels, and a pixel in one crop is guaranteed to have its corresponding pixel in the same pixel row of the other crop~\cite{bradski2008learning}. Next, we extract patch features from each of the crops using DINOv2 ViT-S~\cite{oquab2023dinov2}, and establish 2D-2D correspondences between the crops by linking each patch from one crop with the nearest patch (in terms of L2 distance between DINOv2 feature vectors) from the same row in the other view. We retain up to 500 correspondences with the smallest cyclic distance~\cite{goodwin2022zero}, and triangulate them to obtain a set of 3D points. The 3D object location is estimated using the robust mean of the 3D point set.

\customparagraph{Results (Tab.~\ref{tab:inhandobj_lift3d}).}
The StereoMatch method outperforms MonoDepth regardless of whether the ground-truth segmentation masks or masks predicted by MRCNN-DA are used as input. The 3D localization error of MonoDepth is primarily along the optical axis and therefore caused by inaccurate depth predictions. This is evident in Fig.~\ref{fig:inhandobj_lift3d} where the localization error is most pronounced in camera viewpoints with a larger baseline \wrt the input RGB camera. Finally, the HandProxy baseline demonstrates that while hand tracking alone is sufficient for localizing objects within a 30\,cm uncertainty radius (similar to evaluations in \cite{plizzari2024spatial}), a dedicated method for multi-view 3D lifting of in-hand objects provides additional value for finer-grained localization within 10\,cm.

\section{Conclusion}

We have introduced HOT3D, a large-scale
dataset designed to facilitate the training and evaluation of methods for various 2D and 3D egocentric tasks related to hand-object interaction. Our experiments show that multi-view methods, whose benchmarking is uniquely enabled by HOT3D, significantly outperform their single-view counterparts across several popular tasks. Besides multi-view 3D object tracking, we addressed the tasks of multi-view 6DoF object pose estimation and 3D lifting of in-hand objects, for which we developed strong baselines.
By publicly releasing the dataset
and co-organizing public challenges\footnote{\href{https://bop.felk.cvut.cz/challenges/bop-challenge-2024}{https://bop.felk.cvut.cz/challenges/bop-challenge-2024/}}\footnote{\href{https://github.com/facebookresearch/hand_tracking_toolkit?tab=readme-ov-file\#evaluation}{https://github.com/facebookresearch/hand\_tracking\_toolkit}} on the dataset,
we hope to accelerate research on egocentric vision and contextual AI.

\section*{Acknowledgements}
Big thanks to Lingni Ma for the initial discussions, insightful ideas, and sharing her expertise with motion capture, which all served as the foundation for the HOT3D project. Thanks to Kevin Harris and Steve Olsen for their invaluable expertise and help with setting up the motion-capture lab.
Thanks to Ben Schneiders, Mateja Kodila, Elizabeth Béres, Pablo Gomez Ponce, and Nicolas Greiner for organizing data collections and helping with data processing. Thanks to Fadime Sener, Bugra Tekin, Yann Labb{\'e}, Christian Forster, Mariano Jaimez, Rainer Stal, Timo Stoffregen, Pierluigi Taddei, Eric Sauser, Shitai Li, Lukas Bode, and Ben Schneiders for participating in the data collections. Thanks to Evgeniy Oleinik and Maien Hamed for help with hosting the dataset on Meta infrastructure. Thanks to Xiaqing Pan for his guidance on open sourcing the HOT3D toolkit. Thanks to Mark Schwesinger, Christopher Pistritto, Elizabeth Argall, and Josiah Zacharias for providing legal and writing support to release HOT3D on time. Thanks to Nick Charron, Alexander Gamino, and David Caruso for their help with calibration and SLAM-OptiTrack alignment. Thanks to Ka Chen, Allison Tilp, and Thibaud Gayraud for supporting the creation of high-fidelity CAD models of the HOT3D objects. Thanks to Abha Arora, Luis Pesqueira, and Yuyang Zou for helping with organizing the Aria data campaign and setting up the Aria data collection infrastructure. Thanks to Tomasz Malisiewicz for his help with training models for 2D in-hand object segmentation. Thanks to Weiguang Si for his help with preparing HOT3D for the Hand Tracking Challenge at ECCV 2024.

{\small
\bibliographystyle{ieee_fullname}
\bibliography{references}
}

\appendix

\section*{Appendix}

In this appendix, we provide details about the Aria glasses (Sec.~\ref{sec:appendix-aria}) and the Quest 3 headset (Sec.~\ref{sec:appendix-quest}) which were used to record the HOT3D dataset. We also describe our procedure for ground-truth annotation (Sec.~\ref{sec:appendix-mocap}), provide additional data statistics (Sec.~\ref{sec:appendix-statistics}) and quantitative results (Sec.~\ref{sec:appendix-results}).

\section{Aria glasses} \label{sec:appendix-aria}

\begin{figure}[t]
    \centering
    \includegraphics[width=0.9\linewidth]{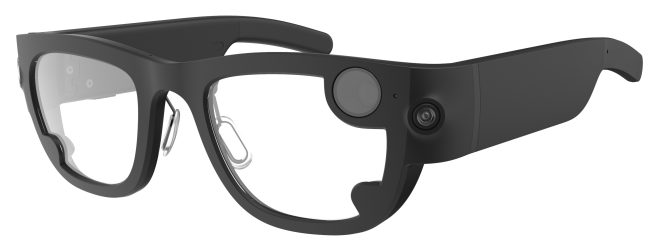}
    \caption{
        \textbf{Project Aria research glasses.}%
    }
    \vspace{-0.7ex}
    \label{fig:appendix:aria:device}
\vspace{2.6ex}
    \centering
\begin{minipage}{0.50\linewidth}%
\includegraphics[width=\linewidth]{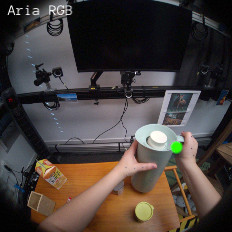}%
\end{minipage}
\begin{minipage}{0.235\linewidth}%
\includegraphics[width=\linewidth]{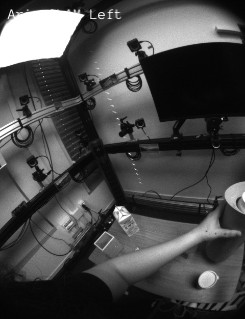}\\[0.5mm]%
\includegraphics[width=\linewidth]{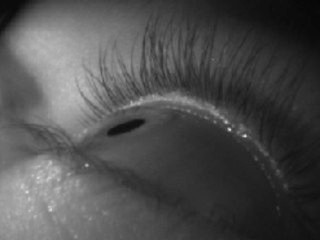}%
\end{minipage}
\begin{minipage}{0.235\linewidth}%
\includegraphics[width=\linewidth]{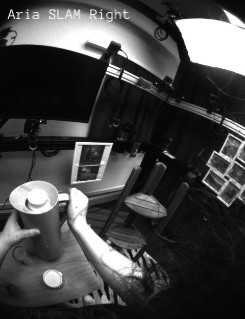}\\[0.5mm]%
\includegraphics[width=\linewidth]{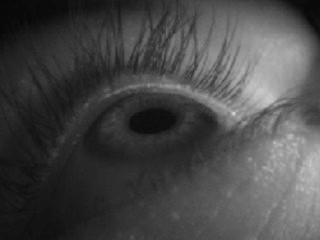}%
\end{minipage}\\%
\includegraphics[width=0.19\linewidth]{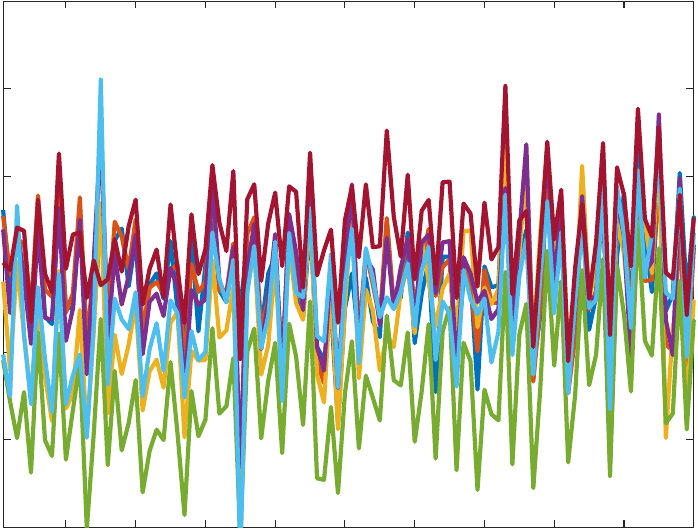}
\includegraphics[width=0.19\linewidth]{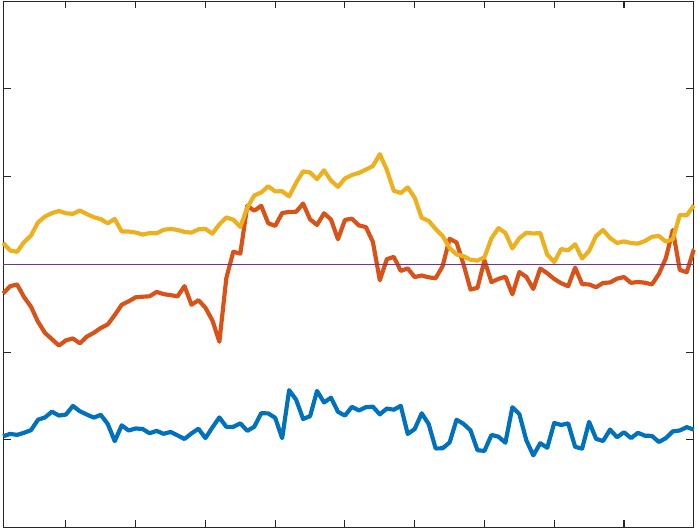}
\includegraphics[width=0.19\linewidth]{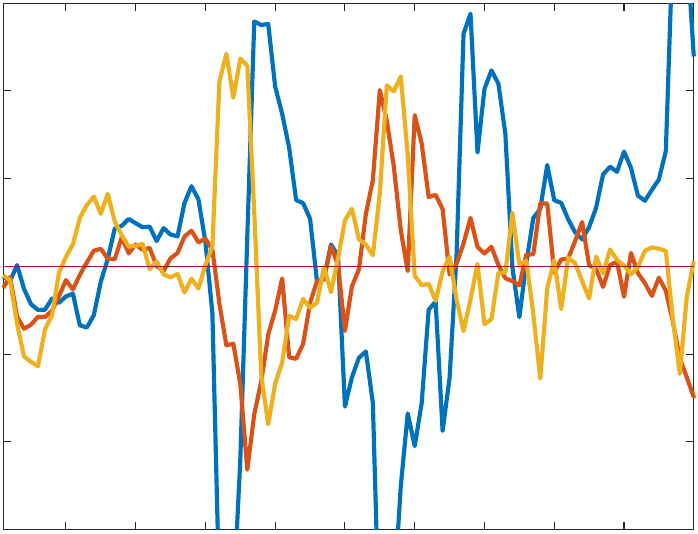}
\includegraphics[width=0.19\linewidth]{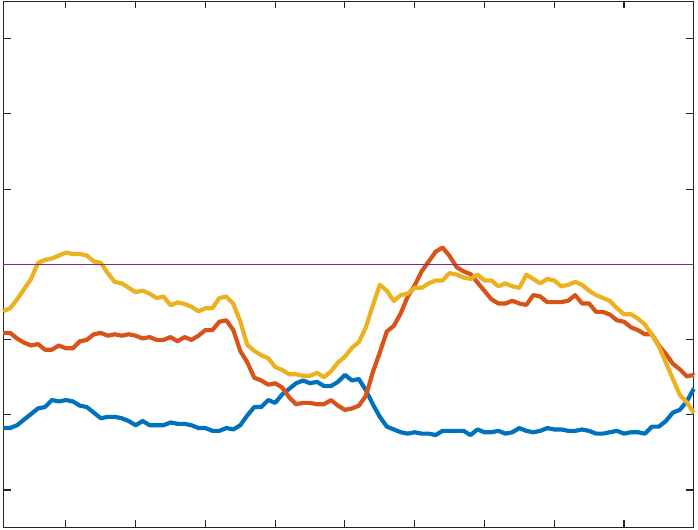}
\includegraphics[width=0.19\linewidth]{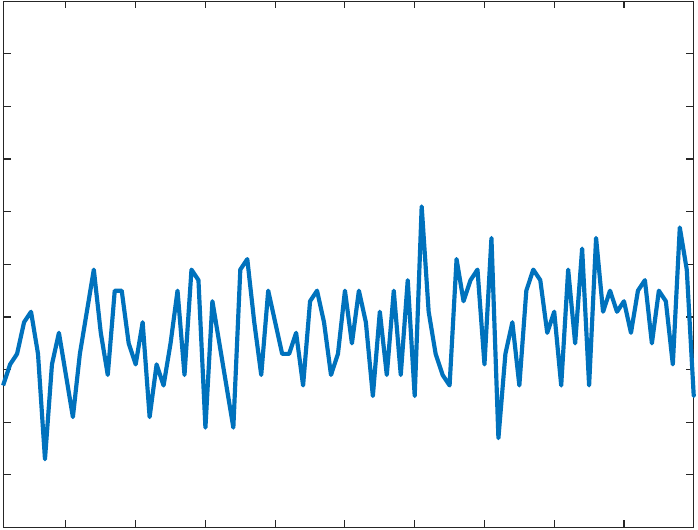}
    \caption{\textbf{Sensor streams recorded by the Project Aria device.} Top: RGB camera, left and right monochrome and eye cameras. Bottom: 10-second extracts from microphones, accelerometer, gyroscope, magnetometer and barometer respectively.}
    \label{fig:appendix:aria:sensors}
\end{figure}

Project Aria~\cite{engel2023project} is an egocentric recording device in glasses form-factor by Meta, designed as a \textit{research tool} for egocentric machine perception and contextualized AI research, and available to researchers across the world via \href{http://projectaria.com}{projectaria.com} (Fig.~\ref{fig:appendix:aria:device}).

\subsection{Device and sensors}
\label{appendix:aria:device}
Project Aria is built to emulate future AR/smart glasses catering to machine perception and egocentric AI rather than human consumption. Aria is designed to be wearable for long periods of time without obstructing or impeding the wearer, allowing for natural motion even when performing highly dynamic activities, such as playing soccer or dancing. Its total weight is 75g (a single GoPro camera has over 150g) and fits just like a pair of glasses. 

Further, the device integrates a rich sensor suite that is tightly calibrated and time-synchronized, capturing a broad range of modalities. For HOT3D, \textit{recording profile 15} is used, which uses the following sensor configuration (all streams come with metadata such as precise timestamps and per-frame exposure times):
\vspace{1ex}
\begin{itemize}
    \item \textbf{One rolling-shutter RGB camera} recording at 30\,fps and $1408 \times 1408$\,px. The camera is fitted with an F-Theta fisheye lens with a field of view (FOV) of $110^{\circ}$.
    \item \textbf{Two global-shutter monochrome cameras} recording at 30\,fps and $640 \times 480$\,px. These cameras provide additional peripheral vision and are fitted with F-Theta fisheye lenses with $150^{\circ}$~FOV.
    \item \textbf{Two monochrome eye-tracking cameras} recording at 10\,fps and $320 \times 240$\,px resolution.
    \item \textbf{Two IMUs} (800\,Hz and 1000\,Hz respectively), \textbf{a barometer} (50\,fps) and \textbf{a magnetometer} (10\,fps).
    \item \textbf{GNSS and WiFi} scanning was disabled for HOT3D. %
    \item \textbf{Audio} recording was disabled for HOT3D for privacy reasons. 
\end{itemize}

\begin{figure}[!th]
    \centering
{\setlength{\fboxsep}{0pt}\setlength{\fboxrule}{0.5pt}
\fbox{\includegraphics[width=0.325\linewidth]{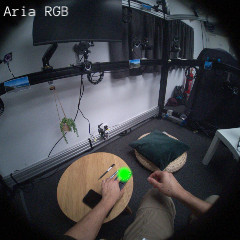}}\hfill%
\fbox{\includegraphics[width=0.325\linewidth]{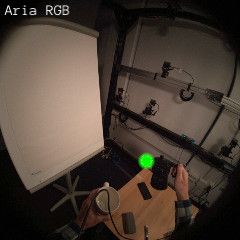}}\hfill%
\fbox{\includegraphics[width=0.325\linewidth]{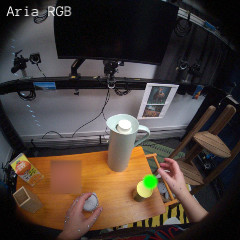}}\\[0mm]%
\fbox{\includegraphics[width=0.325\linewidth]{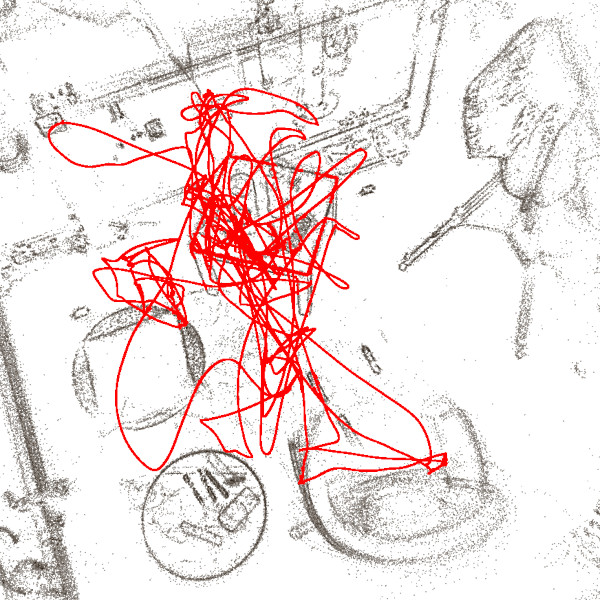}}\hfill%
\fbox{\includegraphics[width=0.325\linewidth]{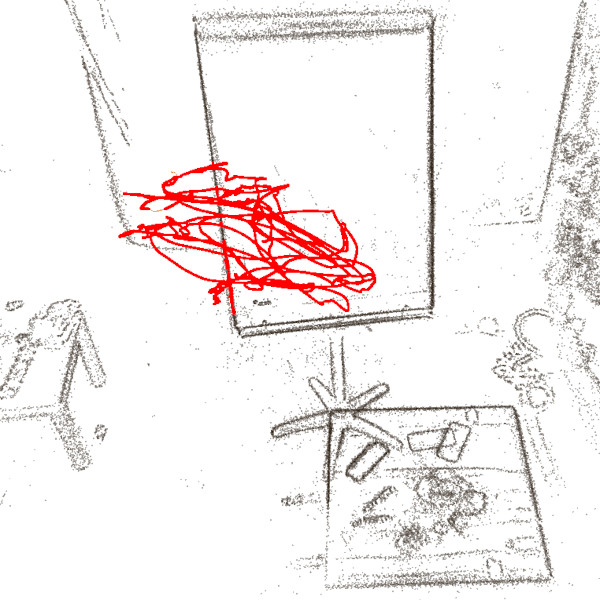}}\hfill%
\fbox{\includegraphics[width=0.325\linewidth]{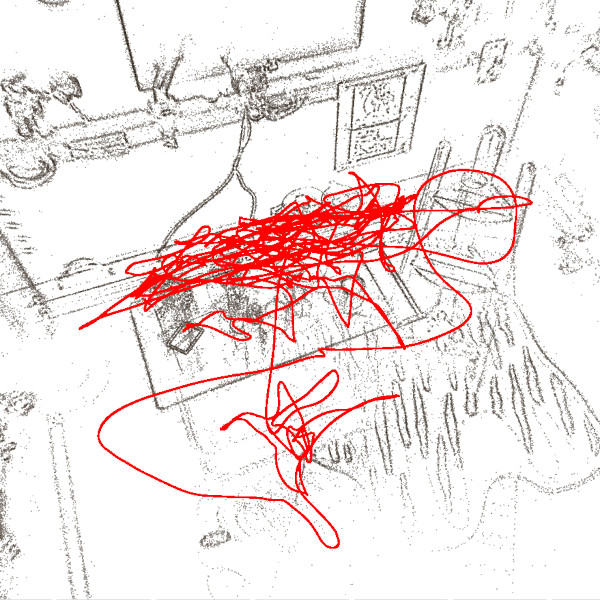}}\\[0mm]%
\fbox{\includegraphics[width=0.325\linewidth]{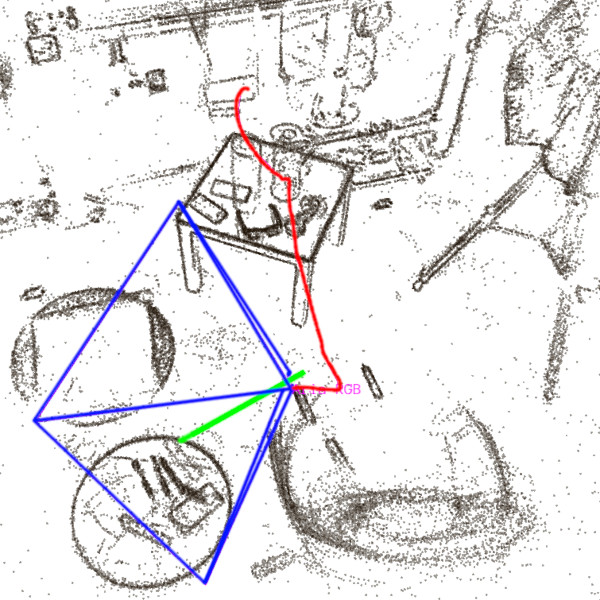}}\hfill%
\fbox{\includegraphics[width=0.325\linewidth]{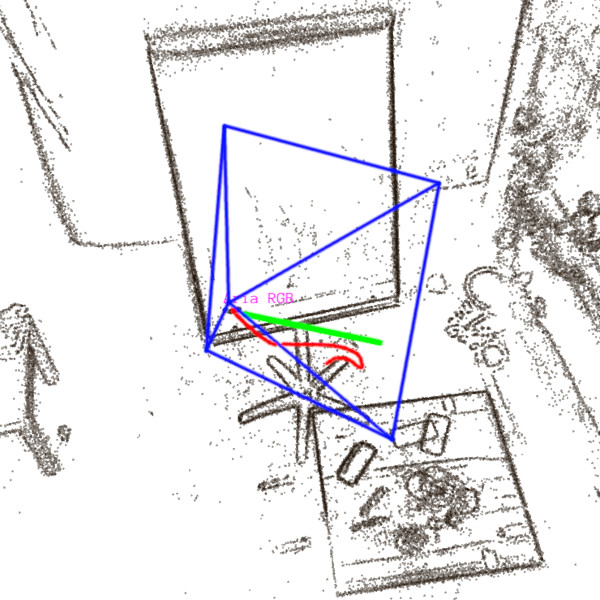}}\hfill%
\fbox{\includegraphics[width=0.325\linewidth]{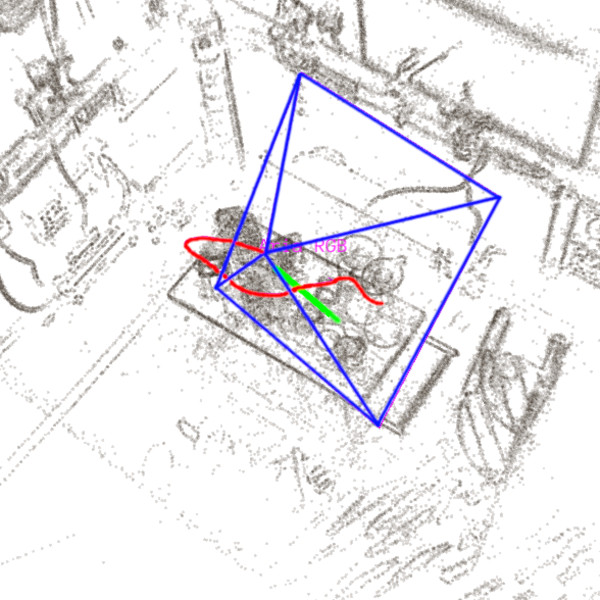}}%
}
    \caption{\textbf{Aria MPS output.} Shown is output for three recordings in a living room, office and kitchen scenario respectively (left to right). Top: RGB view and gaze (green dot). Middle: Point cloud and estimated egocentric camera trajectory for the full recording. Bottom: 3D view of a specific point in time, showing the RGB camera frustum (blue), gaze vector (green) and trajectory from the previous second (red).}
    \label{fig:appendix:aria:mps}
\end{figure}

\subsection{Machine Perception Services (MPS)}
\label{appendix:aria:mps}
Project Aria's machine perception service (MPS) provides software building blocks that simplify leveraging the different modalities recorded. These functionalities are likely to be available as real-time, on-device capabilities in future AR- or smart-glasses. We use the following core functionalities currently offered by Project Aria, and include their raw output as part of the dataset. See \cite{engel2023project} and the technical documentation\footnote{\href{https://facebookresearch.github.io/projectaria_tools/docs/intro}{https://facebookresearch.github.io/projectaria\_tools/docs/intro}} for more details. 

\customparagraph{Calibration.} All sensors are intrinsically and extrinsically~calibrated, and tiny deformations due to temperature changes or stress applied to the glasses frame are further corrected by time-varying online calibration from MPS.

\customparagraph{Aria 6\,DoF localization.} Every recording is localized precisely and robustly in a common, metric, gravity-aligned coordinate frame, using a state-of-the-art VIO and SLAM algorithm. This provides millimeter-accurate 6\,DoF poses for every captured frame and high-frequency (1\,kHz) motion in-between frames. 

\customparagraph{Eye gaze.} The gaze direction of the user is estimated as two outward-facing rays anchored approximately at the wearer's eyes, allowing to approximately estimate not only the direction the user is looking in, but also the depth their eyes are focused on. We use an optional eye gaze calibration procedure, where the mobile companion app directs the wearer to gaze at a pattern on the phone screen while performing specific head movements. This information was then used to generate a more accurate eye gaze direction, personalized to the particular wearer.

\customparagraph{Point clouds.} A 3D point cloud of static scene elements is triangulated from the moving Aria device, using photometric stereo over consecutive frames or left/right SLAM camera. Points are added causally over time, and will include points on any object that is observed while static for several seconds. The output contains both the 3D point clouds as well as the raw 2D observations of every point in the camera images it was triangulated from.

\subsection{Processing summary} 
All Aria recordings are anonymized in a very first step, using the public EgoBlur \cite{raina2023egoblur} model and following Project Aria's responsible innovation principles. 

Then, the MPS pipeline is invoked for each full Aria recording, which are typically about 2 minutes long and include many instances of hand-object interactions with different objects.
Next, we 7DoF-align the MPS output with the OptiTrack coordinate frame (Sec.~\ref{sec:appendix-mocap}). In total, we have processed 199 Aria recordings with a total length of 391 minutes.
\vspace{3ex}

\subsection{Tools and ecosystem}
\label{appendix:aria:tools}
Documentation and open-source tooling for Aria recordings and MPS output is available on GitHub\footnote{\href{https://github.com/facebookresearch/projectaria_tools}{https://github.com/facebookresearch/projectaria\_tools}}
and includes Python and C++ tools to convert, load, and visualize data, as well as sample code for common
computer vision tasks.

\section{Quest 3 headset} \label{sec:appendix-quest}

\begin{figure}[t]
    \centering
    \includegraphics[width=0.8\linewidth]{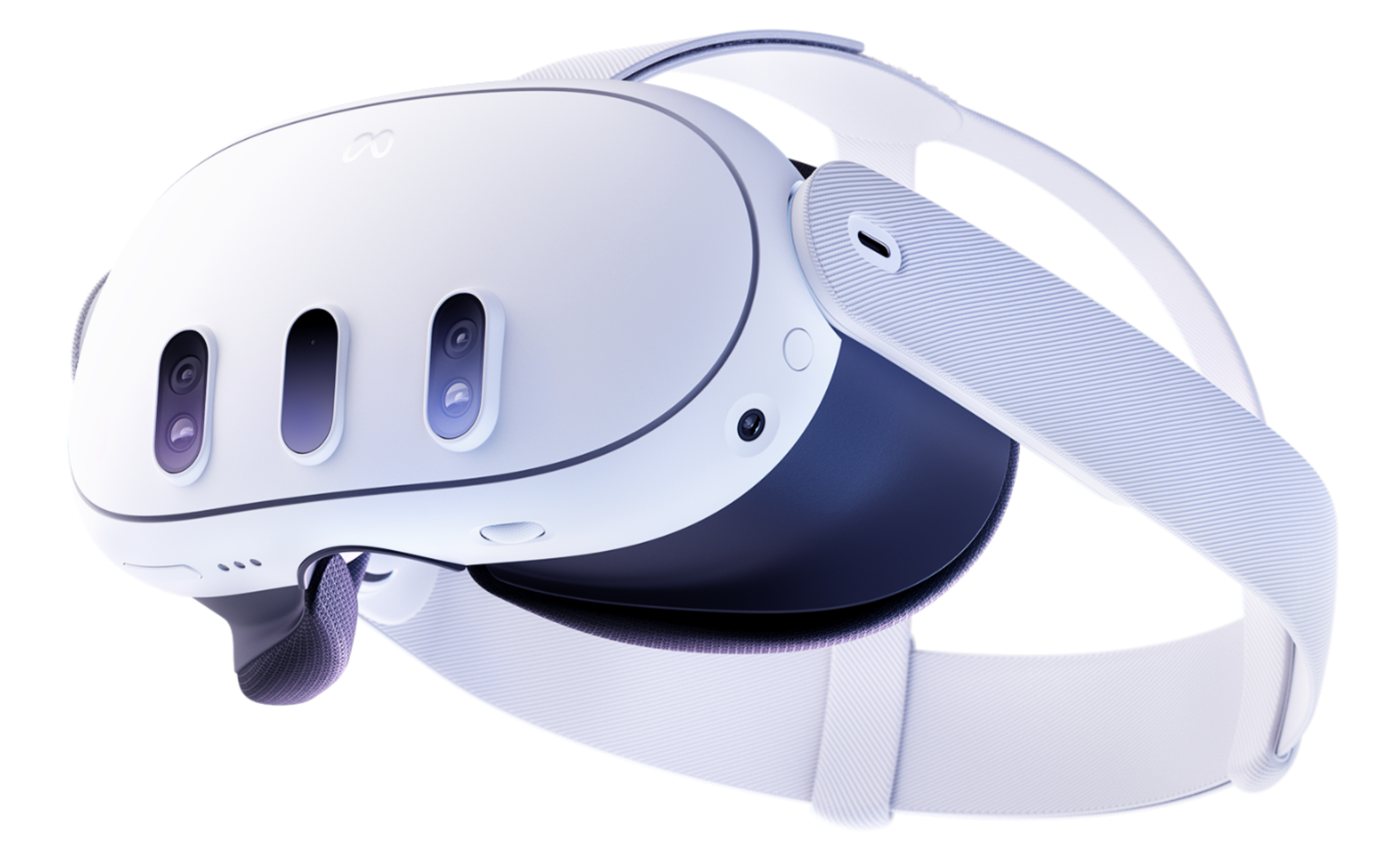}
    \caption{
        \textbf{Meta Quest 3 headset for virtual and mixed reality.}
    }
    \vspace{-0.1ex}
    \label{fig:appendix:quest3:device}
\end{figure}

\begin{figure}[t]
    \centering
    \setlength{\tabcolsep}{1pt} %
    \renewcommand{\arraystretch}{0.6} %

    \begin{tabular}{ccccc}
        \includegraphics[width=0.235\linewidth]{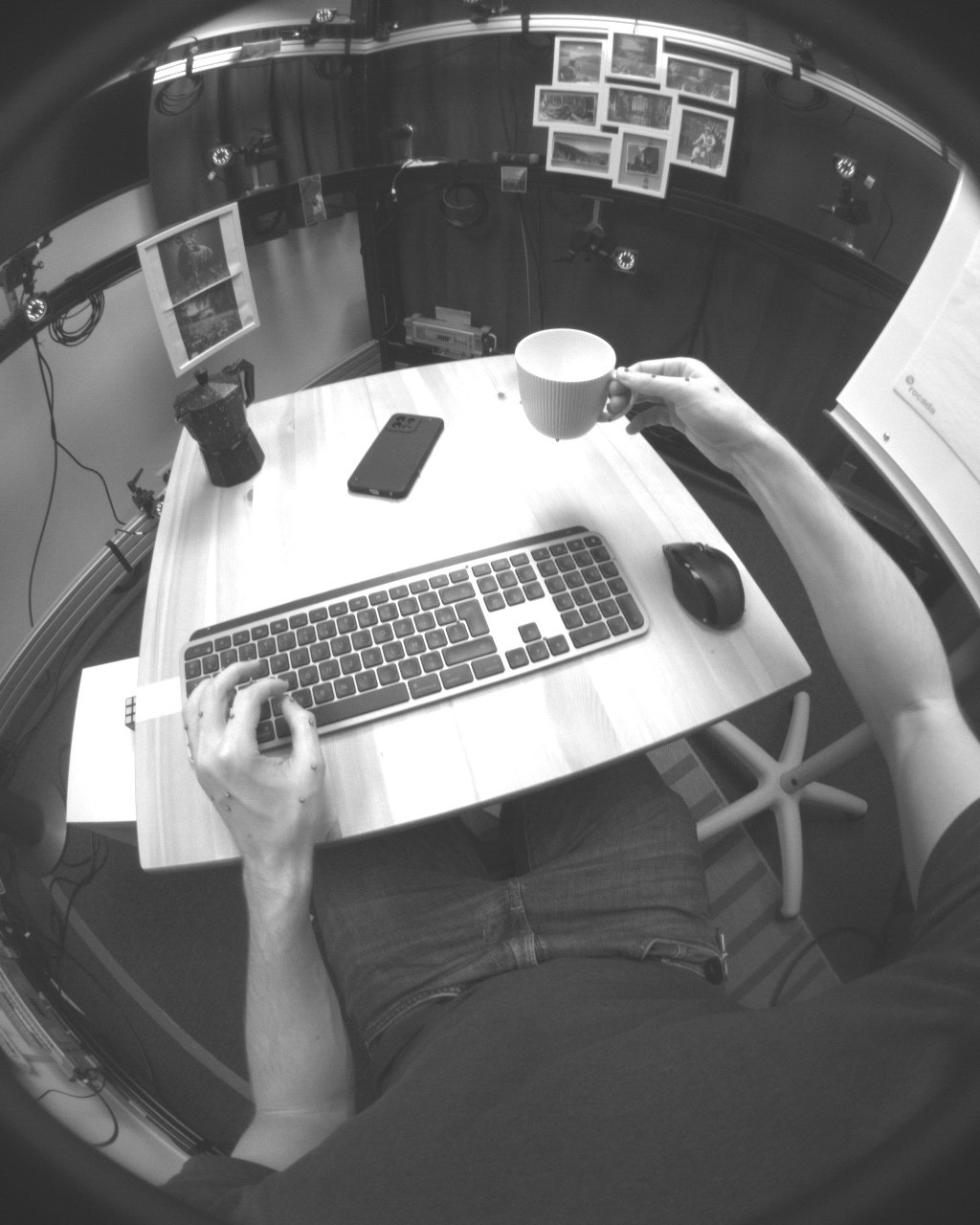} &
        \includegraphics[width=0.235\linewidth]{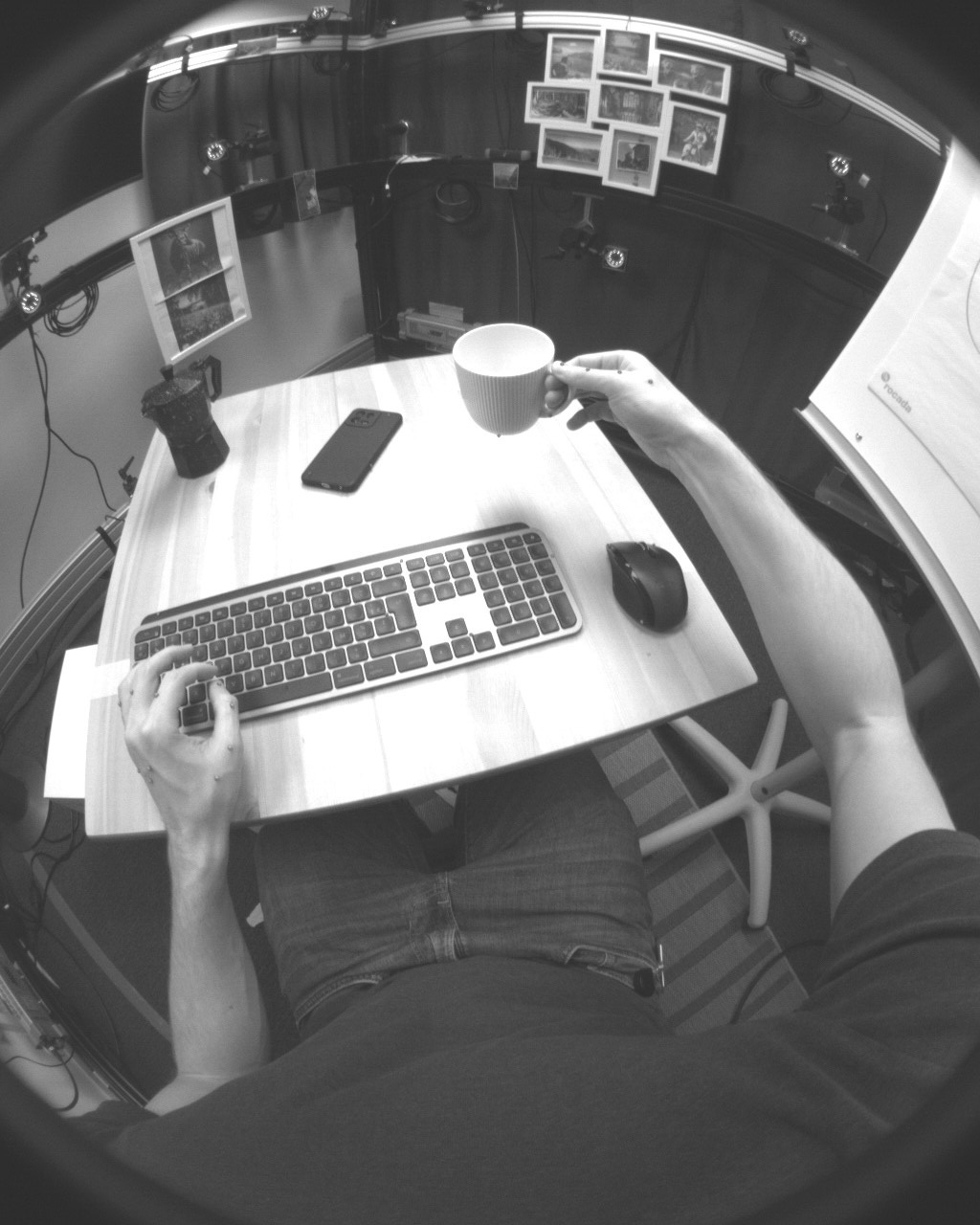} & \phantom{0} &
        \includegraphics[width=0.235\linewidth]{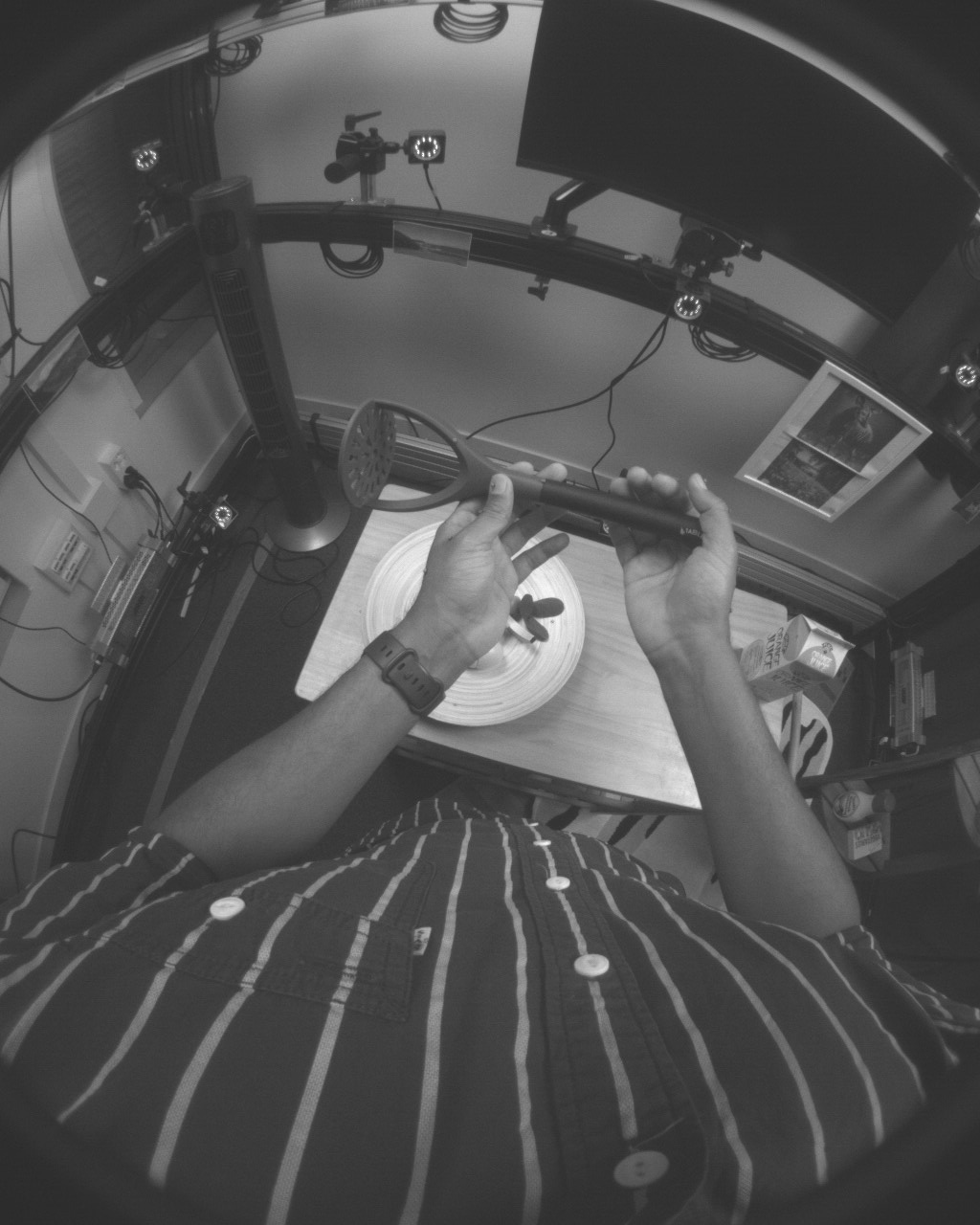} &
        \includegraphics[width=0.235\linewidth]{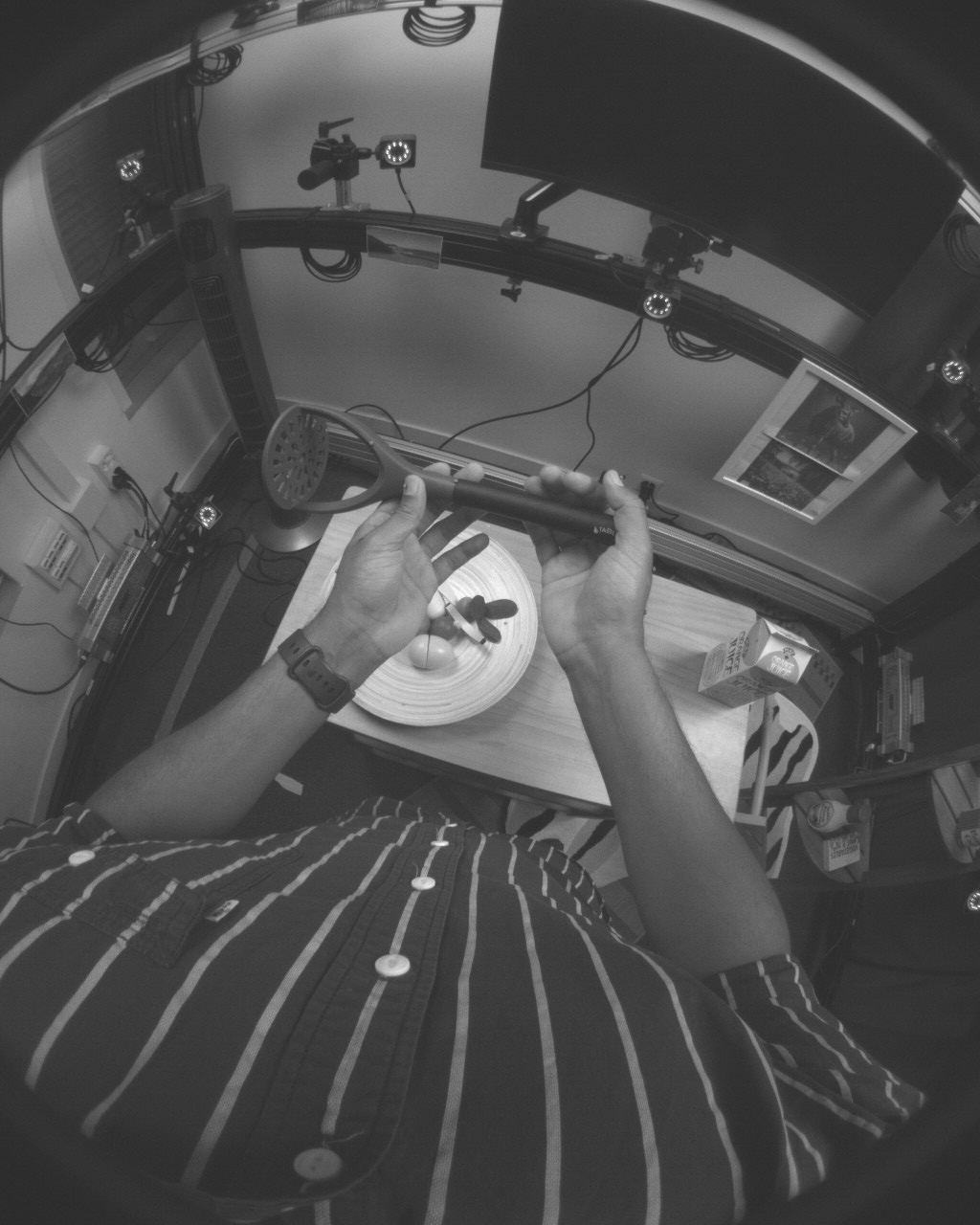}
    \end{tabular}
    \vspace{-0.7ex}
    \caption{
        \textbf{Sample images from Quest 3.} Shown are synchronized images from the two front Quest 3 cameras used for the HOT3D collection.
    }
    \label{fig:appendix:quest3:images}
\end{figure}

Quest 3~\cite{Quest3}, shown in Fig.~\ref{fig:appendix:quest3:device}, is the latest production headset from Meta for virtual- and mixed-reality experiences.
For the HOT3D data collection we used
a developer version of the Quest 3 headset. This version has four global-shutter monochrome cameras with fisheye lenses, 1280x1024\,px image resolution, 18\,PPD (Pixels Per Degree), and records at 30\,fps.
Two of the cameras are on the front side of the headset, roughly aligned with eyes, and two on the sides. HOT3D only includes images from the two front cameras as those capture the relevant scene part (the two side cameras are useful for applications like SLAM). Example images are in Fig.~\ref{fig:appendix:quest3:images}.
Data from other sensors present in the consumer version of Quest~3, including a gyroscope and an accelerometer, were not recorded. The intrinsic and extrinsic parameters of the headset cameras were calibrated with a ChArUco board. Both the headset and the board were attached a set of optical markers and tracked by the motion-capture system described in Sec.~\ref{sec:appendix-mocap}, which allowed to estimate camera-to-headset transformations. At recording time, the headset pose was still tracked by the motion-capture system and used to calculate per-frame camera-to-world transformations.

\section{Marker-based motion capture} \label{sec:appendix-mocap}

The poses of hands and objects were tracked using optical markers attached on their surface. For both hands and objects we used 3\,mm markers with an adhesive layer at their bottom. Such markers are small enough not to influence hand-object interactions. Each hand was attached 19 markers and each object around 10. The marker locations were then semi-automatically registered to 3D models of hands and objects obtained by custom 3D scanners.

At recording time, the optical markers were tracked by multiple infrared OptiTrack
cameras attached on a rig shown in Fig.~5 of the main paper. The intrinsic and extrinsic parameters of the infrared cameras were calibrated before every capturing session.
Hand poses were calculated by fitting the participant's UmeTrack hand model~\cite{han2022umetrack} to the tracked optical markers, as in~\cite{MocapHT_Siggraph2018}. Object poses were estimated by aligning the tracked markers to their registered 3D locations in the model coordinate frame. To achieve reliable tracking, it was important to ensure that the marker constellation on each object is sufficiently distinct. Data frames from different sources were synchronized with SMPTE timecode.

\begin{figure}[t!]
    \centering
\begin{minipage}{\linewidth}%
\includegraphics[width=\linewidth]{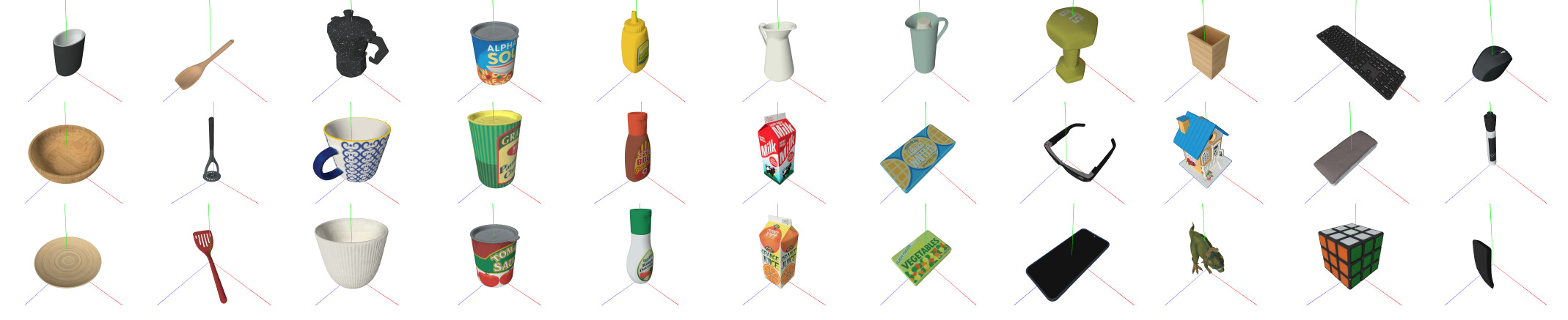}%
\end{minipage}
\begin{minipage}{\linewidth}%
\includegraphics[width=\linewidth]{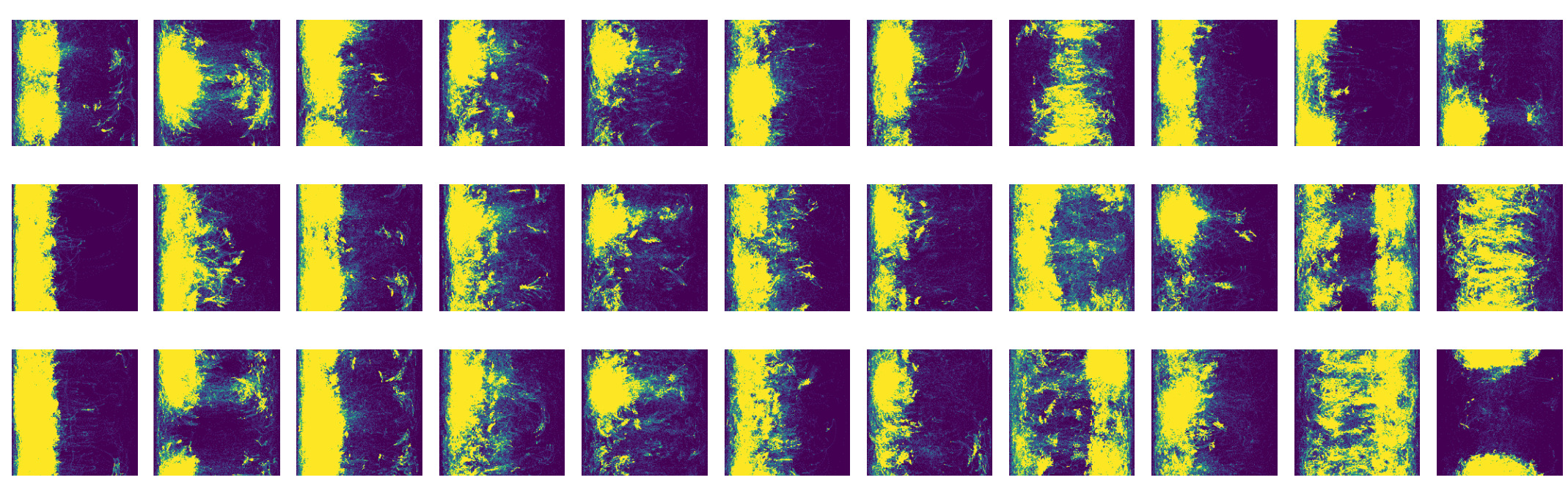}%
\end{minipage}
\vspace{-0.7ex}
\caption{\textbf{Object orientation statistics.} Top: 3D object models in their canonical poses. Bottom: Distribution of azimuth and elevation angles under which the objects are observed across the dataset. The vertical axis is the azimuth angle $[0^\circ,\;360^\circ]$ (angle along the green axis), and the horizontal axis is the elevation angle $[-90^\circ,\;90^\circ]$ (angle \wrt the plane defined by the red and blue axes).}
\label{fig:appendix:statistics:angular}
\end{figure}

\section{Object orientation statistics} \label{sec:appendix-statistics}

When recording, we asked subjects to naturally interact with the objects.
Consequently, orientation distributions of the observed objects (Fig.~\ref{fig:appendix:statistics:angular}) reveal clear object-specific pose biases, which may be useful as prior information at inference (we see 
that the bowl tends to be seen upright, the birdhouse from the front and upright, \etc).

\section{Additional quantitative results} \label{sec:appendix-results}

The results of 2D segmentation and 3D lifting of in-hand objects presented in Tab.~\ref{tab:2d_seg} and \ref{tab:inhandobj_lift3d} were obtained by evaluating methods on clips from both training and test splits. To allow the community to compare their results against our results on these two tasks,
in Tab.~\ref{tab:2d_seg_supp} and \ref{tab:inhandobj_lift3d_supp} we additionally provide results obtained on clips from the training split for which the ground-truth annotations are publicly available. Evaluating on the training split is possible as both of these tasks are training-free and therefore do not require any training split.

\begin{table*}[t!]
	\setlength{\tabcolsep}{3.5pt}
	\small
	\begin{center}
		\begin{tabularx}{1.0\linewidth}{c c Y Y Y Y}
			\toprule
			    & & \multicolumn{4}{c}{Object in hand (mIoU $\uparrow$) for training\,+\,test / training / test split} \\
			\cmidrule(lr){3-6}
			Method & Test dataset & Either & Left & Right & Both \\
			\toprule
			EgoHOS~\cite{zhang2022fine} & EgoHOS & -- & 62.2 & 44.4 & 52.8 \\
			\midrule
			EgoHOS~\cite{zhang2022fine} & HOT3D-Aria & 42.6 / 43.5 / 40.1 & 21.0 / 21.0 / 21.3 & 26.3 / 25.8 / 28.2 & 32.5 / 33.0 / 30.5 \\
			MRCNN & HOT3D-Aria & 47.1 / 48.1 / 43.7 & -- & -- & -- \\
			MRCNN-DA & HOT3D-Aria & \textbf{55.2} / \textbf{56.3} / \textbf{51.6} & -- & -- & -- \\
                \midrule
            EgoHOS~\cite{zhang2022fine} & HOT3D-Quest3 & 33.1 / 33.8 / 31.4 & 13.5 / 12.9 / 15.0 & 14.4 / 13.9 / 15.6 & 24.8 / 25.6 / 22.9 \\
            MRCNN & HOT3D-Quest3 & 37.8 / 38.2 / 36.9 & -- & -- & -- \\
            MRCNN-DA & HOT3D-Quest3 & \textbf{54.7} / \textbf{54.7} / \textbf{54.8} & -- & -- & -- \\
        \bottomrule
	\end{tabularx}
	\end{center}
        \vspace{-0.7ex}
	\caption{\textbf{2D segmentation of in-hand objects.} Each cell shows the mIoU score achieved on the training\,+\,test, training, and test split, respectively.
    }
	\label{tab:2d_seg_supp}
\end{table*}

\setlength{\dashlinedash}{1pt}
\setlength{\dashlinegap}{1pt}
\setlength{\arrayrulewidth}{0.2pt}
\begin{table*}[t!]
	\setlength{\tabcolsep}{3.5pt}
	\small
	\begin{center}
		\begin{tabularx}{1.0\linewidth}{c c c Y Y Y Y}
			\toprule
                & & & \multicolumn{4}{c}{Recall [\%] $\uparrow$ for training\,+\,test / training / test split} \\
                \cmidrule{4-7}
                Method & Test dataset & Views & 5\,cm & 10\,cm & 20\,cm & 30\,cm \\
                \toprule
                HandProxy & HOT3D-Aria & -- & 0.5 / 0.5 / 0.6 & 13.5 / 11.6 / 20.2 & 90.6 / 89.9 / 93.3 & 98.4 / 98.0 / 99.3\\
			\midrule
                \multicolumn{7}{l}{{\footnotesize Using ground-truth 2D segmentation masks:}}\vspace{1.7pt} \\
			  MonoDepth & HOT3D-Aria & 1 & 14.3 / 13.4 / 17.5 & 30.2 / 28.8 / 34.8 & 53.6 / 51.7 / 60.4 & 69.9 / 68.2 / 76.0 \\
			StereoMatch\ & HOT3D-Aria & 3 & \textbf{64.4} / \textbf{65.0} / \textbf{62.6} & \textbf{86.2} / \textbf{86.3} / \textbf{86.0} & \textbf{95.5} / \textbf{95.1} / \textbf{96.8} & \textbf{96.9} / \textbf{96.6} / \textbf{98.3} \\
                \hdashline
                StereoMatch\ & HOT3D-Quest3 & 2 & 76.4 / 78.0 / 72.8 & 96.8 / 96.9 / 96.5 & 99.1 / 99.2 / 99.1 & 99.2 / 99.2 / 99.1 \\
			\midrule
                \multicolumn{7}{l}{{\footnotesize Using 2D segmentation masks predicted by MRCNN-DA:}}\vspace{1.7pt} \\
			MonoDepth & HOT3D-Aria & 1 & 11.1 / 10.6 / 12.7 & 23.3 / 22.4 / 26.5 & 43.7 / 42.6 / 47.5 & 58.2 / 57.5 / 60.8 \\
                StereoMatch\ & HOT3D-Aria & 3 & \textbf{42.6} / \textbf{43.9} / \textbf{38.4} & \textbf{56.4}/ \textbf{57.6} / \textbf{52.2}  & \textbf{63.6} / \textbf{64.9} / \textbf{59.1} & \textbf{66.0} / \textbf{67.4} / \textbf{61.2} \\
                \hdashline
                StereoMatch\ & HOT3D-Quest3 & 2 & 59.1 / 60.1 / 56.9 & 75.3 / 75.6 / 74.6 & 80.4 / 80.6 / 79.9 & 81.3 / 81.6 / 80.7 \\
			\bottomrule
		\end{tabularx}
	\end{center}
        \vspace{-0.7ex}
	\caption{\textbf{3D lifting of in-hand objects.} Each cell shows the recall rate achieved on the training\,+\,test, training, and test split, respectively.
    \vspace{10cm}
    }
	
	\label{tab:inhandobj_lift3d_supp}
\end{table*}

\end{document}